%% file: thesis.tex
\documentstyle[11pt,subfloat,treemacros,dt,slashbox,titlepage]{book}

\begin{document}
\pagestyle{headings}
\pagenumbering{roman}

\input{psfig.tex}

\begin{titlepage}

\title{Corpus-Based Word Sense Disambiguation}

\author{\Large Atsushi Fujii \\
  Ph.D. thesis \\
  Department of Computer Science \\
  Tokyo Institute of Technology \\
  March 1998
}

\date{Supervisor : Hozumi Tanaka}

\maketitle

\end{titlepage}

\newcommand{\E}{\mbox{$e$}}
\newcommand{\EX}[2]{\mbox{${\cal E}_{#1}^{#2}$}}
\newcommand{\EXi}[2]{\mbox{${\cal E}_{#1,#2}$}}
\newcommand{\EXii}[3]{\mbox{${\cal E}^{#1}_{#2,#3}$}}
\newcommand{\V}{\mbox{$v$}}
\newcommand{\SS}{\mbox{$s$}}
\newcommand{\SSi}[1]{\mbox{$s_{#1}$}}
\newcommand{\C}{\mbox{$c$}}
\newcommand{\Ci}[1]{\mbox{$c_{#1}$}}
\newcommand{\Ni}[1]{\mbox{$n_{#1}$}}
\newcommand{\Mi}[1]{\mbox{$m_{#1}$}}
\newcommand{\X}[1]{\mbox{${\cal X}_{#1}$}}
\newcommand{\eq}[1]{(\ref{#1})}
\newcommand{\set}[1]{\mbox{\bf #1}}
\newcommand{\seti}[2]{\mbox{$\set{#1}_{#2}$}}
\newcommand{\colcenter}[1]{\hfill\centering #1\hfill}
\newcommand{\shortcite}[1]{\cite{#1}}
\newcommand{\etal}{et~al.}
\newcommand{\etaleos}{et~al}

\newcommand{\capital}[1]{{\footnotesize #1}}

\newcounter{exnum}
\setcounter{exnum}{1}
\newcommand{\exnum}{\theexnum\addtocounter{exnum}{1}}

\chapter*{Preface}
\addcontentsline{toc}{chapter}{Preface}

Resolution of lexical ambiguity, commonly termed ``word sense
disambiguation'', is expected to improve the analytical accuracy for
tasks which are sensitive to lexical semantics. Such tasks include
machine translation, information retrieval, parsing, natural language
understanding and lexicography.  Reflecting the growth in utilization
of machine readable texts, word sense disambiguation techniques have
been explored variously in the context of corpus-based
approaches. Within one corpus-based framework, that is the
similarity-based method, systems use a database, in which example
sentences are manually annotated with correct word senses. Given an
input, systems search the database for the most similar example to the
input.  The lexical ambiguity of a word contained in the input is
resolved by selecting the sense annotation of the retrieved example.

In this research, we apply this method of resolution of verbal
polysemy, in which the similarity between two examples is computed as
the weighted average of the similarity between complements governed by
a target polysemous verb.  We explore similarity-based verb sense
disambiguation focusing on the following three methods. First, we
propose a weighting schema for each verb complement in the similarity
computation. Second, in similarity-based techniques, the overhead for
manual supervision and searching the large-sized database can be
prohibitive.  To resolve this problem, we propose a method to select a
small number of effective examples, for system usage. Finally, the
efficiency of our system is highly dependent on the similarity
computation used. To maximize efficiency, we propose a method which
integrates the advantages of previous methods for similarity
computation.

\chapter*{Acknowledgments}
\addcontentsline{toc}{chapter}{Acknowledgments}

First and foremost, I would like to thank my supervisor, Hozumi Tanaka
(TITECH, Japan) for his guidance, support and encouragement throughout
the years of my studentship. I would also like to thank Takenobu
Tokunaga (TITECH, Japan) for his enormous support with this research,
and other members of my thesis committee: Sadaoki Furui, Hiroshi
Maruyama and Masayuki Numao (TITECH, Japan).

Additionally, I would like to express my gratitude to members of the
Tanaka \& Tokunaga laboratories (TITECH, Japan) for their
contributions to this research, amongst whom, I can never thank
Kentaro Inui and Timothy Baldwin enough.

Manabu Okumura (JAIST, Japan), Michael Zock (LIMSI, France) and Dan
Tufis (Romanian Academy, Romania) provided invaluable comments on this
research, for which I am greatly indebted.

Last but not least, I would also like to thank Masayuki Kameda (RICOH
Co., Ltd., Japan) for his support with the QJP parser, and Wakako
Kashino and Minako Hashimoto (Information-technology Promotion
Agency, Japan) for their support with the corpus used in this
research.

This research was partially supported by a Research Fellowship of the
Japan Society for the Promotion of Science for Young Scientists.

\tableofcontents
\listoftables
\listoffigures
\clearpage
\pagenumbering{arabic}
\setcounter{page}{1}

\chapter{Introduction}
\label{cha:introduction}

\section{Background}
\label{sec:background}

Natural language processing (NLP) involves resolution of various types
of {\em ambiguity}. Lexical ambiguity is one of these ambiguity types,
and occurs when a single word (lexical form) is associated with
multiple senses or meanings. For applications which are sensitive to
semantic denotation, or more precisely lexical semantics, this
ambiguity type can pose a major obstacle.  Resolution of lexical
ambiguity, which is commonly termed ``word sense disambiguation''
(WSD), is expected to improve the quality of the following research
fields.

\begin{itemize}
\item Machine translation (MT) can safely be identified as one of the
  major beneficiaries of word sense disambiguation, because a single
  word in a source language is frequently associated with multiple
  translations in a target language, each of which is often associated
  with a different sense to the source word. For example, the ``tax''
  and ``obligation'' senses of the English word {\it duty\/}
  correspond to the French translations of {\it droit\/} and {\it
    devoir}, respectively. Sense disambiguation of {\it duty\/} is
  expected to allow this word to be translated appropriately for a
  given context.  In fact, a number of MT-oriented word sense
  disambiguation methods have been explored based on this
  notion~\cite{brown:acl-91,dagan:cl-94}.

\item Information retrieval (IR) and text categorization (TC) suffer
  from the effects of noisy words associated with multiple senses, and
  IR/TC systems can easily end up relating documents containing the
  same words but in different senses (usages). For example, documents
  containing the word {\it AIDS\/} can easily be associated with those
  containing the word {\it aids}. Conventional systems have
  tentatively avoided this problem through usage of inverse document
  frequency (IDF)~\cite{salton:83}. The rationale behind IDF is that
  words which rarely occur over document collections are valuable, or
  that in other words, the IDF of a word is inversely proportional to
  the number of documents containing that word.  As
  Krovets~\shortcite{krovets:acmtois-92} identified, this can be
  recast in the context of lexical ambiguity. That is, words of higher
  frequency also tend to be associated with a greater number of
  senses, and therefore the degree to which these noisy words affect
  the system output can be minimized by introducing IDF.  However, as
  a number of experimental results have shown/suggested, word sense
  disambiguation is a crucial task for the further improvement of
  IR/TC
  systems~\cite{fukumoto:coling-96,krovets:acmtois-92,krovets:acl-eacl-97,sanderson:sigir-94,schutze:cl-x,schutze:sdair-95}.

\item Syntactic analysis (or parsing) often fails to identify the
  correct syntactic structure for an input sentence when syntactic
  relations are associated with semantic content. Prepositional phrase
  (PP) attachment problems\footnote{Ravin~\shortcite{ravin:acl-90},
    for example, focused on the resolution of PP-attachment problem
    through word sense disambiguation.} and predicate-argument
  structures associated with selectional restrictions are immediate
  examples of this problem type, in that they require the intervention
  of the semantic content of lexical entries for knowledge
  representation. Given the fact that syntactic and semantic (lexical)
  ambiguity are not independent of each other, a number of methods
  have proposed the mutual resolution of these two ambiguity
  types~\cite{lytinen:aaai-86,maruyama:acl-90,k.nagao:ieice-94}.

\item So called class-based NLP
  approaches~\cite{ker:cl-97,resnik:phd-93} are also potential
  beneficiaries of word sense disambiguation\footnote{A number of
    methods described in the previous item (``parsing'') can also be
    seen as instances of this category.}. These approaches involve
  mapping of each word entry to a semantic class (usually taken from a
  thesaurus taxonomy). Consequently, disambiguation of word senses is
  poignant, because each word is often associated with multiple class
  candidates, which are closely related to the senses of that word.

\item In natural language understanding (NLU), semantic structures are
  constructed by considering the meaning of each
  word~\cite{charniak:acl-88}. Kilgarriff~\shortcite{kilgarriff:nlprs-97}
  argued that current practical NLU systems, such as dialogue and
  information extraction (IE) systems, have commonly employed
  domain-specific knowledge representation rather than word sense
  disambiguation, in order to counter lexical ambiguity. However, we
  would like to note that his observation does not immediately reject
  the possibility of the potential contribution of word sense
  disambiguation to NLU systems.

\item Kilgarriff~\shortcite{kilgarriff:nlprs-97} also points to the
  advantages of word sense disambiguation in lexicography. By this is
  meant that sense-annotated linguistic data reduces the considerable
  overhead imposed on lexicographers in sorting large-scaled corpora
  according to word usage for different senses. Not only lexicography,
  but the general process of compiling linguistic resources is
  expected to improve through the interaction between lexicographers
  and computers~\cite{basili:ai-96,velardi:cl-91}.  In addition, word
  sense disambiguation techniques can also allow language learners to
  access example sentences containing a certain word usage from large
  corpora, without excessive overhead.
\end{itemize}
It should be noted that past word sense disambiguation methods have
not contributed to all the research given above. In fact, no
quantitative data has been forthcoming documenting empirical
improvement through word sense disambiguation, except for in the
fields of MT~\cite{brown:acl-91} and IR/TC~\cite{fukumoto:coling-96}.
One reason for this past failure to enhance existing methods is
presumably due to the immaturity of word sense disambiguation
research, which stimulates us to further explore this exciting
research area.

Reflecting the growth in utilization of machine readable texts, word
sense disambiguation techniques have been explored variously in the
context of ``corpus-based NLP approaches''\footnote{Corpus-based
  approaches have been explored in terms of other types of NLP
  research, a sample of which are reviewed, for example, by Church and
  Mercer~\shortcite{church:cl-93}.}.  These methods generally use a
corpus in which component words of each example sentence are annotated
(either manually or automatically) with their correct word sense, to
automatically induce rules or probabilistic models for disambiguation.
Unlike conventional rule-based approaches relying on {\em
  hand-crafted\/} selectional rules (some of which are reviewed, for
example, by Hirst~\shortcite{hirst:87} and
Small~\etal~\shortcite{small:88}), corpus-based approaches release us
from the task of generalizing observed phenomena through a set of
rules. While certain methods require manual annotation of the given
corpora (namely ``supervised methods'')\footnote{One may argue that
  supervised corpus-based methods have not released us from
  hand-encoding tasks. However, we would like to note that manual
  annotation is still easier than describing rule sets based on human
  introspection.}, other methods exclude or minimize the overhead for
supervision (namely ``unsupervised methods''). However, we observe
that the applicability of unsupervised methods has so far been limited
to relatively specific applications, and that the overhead for
supervision still remains as a major drawback of corpus-based word
sense disambiguation (we will elaborate on this issue in
Chapter~\ref{cha:past_research}).

\section{Focus of this Research}
\label{sec:focus}

First, let us precisely state the focus of this research, i.e. which
subcategory of word sense disambiguation we are targetting, given the
considerable variation in types and associated methods of lexical
disambiguation (between noun and adjective senses, for example). At
the same time, we consider the difficulty of the task, which can range
between coarse-grained (totally distinct) and fine-grained (closely
related) word sense distinctions. In this research, we explore
disambiguation of verb senses (verbal polysemy\footnote{We will define
  ``polysemy'' in Section~\ref{sec:task_description}.}) based on an
existing machine readable dictionary.  Note that the dictionary we use
provides relatively fine-grained verbal polysemy. Among the research
fields described in Section~\ref{sec:background}, our research focus
is expected to improve the quality of machine translation, parsing,
natural language understanding and lexicography. With regard to
current information retrieval systems, the disambiguation of noun
senses seems to be more crucial a task than that for verb senses (for
one thing, keywords and user queries usually comprised of noun
phrases). However, the potential contribution of verb sense
disambiguation extends to a significant proportion of the tremendous
range of information retrieval systems.

Second, let us describe the approach we use to tackle the verb sense
disambiguation task. Obviously, our research methodology must be
contextualized in terms of past research literature associated with
corpus-based word sense disambiguation, for which the reader should
refer to Chapter~\ref{cha:past_research}. In this process, it is
important $\underline{\mbox{not}}$ to limit our focus only to {\em
  verb\/} sense disambiguation, because methods employed in other
types of disambiguation may also be applicable. In brief, our system
employs a similarity-based method, in which disambiguation is
performed based on the similarity between a given input sentence and
example sentences associated with each verb sense. The similarity is
computed by averaging the similarity for each case (or case filler
noun) syntactically governed by the target verb. Through preliminary
experimentation on Japanese verbs, we identified a number of problems
associated with our research focus. In the following, we describe our
approaches to these problems.
\begin{itemize}
\item The degree to which each case contributes to verb sense
  disambiguation is not consistent. One may intuitively understand
  that in the case of English, for example, object case is more
  closely related to verb senses than subject case. The same
  observation can also be made in the case of Japanese. We explore a
  method of introducing this notion {\em computationally\/} into
  similarity-based methods, and demonstrate the effectivity of our
  proposal through comparison with a number of different methods (see
  Chapter~\ref{cha:vsd} for details).
\item As with most corpus-based word sense disambiguation systems, our
  system uses a large-scaled corpus annotated with correct verb
  senses.  However, a considerable overhead is required when one tries
  to manually perform the annotation process. One possible solution
  would involve automatic annotation, that is, an unsupervised method.
  However, our experiments show that at least for the particular
  unsupervised method targetted in our research, unsupervised methods
  still find it difficult to match the performance achieved for
  supervised methods. In view of this result, we explore a {\em
    semi\/}-supervised method in which we {\em selectively\/} sample a
  small number of effective example sentences from a given corpus in
  the annotation process. In other words, our sampling method aims at
  minimizing the overhead for manual supervision, without degrading
  the system performance (see Chapter~\ref{cha:sampling} for details).
\item We also identified that the performance of our system is highly
  dependent on the similarity computation between the input and
  example sentences. Roughly speaking, past approaches for similarity
  computation can be subdivided into statistically-driven and
  thesaurus-driven methods. In addition, integration methods combined
  these two approaches have recently been proposed. We also explore
  the similarity computation in the context of the integration of
  different methods, and demonstrate the effectivity of our proposal
  for our verb sense disambiguation system (see
  Chapter~\ref{cha:word_similarity} for details).
\end{itemize}

\section{Outline of the Proceeding Chapters}
\label{sec:outline}

Chapter~\ref{cha:past_research} surveys past research on word sense
disambiguation, and identifies open questions in the field.
Chapter~\ref{cha:vsd} describes the overall architecture of our
similarity-based verb sense disambiguation system, in which we newly
introduce the notion of degree of case contribution to verb sense.
Chapter~\ref{cha:sampling} proposes a selective sampling method, which
samples a smaller-sized, effective example set for use with our verb
sense disambiguation system, so as to minimize the overhead for
supervision and that to search a large-sized corpus.
Chapter~\ref{cha:word_similarity} explores and discusses the
similarity computation used in our system, through integration of
thesaurus taxonomy and co-occurrence statistics. Finally,
Chapter~\ref{cha:conclusion} summarizes our contributions and
discusses outstanding issues.

\chapter{Past Research on Word Sense Disambiguation}
\label{cha:past_research}

\begin{quote}
  \it Reflecting the rapid growth in utilization of machine readable
  texts, word sense disambiguation techniques have been explored
  variously in the context of ``corpus-based NLP approaches''. This
  chapter surveys past research associated with corpus-based word
  sense disambiguation, focusing mainly on methodology and evaluation
  criteria.
\end{quote}

\section{Overview: Terminology and Task Description}
\label{sec:task_description}

The task of a word sense disambiguation system (``system'' or ``WSD
system'', hereafter) is to resolve the lexical ambiguity of a
word\footnote{In most cases, past methods disambiguate only ``content
  word'', and ``functional words'' such as prepositions are beyond the
  scope of word sense disambiguation.} in a given context.  To put it
more precisely, the term ``lexical ambiguity'' refers to two different
concepts: ``homonymy'' and ``polysemy''. The former is the case where
two different words happen to have the same lexical form, and the
latter is the case where one word has several (related) meanings.
Conventionally, the distinction between {\it bank\/}~(``river edge'')
and {\it bank\/}~(``financial institution'') has been used as an
example of homonymy, and {\it rust\/}~(verb) and {\it rust\/}~(noun)
for polysemy.  In this dissertation, we will generally use the term
``polysemy'' to refer to both lexical ambiguity types, because (a) the
difference between these two ambiguity types has been relatively less
controversial in word sense disambiguation tasks (although from a
linguistic point of view, the two ambiguity types should be rigorously
defined), and (b) the focus of this research is on disambiguating
verbal polysemy rather than homonymy. To derive plausible word senses
(polysemy), most past WSD~systems have used lexical resources, such as
machine readable dictionaries (MRDs) or thesauri. Thus, the task of
WSD~systems can be termed as ``categorization'' because a plausible
word sense is selected from {\em predefined\/} candidates. Note that
this task should not to be confused with automatic identification of
word senses~\cite{fukumoto:coling-94,utsuro:coling-96,zernik:ijcai-89}
and word clustering
(grouping)~\cite{hatzivassiloglou:acl-93,hindle:acl-90,pereira:acl-93,tokunaga:ijcai-95}.
One may notice that the distinction between ``ambiguity'' and
``vagueness'' is also a controversial linguistics
issue~\cite{geeraerts:cogling-93,kilgarriff:phd-92,tuggy:cogling-93}.
A comprehensible example of vagueness would be {\it aunt}, which is
unspecific between ``father's sister'' and ``mother's sister'' without
explicit context.  However, this type of under-specification is beyond
the scope of word sense disambiguation unless dictionaries list
father/mother's sister as different senses of {\it aunt}.

Let us leave the definition of lexical ambiguity and turn to the
process of word sense disambiguation. Given an input sentence
containing polysemous word(s), most WSD~systems first preprocess the
input to extract a set of features (clues) used for the
disambiguation.  This preprocessing typically involves
morphological/syntactic analysis because the parts-of-speech of words
appearing in the input and syntactic relations involving polysemous
words, can be informative features. Macroy~\shortcite{macroy:cl-92},
for example, identified syntactic tags, morphology, collocations, and
word associations as the most important sources of information for
word sense disambiguation.  Thereafter the system interprets
polysemous word(s) by selecting a single plausible word sense. While
certain systems interpret only one polysemous word in the input, other
systems simultaneously interpret all polysemous words appearing in the
input.  Wilks and Stevenson~\shortcite{wilks:acl-siglex-ws-97} call
this second task type ``word sense tagging'' (based on the analogy of
part-of-speech tagging). However, the difference between the two task
types is relatively unimportant for this research.

Section~\ref{sec:different_methodologies} classifies past word sense
disambiguation methods.  Section~\ref{sec:evaluation_methodology} then
describes the way past research has evaluated word sense
disambiguation methods.  Section~\ref{sec:related_nlp_research}
describes related NLP research which is expected to enhance word sense
disambiguation.  Finally, discussion associated with word sense
disambiguation is added in Section~\ref{sec:past_discussion}.

\section{Different Methodologies for Word Sense Disambiguation}
\label{sec:different_methodologies}

This section surveys different past methods for corpus-based word
sense disambiguation, to clarify problems tackled in this research.
First, Section~\ref{subsec:classification} classifies past methods,
according to their induction mechanism (rule-based method or
probabilistic model, say). Second,
Section~\ref{subsec:unsupervised_learning} focuses on a different
viewpoint, that is, supervised vs. unsupervised learning methods.
While supervised methods require manual annotation of correct sense to
each polysemous word contained in a corpus, unsupervised methods
automatically acquire corpora annotated with (presumably correct) word
senses.  These automatic methods are what
Section~\ref{subsec:unsupervised_learning} focuses on principally.
It should be noted that criteria for the classification of past
methods can vary depending on the viewpoint or interest of the
researcher\footnote{For example, Wilks and
  Stevenson~\shortcite{wilks:acl-siglex-ws-97} classified past word
  sense disambiguation methods differently.}.

\subsection{General classification of methodologies}
\label{subsec:classification}

\begin{figure}[htbp]
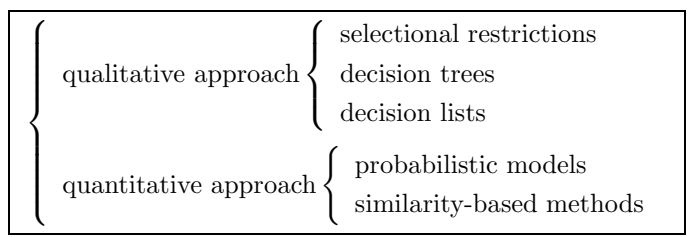

  \begin{center}
    \small
    \leavevmode
    \fbox{
      $\left\{
      \begin{array}{l}
        \mbox{qualitative approach}
        \left\{
          \begin{array}{l}
            \mbox{selectional restrictions} \smallskip \\
            \mbox{decision trees} \smallskip \\
            \mbox{decision lists}
          \end{array}
        \right.
        \medskip \\
        \mbox{quantitative approach}
        \left\{
          \begin{array}{l}
            \mbox{probabilistic models} \smallskip \\
            \mbox{similarity-based methods}
          \end{array}
        \right.
      \end{array}\right.$}
  \end{center}
  \caption{Classification of different methodologies of word sense disambiguation}
  \label{fig:different_methods}
\end{figure}

This section classifies past methodologies as shown in
Figure~\ref{fig:different_methods}, which first divides methods into
two different approaches. The first approach can be called
``qualitative approach''\footnote{One may notice that the terms
  ``symbolic/rule-based approach'' can be interchangeably used for
  qualitative approach. However, we used the term ``qualitative
  approach'' to make the direct contrast to \mbox{``quantitative
    approach''}.} and uses selectional rules associated with each word
sense candidate.  Given an input containing a polysemous word, the
system deterministically selects the sense(s) for which the rules are
satisfied.  Generally speaking, the granularity of selectional rules
can be a problem for this approach. That is, specified rules often
fail to select a word sense for exceptional inputs.  On the other
hand, generalized rules run the risk of selecting incorrect word
senses.  To counter this problem, the second approach -- what we call
the ``quantitative'' approach -- computes scalable values for each
word sense candidate, and selects the sense with maximal value as the
interpretation of the polysemous word. Compared with the rule-based
approach, this approach is more robust for exceptional
inputs\footnote{A number of {\em compromised\/} methods can be found
  in past research. Wilks~\shortcite{wilks:ai-75} used selectional
  restrictions as {\em preferences\/} rather than constraints.
  Uramoto~\shortcite{uramoto:ieice-94} combined qualitative and
  quantitative approaches.}.  Both approaches generally use a corpus,
in which examples are associated with each word sense candidate. We
shall call this corpus the ``training data''. Note that in this
section, examples in the training data can be annotated either {\em
  manually\/} or {\em automatically}, without loss of generality.  The
qualitative approach can be further subdivided in terms of rule
format. Subcategories described in this section are ``selectional
restrictions'', ``decision trees'' and ``decision lists''. The
quantitative approach can be divided in terms of the method used to
compute scalable values, as ``probabilistic models'' and
``similarity-based methods''.  All subcategories of these two
approaches correspond to the different titles of the following items.

\paragraph{Selectional restrictions}

Selectional restrictions~\cite{katz:63}, which impose constraints on
arguments for a given word (sense), have commonly been used in word
sense disambiguation relying on hand-crafted
rules~\cite{charniak:aaai-86,hirst:87,lytinen:aaai-86}.  Let us take
the following example sentences (a) and (b), which contain different
usages of the verb {\it employ}, that is, ``to hire'' and ``to
accept'':
\begin{list}{}{}
\item
  \label{ex:employ}
  \newcounter{employ}
  \setcounter{employ}{\theexnum}
  \begin{tabular}{lll}
    (\exnum) & a. The facility will {\it employ\/} new employees. & (``to
    hire'') \\
    & b. The committee {\it employed\/} his proposal. & (``to
    accept'')
  \end{tabular}
\end{list}
One can intuitively differentiate the senses of {\it employ\/} in
sentences (\mbox{\theemploy}-a) and (\mbox{\theemploy}-b) with the
complements of each {\it employ}. To be more precise, {\it employ\/}
in (\mbox{\theemploy}-a) restricts its subject and object nouns to
those associated with the semantic
features~\capital{HUMAN}/\capital{ORGANIZATION}~and~\capital{HUMAN},
respectively. On the other hand, {\it employ\/} in
(\mbox{\theemploy}-b) restricts its subject and object nouns to those
associated with the semantic
features~\capital{HUMAN}/\capital{ORGANIZATION}~and~\capital{IDEA},
respectively.  Consequently, given {\it employees\/} as the object,
the sense ``to hire'' is selected as the interpretation of {\it
  employ\/} in (\mbox{\theemploy}-a), and the sense ``to accept'' is
ruled out. The same reasoning can be used to select the sense ``to
accept'' as the interpretation of {\it employ\/} in
(\mbox{\theemploy}-b).  One may notice that selectional restriction
can also disambiguate polysemy of verb complements (the subject and
object). For example, {\it facility\/} in (\mbox{\theemploy}-a) has
multiple senses, a sample of which are ``installation'',
``proficiency'' and ``readiness''. However, the selectional
restriction imposed for the subject of {\it employ\/}~(``to hire'')
can correctly select the sense ``installation'' as the interpretation
of {\it facility}. It should be noted that the polysemy of both {\it
  facility\/} and {\it employ\/} are theoretically disambiguated
simultaneously. In fact, the disambiguation process can be seen as
mutually propagating semantic constraints to each polysemous word
through selectional
restriction\footnote{Lytinen~\shortcite{lytinen:aaai-86} and
  Nagao~\shortcite{k.nagao:ieice-94} used constraint propagation
  method to simultaneously resolve syntactic and semantic ambiguity.}.

However, considerable human effort is required to describe
large-scaled selectional restrictions. Manual construction is also
associated with human bias and inconsistency of granularity. Besides
this, revision requires additional human overhead, that is,
lexicographers have to laboriously identify and revise associated
entries.  Resnik~\shortcite{resnik:phd-93} proposed an
information-theoretic method to automatically identify selectional
restrictions, which is expected to counter this problem to some
degree. Resnik identified selectional restrictions as semantic classes
defined in the taxonomy of the well-known English semantic network
WordNet~\cite{miller:techrep-93}\footnote{WordNet terms semantic
  classes as ``synsets''.}, and only those nouns dominated in the
taxonomy by the identified class can satisfy the restriction.  While
Resnik used this method to resolve syntactic ambiguity,
Ribas~\shortcite{ribas:eacl-95} applied it to disambiguate senses of
case filler nouns (for example, disambiguation of the polysemy of {\it
  facility\/} in (\mbox{\theemploy}-a) as performed above). The basis
of this method is to estimate the information-theoretic association
degree between case fillers and semantic classes, for each verb (or
verb sense). Intuitively speaking, the association degree gives a
greater values for semantic classes that are likely to appear as
complement of a given verb sense.  Formally speaking, the association
degree between verb sense~$\SS$ and class~$r$ (restriction candidate)
with respect to case $\C$ is computed by
Equation~\ref{eq:ribas:95}~\cite{resnik:phd-93,ribas:eacl-95}.
\begin{equation}
  \label{eq:ribas:95}
  A(\SS,\C,r) = P(r|\SS,\C)\cdot\log\frac{\textstyle
    P(r|\SS,\C)}{\textstyle P(r|\C)}
\end{equation}
Here, \mbox{$P(r|\SS,\C)$} is the conditional probability that a case
filler marked with case~$\C$ of sense~$\SS$ is dominated by class~$r$
in the WordNet taxonomy.  \mbox{$P(r|\C)$} is the conditional
probability that a case filler marked with case~$\C$ (disregarding
verb sense) is dominated by class~$r$.  The distribution obtained from
training data is usually used to estimate each probability.

\paragraph{Decision trees}

\begin{figure}[htbp]
  \begin{center}
    \leavevmode
    \psfig{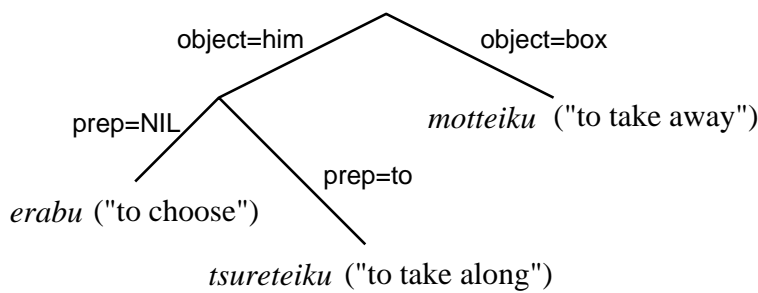}
  \end{center}
  \caption{A fragment of the decision tree for the English verb {\it
      take\/}}
  \label{fig:decision_tree}
\end{figure}

In spite of its long history of applications in AI research,
``decision trees'' have rarely been applied to word sense
disambiguation\footnote{Okumura and Tanaka~\shortcite{okumura:aaai-90}
  proposed generalized discrimination networks (GDNs), which can be
  seen a variant of decision trees, for word sense
  disambiguation. However, we will not discuss their method further
  here, because they focus on exploration of an ``incremental
  disambiguation model''~\cite{mellish:85} and not automatic
  construction of the networks the main topic of this research.}.
Among a number of proposed decision tree algorithms
\mbox{``C4.5''}~\cite{quinlan:93} has been used relatively commonly as
a benchmark comparison.  Mooney~\shortcite{mooney:emnlp-96} and
Pedersen and Bruce~\shortcite{pedersen:aaai-iaai-97} (individually)
compared the performance of various word sense disambiguation methods
with the \mbox{C4.5} algorithm.
Tanaka~\shortcite{hideki_tanaka:coling-94} used the \mbox{C4.5}
algorithm to acquire English-Japanese verbal translation rules. One
may notice that this task can also be seen as a type of rule induction
for word sense disambiguation, because a unique English verb can be
interpreted as different Japanese
verbs. Figure~\ref{fig:decision_tree} shows a fragment of the decision
tree for the English verb {\it take}, which corresponds to multiple
Japanese translations such as {\it erabu\/}~(``to choose''), {\it
  tsureteiku\/}~(``to take along'') and {\it motteiku\/}~(``to take
away''). Given an input containing {\it take\/} followed by the object
noun {\it him\/} and preposition {\it to}, the decision tree selects
{\it tsureteiku\/} as the interpretation, by traversing the branches
corresponding to \mbox{``{\tt object=him}''} and \mbox{``{\tt
    prep=to}''}.

Here, let us devote a little space to explaining the basis of the
\mbox{C4.5} algorithm.  In this decision tree algorithm,
classification rules are formulated by recursively partitioning the
training data. Each nested partition is based on the feature value
that provides the greatest increase in the information gain ratio for
the current partition. The final partitions correspond to a set of
classification rules where the antecedent of each rule is a
conjunction of the feature values used to form the corresponding
partition. Let \mbox{\{$C_1,C_2,\ldots,C_k$\}} denote class
candidates, one of which is assigned to the input. Suppose we already
have a partition, which divides the set~$T$ of training data into
subsets \mbox{\{$T_1,T_2,\ldots,T_n$\}}. Given that
\mbox{$freq(C_i,T)$} and $|T|$ denote the number of training data in
$T$ that belong to class $C_i$ and the number of training data in
set~$T$, respectively, the ``entropy'' (uncertainty) of set~$T$ can be
estimated by Equation~\ref{eq:c4.5_1}.
\begin{equation}
  \label{eq:c4.5_1}
  info(T) = -\sum_{i}\frac{\textstyle freq(C_i,T)}{\textstyle
    |T|}\cdot\log\frac{\textstyle freq(C_i,T)}{\textstyle |T|}
\end{equation}
Suppose $T$ is partitioned into $n$~subsets using feature value $X$ as
a classification rule, the expected entropy over partitioned subsets,
\mbox{$info_{X}(T)$}, is estimated by Equation~\ref{eq:c4.5_2}.
\begin{equation}
  \label{eq:c4.5_2}
  info_{X}(T) = \sum_{i}\frac{\textstyle |T_i|}{\textstyle
    |T|}\cdot info(T_i)
\end{equation}
The decrease in entropy resulting from the partition, which represents
the information gain for $X$, is then estimated by
Equation~\ref{eq:c4.5_3}.
\begin{equation}
  \label{eq:c4.5_3}
  gain(X) = info(T) - info_{X}(T)
\end{equation}

\paragraph{Decision lists}

Decision lists are a form of rule representation as proposed by
Rivest~\cite{rivest:ml-87}, and consist of tuples of the form
\mbox{``(condition, value)''}. As Rivest observes, decision lists can
be seen as ``if-then-else'' rules, in other words, exceptional
conditions appear earlier while general conditions appear later in the
list\footnote{The last condition accepts all cases (namely ``true''),
  otherwise the system could potentially fail to make any decision for
  certain input types.}.  Given a query, each condition in the
decision list is applied sequentially until a condition which is
satisfied by the query is found. Thereupon, the value which
corresponds to that condition is selected as the answer.  Yarowsky
applied decision lists to the task of accent
restoration~\cite{yarowsky:acl-94} (this is one type of lexical
disambiguation, in which a single word is associated with multiple
pronunciations), and word sense disambiguation~\cite{yarowsky:acl-95}.
In Yarowsky's case~\cite{yarowsky:acl-94,yarowsky:acl-95}, each
condition corresponds to a word collocation which can be used as
evidence to resolve lexical ambiguity, and each value corresponds to a
correct word sense (or pronunciation).  Since manual identification of
effective conditions is expensive and inconsistent, Yarowsky used word
collocation (within a fixed window size) obtained from a large corpus
to automatically identify effective evidence types. The effectivity
degree of a given piece of evidence is estimated as the likelihood
that it supports a given sense candidate more strongly than
another\footnote{In Yarowsky's
  case~\shortcite{yarowsky:acl-94,yarowsky:acl-95}, the number of
  sense candidates for each word was limited to two.}. Formally
speaking, this notion is represented as the log-likelihood, that is,
the ratio between the conditional probability that sense~$\SSi{1}$ and
sense~$\SSi{2}$ occur, respectively, given evidence~$E$
(Equation~\ref{eq:yarowsky:94}).
\begin{equation}
  \label{eq:yarowsky:94}
  \log\frac{\textstyle P(\SSi{1}|E)}{\textstyle P(\SSi{2}|E)}
\end{equation}
In decision lists, evidence (along with their supporting word senses)
is sorted according to log-likelihood, in descending order.
Figure~\ref{fig:yarowsky:95} shows a fragment of the decision list
trained for the disambiguation of the word~{\it
  plant\/}~(``organism''/``factory'')~\cite{yarowsky:acl-95}, where
each piece of evidence denotes a specific collocational pattern or
word collocating with a certain distance associated with {\it
  plant}. For example, given the input containing the pattern {\it
  plant~height\/}, the interpretation for {\it plant\/} is
``organism''.  Note that
Yarowsky~\shortcite{yarowsky:acl-94,yarowsky:acl-95} used a method
binary sense distinction method, in that the number of sense
candidates was limited to two. To apply this method to the
disambiguation of words with multiple ambiguity, the denominator in
Equation~\ref{eq:yarowsky:94} is presumably computed based on the
probability of (a) all the other sense candidates or (b) only the most
competitive sense candidate (that is, the sense candidate with the
second highest probability given evidence~$E$).

\begin{figure}[htbp]
  \begin{center}
    \small
    \leavevmode
    \begin{tabular}{|l|c|} \hline
      {\hfill\centering evidence\hfill} & sense \\ \hline
      {\it plant\/} growth & organism \\
      car (within $\pm k$ words) & factory \\
      {\it plant\/} height & organism \\
      union (within $\pm k$ words) & factory \\
      equipment (within $\pm k$ words) & factory \\
      {\hfill\centering $\vdots$ \hfill} & {\hfill\centering $\vdots$
        \hfill} \\ \hline
    \end{tabular}
  \end{center}
  \caption{A fragment of the decision list for the word~{\it plant\/}}
  \label{fig:yarowsky:95}
\end{figure}

Pedersen and Bruce~\shortcite{pedersen:aaai-iaai-97} used the
\mbox{``CN2''} rule induction algorithm~\cite{clark:ml-89} as one form
of comparison in their ``decomposable model''\footnote{See the
  following ``Probabilistic models'' item for details of the
  decomposable model.}.  The CN2 algorithm generates decision lists
based on a given set of training data. The rule induction algorithm
consists of finding rules, and measuring the significance of those
rules based on entropy and their coverage, over the range of the
training data. The quality of decision lists is highly dependent on
the sequence of rules, and thus naive rule sequences end up degrading
the disambiguation performance. To overcome this problem, the latest
version of CN2~\cite{clark:ewsl-91} optionally employs an ``unordered
rule'' technique. Let us explain, in passing, the basic algorithm
here. The training procedure is the same as in the original
version. However, given an input and decision rule set, rather than
applying the rules in a fixed order, the algorithm collects {\em
  all\/} the rules which the input satisfies, and their associated
conditions.  Note that by disregarding the rule order (in other words
the ``context''), different rules can potentially support different
values and consequently ambiguity often remains unresolved. In such a
case, additional consideration is given to the distribution of the
training data covered by the collected rules, in choosing a unique
value. Let us take Figure~\ref{fig:cn2} as an example rule
set~\cite{clark:ewsl-91}, where each line corresponds to a rule for
the two values of ``bird'' and ``elephant''. In this figure, the
``coverage'' column denotes the number of training examples which
satisfy the corresponding rule, for the values ``bird'' and
``elephant'', respectively. Let us consider the following example
input.
\begin{eqnarray*}
  \label{eq:cn2_input}
  <\!\mbox{\tt size=large, beaked=yes, legs=two, feather=yes,
    flies=no}\!>
\end{eqnarray*}
The rule sets in Figure~\ref{fig:cn2} cannot uniquely decide the
answer for this input, which satisfies all three rules. However, the
summation of the coverage for these rules ``(35,10)'' puts the
preference for the final answer on ``bird'' over ``elephant''.

\begin{figure}[htbp]
  \begin{center}
    \small
    \leavevmode
    \begin{tabular}{|llll|c|} \hline
      \multicolumn{4}{|c|}{rule} & coverage \\ \hline
      {\tt if} & {\tt legs=two and feather=yes} & {\tt then} & {\tt
        class=bird} & (13,0) \\
      {\tt if} & {\tt size=large and flies=no} & {\tt then} & {\tt
        class=elephant} & (2,10) \\
      {\tt if} & {\tt beaked=yes} & {\tt then} & {\tt class=bird} &
      (20,0) \\ \hline
    \end{tabular}
    \caption{An example rule set for CN2}
    \label{fig:cn2}
  \end{center}
\end{figure}

\paragraph{Probabilistic models}
\label{para:prob}

From the viewpoint of probability theory, the task of word sense
disambiguation is to select the sense with maximal probability for a
given input\footnote{Some statistical methods are reviewed, for
  example, by Charniak~\shortcite{charniak:93}.}. The probability for
word sense~$\SS$, \mbox{$P(\SS|x)$}, is commonly transformed into
Equation~\ref{eq:bayes} through use of the Bayesian theorem.
\begin{equation}
  \label{eq:bayes}
  P(\SS|x) = \frac{\textstyle P(\SS)\cdot P(x|\SS)}{\textstyle P(x)}
\end{equation}
In practice, \mbox{$P(x)$} can be omitted because this factor is
constant for all the sense candidates, and therefore does not affect
the relative probability for different senses
(Equation~\ref{eq:bayes2}).
\begin{equation}
  \label{eq:bayes2}
  \arg\max_{\SS}P(\SS|x) = \arg\max_{\SS}P(\SS)\cdot P(x|\SS)
\end{equation}
The probability of sense~$\SS$, \mbox{$P(\SS)$}, is usually estimated
based on its distribution, obtained from training data.  Thus, the
performance of the method fundamentally depends on the method for
approximating \mbox{$P(x|\SS)$}. For the purposes of general
discussion, let the input be represented by a vector comprised of word
sense disambiguation features, as given below.
\begin{eqnarray*}
  \label{eq:feature_vector}
  <\!F_1=f_1,~F_2=f_2,~\ldots,~F_N=f_N\!>
\end{eqnarray*}
Here, $F_i$ and $f_i$ are the $i$-th feature type and its value,
respectively.  Let us summarize the diverse range of past methods
based on the following two principles.

The first principle is to identify an informative feature set, and
ideally, different feature sets for each different target polysemous
word.  Typically, words which {\em saliently\/} collocate with a word
sense are used as features. In this case, each feature takes a binary
value, that is, 1 for existence and 0 for absence in the input.  A
number of methods have been proposed to automatically identify
informative collocating words (termed ``salient
words''~\cite{yarowsky:coling-92} or
\mbox{``indicators''~\cite{justeson:cl-95}}).
Yarowsky~\shortcite{yarowsky:coling-92} used mutual information
between $w$ and sense~$\SS$ (in Yarowsky's case, word senses are
semantic categories defined in Roget's thesaurus~\cite{chapman:84}) to
estimate the degree of salience of word~$w$ to sense~$\SS$.
Intuitively speaking, the mutual information between two phenomena
gives a greater value when these phenomena are more likely to
co-occur.  The mutual information of $w$ and $\SS$, \mbox{$I(w;\SS)$},
is computed as shown in Equation~\ref{eq:mutual_information}.
\begin{equation}
  \label{eq:mutual_information}
  I(w;\SS) = \log\frac{\textstyle P(w|\SS)}{\textstyle P(w)}
\end{equation}
Here, \mbox{$P(w|\SS)$} is the probability that $w$ appears given
$\SS$, and \mbox{$P(w)$} is the probability that $w$ appears in the
context\footnote{Strictly speaking, \mbox{$P(w|\SS)$} and
  \mbox{$P(w)$} should be denoted as \mbox{$P(F_{w}\!=\!1|\SS)$} and
  \mbox{$P(F_{w}\!=\!1)$}, respectively, where $F_w$ is the feature
  representing the existence of word~$w$. However, we use a simplified
  notation here.}. These factors are estimated based on the relative
distributions of $w$ and $\SS$ in the training data.
Figure~\ref{fig:yarowsky:92} shows examples of salient words related
to the categories \capital{ANIMAL} and
\capital{TOOLS}~\cite{yarowsky:coling-92}. Intuitively speaking, when
words like {\it species\/} and {\it family\/} appear in the input, the
probability for \capital{ANIMAL} tends to be greater than that for
\capital{TOOLS}. Justeson and Katz~\shortcite{justeson:cl-95} select
word~$w$ as indicator of sense $\SS$ such that $w$ appears more
frequently with sense~$\SS$ than with other sense candidates.  This is
equivalent to selecting a word~$w$ that satisfies
Equation~\ref{eq:justeson:95}, where $t$ is any other sense candidate
of the polysemous word under evaluation\footnote{In practice,
  \mbox{$P(\SS|w)$} and \mbox{$P(t|w)$} are transformed using the
  Bayesian theorem as for Equation~\ref{eq:bayes}.}.
\begin{equation}
  \label{eq:justeson:95}
  P(\SS|w) > P(t|w)
\end{equation}
Their objective is to disambiguate adjective senses, or in their case,
polysemy of an adjective as defined by its antonyms. For example, the
word {\it old\/} has two senses, i.e. ``not new'' and ``not
young''. Figure~\ref{fig:justeson:95} shows examples of indicator
nouns identified in their paper~\cite{justeson:cl-95}.  Note that in
Justeson and Katz's case, the disambiguation process in itself does
not rely on a probabilistic model, or more precisely, indicators are
used simply as restriction rules. Ng and Lee~\shortcite{ng:acl-96} and
Pedersen~\etal~\shortcite{pedersen:anlp-97} use multiple feature types
along with collocating words. The following additional features are
also usually used\footnote{Ng and Lee~\shortcite{ng:acl-96} use these
  features for exemplar-based word sense disambiguation, which will be
  described in the next ``Similarity-based methods'' item.}:
\begin{itemize}
\item the morphological properties of polysemous words (for example,
  singular/plural in the case of a polysemous noun, and tense in the
  case of a polysemous verb),
\item parts-of-speech of collocating words,
\item syntactic relations associated with polysemous words.
\end{itemize}

\begin{figure}[htbp]
  \begin{center}
    \small
    \leavevmode
    \begin{tabular}{|l|l|} \hline
      {\hfill \centering catogory\hfill} & {\hfill \centering
        salient words\hfill} \\ \hline
      \capital{ANIMAL} & species, family, bird, fish \\
      \capital{TOOLS} & tool, machine, engine, blade \\ \hline
    \end{tabular}
  \end{center}
  \caption{Example salient words for two categories}
  \label{fig:yarowsky:92}
\end{figure}

\begin{figure}[htbp]
  \begin{center}
    \small
    \leavevmode
    \begin{tabular}{|l|l|} \hline
      {\hfill \centering sense\hfill} & {\hfill \centering
        indicators\hfill} \\ \hline
      {\it old\/}~(``not new'') & world, thing, car, way \\
      {\it old\/}~(``not young'') & man, people, woman, wine \\ \hline
    \end{tabular}
  \end{center}
  \caption{Example indicator nouns for the adjective {\it old\/}}
  \label{fig:justeson:95}
\end{figure}

The second principle is to compute (approximate) \mbox{$P(x|\SS)$},
based on the conditional probability of feature(s) given sense~$\SS$.
The most simple model, the ``Naive-Bayes method'', assumes that
features are conditionally independent of each other given
sense~$\SS$. That is, \mbox{$P(x|\SS)$} is approximated simply by the
product of each \mbox{$P(F_i\!=\!f_i|\SS)$}, as shown in
Equation~\ref{eq:bayes3}.
\begin{equation}
  \label{eq:bayes3}
  P(x|\SS) \simeq \prod_{i}P(F_i\!=\!f_i|\SS)
\end{equation}
A number of past research applied the Naive-Bayes method to word sense
disambiguation~\cite{gale:ch-92,mooney:emnlp-96,leacock:arpa-hlt-93,ng:emnlp-97,pedersen:emnlp-97,pedersen:aaai-iaai-97}.

A more complex model, namely the ``decomposable
model''~\cite{bruce:acl-94,pedersen:anlp-97}, considers
interdependency between different features. Here, let features
\mbox{$(F_1,F_2,F_3)$} represent the input. Suppose $F_1$ and
\mbox{$(F_2,F_3)$} are interdependent given $\SS$, such that
\mbox{$P(x|\SS)$} can be expressed by Equation~\ref{eq:bayes4}.
\begin{equation}
  \label{eq:bayes4}
  P(x|\SS) = P(F_1=f_1|\SS)\cdot P(F_2=f_2,F_3=f_3|\SS)
\end{equation}
In the decomposable model, the number of parameters to be estimated,
which is proportional to the number of combinations of values for
interdependent features, can be enormous.  This often leads to the
data sparseness problem. In addition, it is difficult to identify
which features are interdependent given a particular sense.  To
overcome this problem, Pedersen~\etal~\shortcite{pedersen:anlp-97}
proposed a method of automatically identifying the optimal model
(higher performance and fewer parameter estimations), by iteratively
altering the complexity level of the model.  However, their
experimental results show that identified models did not generally
outperform the Naive-Bayes method.

\paragraph{Similarity-based methods}

Given problems along with their solution (the way they have been
solved), humans are able to solve new problems based on the analogy of
the previously observed cases. This analogy-based process has been
explored in many AI applications, under the headings of case-based
reasoning~\cite{kolodner:93}, exemplar-based
reasoning~\cite{bareiss:90}, memory-based reasoning~\cite{weiss:91}
and instance-based
learning~\cite{aha:ml-91}\footnote{Nagao~\shortcite{nagao:ahi-84}, for
  example, explored the analogy principle in NLP applications.}.  A
number of word sense disambiguation systems have variously applied
these methods\footnote{In the case of word sense disambiguation, the
  reasoning mechanism is relatively simple when compared with tasks
  focused on in AI research (for example, resolution of a ``political
  dispute'' focused on by Kolodner~\shortcite{kolodner:93}), because
  previous problems are merely example sentences containing polysemous
  words annotated with senses.}.  One critical issue in this has been
the computation of {\em similarity\/} between an input (a new problem)
and examples in the training data (previous problems), and thus, we
term these methods ``similarity-based methods''. In the $k$-nearest
neighbor method ($k$-NN), one similarity-based method, processing
proceeds as follows. First, $k$ examples similar to the input are
retrieved from the training data.  Thereafter, retrieved examples {\em
  vote\/} on the sense of the polysemous word in the input, or in
other words, the sense receiving the highest frequency of annotation
with the $k$ examples is selected as the interpretation of the input
word.  In the case of \mbox{$k = 1$}, this method is termed the
nearest neighbor method, in which the input word is disambiguated
simply by superimposing the word sense associated with the example of
highest similarity.  Generally speaking, the nearest neighbor method
has been used more commonly than $k$-NN in past word sense
disambiguation research\footnote{Ng~\shortcite{ng:emnlp-97}
  automatically identified the optimal value of $k$ over the range of a
  given training data, which reportedly improved the performance of
  the nearest neighbor method.}. In the following, we classify past
similarity-based methods in terms of the method of computing the
similarity (or distance).

\begin{figure}[htbp]
  \begin{center}
    \leavevmode \psfig{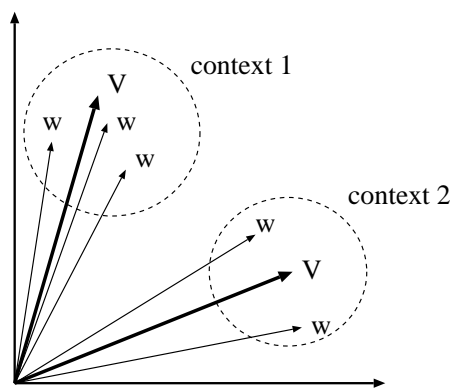}
  \end{center}
  \caption{The word and context vectors for two contexts}
  \label{fig:vsm}
\end{figure}

Suppose each training example and the input (``examples'', hereafter)
are represented by a feature vector defined by
Equation~\ref{eq:feature_vector}, in which one may notice that each
example is positioned in an $N$-dimensional space, where feature $F_i$
corresponds to the $i$-axis.  One common such implementation termed
the ``vector space model'' (VSM) computes the similarity between two
examples by the angle between the two vectors representing the
examples. Note that VSM has a long history of application in
information retrieval (IR) and text categorization (TC)
tasks~\cite{salton:83}. However, in the case of IR/TC, VSM is used to
compute the similarity between documents, which is represented by a
vector comprising statistical factors of content words in each
document. Formally speaking, the similarity between two examples $x$
and $y$ is computed as the cosine of the angle between their
associated vectors. This can be expressed by
Equation~\ref{eq:cosine_sim}, where $\vec{x}$ and $\vec{y}$ are
vectors representing examples $x$ and $y$, respectively.
\begin{equation}
  \label{eq:cosine_sim}
  sim(x,y) = \frac{\textstyle \vec{x}\cdot \vec{y}}{\textstyle
    |\vec{x}||\vec{y}|}
\end{equation}
Sch\"{u}tze~\shortcite{schutze:supercomp-92} applies the vector space
model to word sense disambiguation, although vectors are used for each
word sense not individual examples\footnote{Within the IR/TC
  community, this approach is called ``text-to-category comparison'',
  and contrasted with ``text-to-text comparison'' which computes the
  similarity between the input and individual examples.}. First,
Sch\"{u}tze represents each word by a ``word vector'', that is, a
vector comprising the statistics (e.g. frequency) of its collocating
words. The collocational statistics are usually collected within a
fixed proximity.  Then, each context is represented by a ``context
vector'', which is the sum (or ``centroid'') of word vectors related
to words appearing in the context. Figure~\ref{fig:vsm} illustrates
the idea of the context and word vectors, in which each $w$ denotes a
word vector corresponding to a context vector $V$. It should be noted
that unlike the vector space model in IR/TC, Sch\"{u}tze's method
returns a positive similarity value even when two given context
vectors have no words in common (such as with contexts 1 and 2 in
Figure~\ref{fig:vsm}), that is, two context vectors can be similar
when they comprise similar word vectors. Then, automatic clustering
algorithms~\cite{cheeseman:cml-88,cutting:sigir-92} are used to
cluster each polysemous word into word senses, which are also
represented by ``sense vectors''\footnote{Sch\"{u}tze used two
  different clustering algorithms, depending on the data size.}. In
practice, Sch\"{u}tze used the ``singular value decomposition'' (SVD)
technique~\cite{berry:ijsa-92,golub:89}, which finds the major axes
and reduces the dimension of the vector space.  Finally,
Equation~\ref{eq:cosine_sim} computes the similarity between the input
and each word sense cluster, to select the word sense with maximal
similarity. Leacock~\etal~\shortcite{leacock:arpa-hlt-93} compared the
vector space model, the Naive-Bayes method (see the previous
paragraph) and neural networks, and reported that the vector space
model and neural networks slightly outperformed the Naive-Bayes
method. Niwa and Nitta~\shortcite{niwa:coling-94} also explored the
vector space model, in an implementation resembling Sch\"{u}tze's
method. The vector representation is expressed by
Equation~\ref{eq:niwa_nitta1}, where \mbox{$V(\SS)$} and \mbox{$V(w)$}
are the context vector for sense~$\SS$ and the word vector for
word~$w$ within a fixed proximity from $\SS$, respectively.
\begin{equation}
  \label{eq:niwa_nitta1}
  V(\SS) = \sum_{w \in context} V(w)
\end{equation}
In Niwa and Nitta's case, the word vector for $w$, \mbox{$V(w)$},
consists of the mutual information between $w$ and ``origin words''
(commonly used words), as in Equation~\ref{eq:niwa_nitta2}.
\begin{equation}
  \label{eq:niwa_nitta2}
  V(w) = <I(w;o_1),~I(w;o_2),~\ldots,~I(w;o_i),~\ldots>
\end{equation}
Here, \mbox{$I(w;o_i)$} denotes the mutual information between $w$ and
origin word~$o_i$.  In practice, origin words are made up of the 51st
to 1050th most frequently used words in Collins English Dictionary
definitions~\cite{sinclair:87}. Equation~\ref{eq:cosine_sim} is used
to compute the similarity between two context vectors. However, unlike
Sch\"{u}tze's method, which averages context vectors into sense
vectors, Niwa and Nitta prefer to use the ``text-to-text comparison''
method: they compute the similarity between the input and {\em
  individual\/} examples.

One may argue that the ``Euclidean distance'' between two examples in
$N$-dimensional space can also be applied to the similarity
computation (as performed by Aha~\etal~\shortcite{aha:ml-91} for
instance-based learning).  However, this similarity measure seems less
effective for word sense disambiguation, because Euclidean distance is
sensitive to vector length, which is usually proportional to the
frequency of collocating words.  In other words, Euclidean distance
usually assigns lower similarities to frequently appearing word
senses.

Ng and Lee~\shortcite{ng:acl-96}, following Cost and
Salzberg~\shortcite{cost:ml-93}, used a different similarity measure.
In their method, the distance between two examples is computed by
summing the distances between the feature values associated with those
examples. In other words, two examples are similar when they have
feature values that roughly correspond in distribution as obtained
from the training data.  The distance between two feature values $f_1$
and $f_2$ of a feature $F$ is computed as shown in
Equation~\ref{eq:cost_dist}.
\begin{equation}
  \label{eq:cost_dist}
  dist(f_1,f_2) = \sum_{i}\left|P(\SS|F\!=\!f_1) -
    P(\SS|F\!=\!f_2)\right|
\end{equation}
Here, \mbox{$P(\SS|F\!=\!f_1)$} is the conditional probability of
sense~$\SS$ given feature~$F$ takes value~$f_1$.
\mbox{$P(\SS|F\!=\!f_2)$} denotes a similar probability for
value~$f_2$ of feature~$F$.  Ng and Lee~\shortcite{ng:acl-96} used
multiple features described in the previous item (see the item
``Probabilistic models''), and showed its effectivity over the use of
single type feature.  Their results suggest that feature selection is
also a crucial task in the similarity-based methods.

Cho and Kim~\shortcite{cho:nlprs-95} used ``relative entropy'', which
estimates the degree to which two probability distributions differ,
for disambiguation of verb sense. They used only one feature, in that
each example is represented simply by the object case noun for the
target polysemous verb.  The similarity between examples is computed
based on the distribution of their associated nouns.  Cho and Kim
represented the distribution of noun~$n$ by Equation~\ref{eq:cho:95},
where \mbox{$P(v_i|n)$} is the conditional probability that verb~$v_i$
appears given noun~$n$.
\begin{equation}
  \label{eq:cho:95}
  d(n) = <\!P(v_1|n),~P(v_2|n),~\ldots,~P(v_m|n)\!>
\end{equation}
Given two examples associated with distributions \mbox{$d(n_1)$} and
\mbox{$d(n_2)$}, the similarity between them is computed by
Equation~\ref{eq:relative_entropy}\footnote{In practice, Cho and
  Kim~\shortcite{cho:nlprs-95} used a smoothing technique in the case
  of an occurrence of \mbox{$P(v_i|n) = 0$} in the given
  vectors. However, we do not further describe the technique
  utilized.}.
\begin{equation}
  \label{eq:relative_entropy}
  sim(d(n_1),d(n_2)) = -\sum_{i}P(v_i|n_1)\log\frac{\textstyle
    P(v_i|n_1)}{\textstyle P(v_i|n_2)}
\end{equation}

All the methods reviewed so far compute the similarity based on the
statistical distribution obtained from training data. The remainder of
this section is devoted to reviewing methods relying on hand-crafted
resources (mostly thesauri) for the similarity computation.  Here, let
us assume that sentences~(\theemploy) (see
page~\pageref{ex:employ}) are examples incorporated into training
data, and that we have the following sentence as input.
\begin{list}{}{}
\item
  \label{ex:employ2}
  \newcounter{employ2}
  \setcounter{employ2}{\theexnum}
  \begin{tabular}{ll}
    (\exnum) & The company {\it employed\/} the graduate.
  \end{tabular}
\end{list}
One may notice that the sense of {\it employ\/} in the input would be
``to hire'', because the subject and object of {\it employ\/} in the
input are semantically closer to those in (\mbox{\theemploy}-a),
respectively, than to those in (\mbox{\theemploy}-b).  This is where
hand-crafted thesauri can be applied, based on the intuitively
feasible assumption that words located near each other within the
structure of a thesaurus have similar meaning. In practice, most
methods heuristically predefine the relation between the similarity
between nouns, and the length of the path between them in the
thesaurus structure.  Kurohashi and
Nagao~\shortcite{kurohashi:ieice-94} used the {\it Bunruigoihyo\/}
thesaurus~\cite{bgh:64} for disambiguating senses of Japanese verbs.
Uramoto~\shortcite{uramoto:ieice-94} used LDOCE~\cite{ldoce:78} for
disambiguating senses of English verbs.  Note that complements of
verbs in the training data can be considered as an extensional
description of selectional restriction.  In other words, instead of
merely ruling out inappropriate case filler nouns, example complements
estimate the degree that input complements satisfy the restriction
imposed on them. It should also be noted that lexical ambiguity of
complement nouns can be resolved in the same manner as previously
demonstrated for the disambiguation of {\it facility\/} in
(\mbox{\theemploy}-a). One may notice that {\it company\/} in the
input can be interpreted as ``installation'' because {\it facility\/}
in (\mbox{\theemploy}-a) is also related to ``installation'', and
therefore these two words are close vicinity within located in the
thesaurus\footnote{Regrettably, Kurohashi and
  Nagao~\shortcite{kurohashi:ieice-94} and
  Uramoto~\shortcite{uramoto:ieice-94} did not make significant
  comment on this point.}.  Li~\etal~\shortcite{x.li:ijcai-95} and
Lin~\shortcite{lin:acl-eacl-97} (independently) explored this notion.
In both cases, the similarity between two words is computed based on
the taxonomy defined in WordNet~\cite{miller:techrep-93}.
Li~\shortcite{x.li:ijcai-95} heuristically predefined the relation
between the length of the path in the taxonomy and the similarity.
Lin~\shortcite{lin:acl-eacl-97} used a more formal measure resembling
the information-theoretic taxonomy-based similarity measure proposed
by Resnik~\shortcite{resnik:wvlc-95,resnik:ijcai-95} (see
Section~\ref{sec:ws_related} for details).

\subsection{Restricting supervision}
\label{subsec:unsupervised_learning}

Corpus-based approaches have recently pointed the way to a promising
trend in word sense disambiguation. A number of experiments have shown
that the performance of word sense disambiguation can be significantly
improved by enhancing the volume of supervised
examples~\cite{mooney:emnlp-96,ng:acl-siglex-ws-97}. However, to do
this, supervised methods require considerable manual annotation in
supervising large-sized training data sets. To resolve this problem,
unsupervised and semi-supervised methods, which (semi-)automatically
acquire annotated training data, have been variously explored. On the
other hand, one may wonder if systems without human supervision can
perform to a reasonable level. Below, we survey different approaches
for restricting reliance on supervision. In addition, despite the
promising features of each method, we discuss problems and open
questions related to the various approach types.

\paragraph{Bilingual corpora}

Based on the observation that different word senses in a given
language can correspond to distinct words in other languages, Dagan
and Itai~\shortcite{dagan:cl-94} used bilingual corpora for word sense
disambiguation (they used Hebrew-English and German-English
language-pair corpora, respectively). In their case, word polysemy in
the source language is defined based on the existence of separate
translations in the target language. In other words, the objective of
this research can be seen as translating a source language sentence to
a target language sentence on a word-to-word basis. The method
proceeds as follows:
\begin{enumerate}
\item the source language corpus is parsed to extract syntactic tuples
  (source syntactic tuples),
\item a bilingual lexicon is used to generate alternative target
  syntactic tuples for each source syntactic tuple,
\item the target language corpus is parsed to extract target syntactic
  tuples, which are used to evaluate the plausibility of each
  alternative target tuple generated in the previous step,
\item the target sentence maximizing the combined plausibility of
  target syntactic tuples is generated.
\end{enumerate}
For example, the Hebrew tuple ``{\it higdil sikkuy\/}'' (verb-obj)
containing the polysemous word {\it higdil\/} is associated with three
English tuples: ``increase chance'', ``enlarge chance'' and ``magnify
\mbox{chance''}. However, the polysemy can be resolved by selecting
the tuple which is most likely to occur, base on the target corpus,
i.e. ``increase chance''.  It should be noted that Dagan and Itai's
method uses information extracted {\em independently\/} from the
source and target language corpora, which means that the two corpora
may not necessarily be translations of one another, and that manual
supervision is avoided. This property of potential non-correspondence
leads to a salient contrast with the method proposed by
Brown~\etal~\shortcite{brown:acl-91}, which uses bilingual {\em
  aligned\/} corpora. The applicability of bilingual corpus-based
methods is relatively limited, that is, they are restricted to
translation-oriented applications.  Apparently, word sense
distinctions differ depending on application, and therefore (as
identified by Krovets~\shortcite{krovets:acl-eacl-97}, for example)
word sense disambiguation aimed at machine translation is not
necessarily useful for information retrieval.

\paragraph{Bootstrapping}

The basis of ``bootstrapping'' is, given an initial training data set
(usually consisting of a small number of annotated examples), to
progressively enhance the training data by iteratively acquiring
presumably correctly annotated examples from previous disambiguation
results. As can be imagined, this method often ends up incorporating
noise (incorrectly annotated examples) into the training data.  To
avoid this problem, Hearst~\shortcite{hearst:oed-91} used a relatively
small number of {\em supervised\/} examples as initial training data.
Yarowsky~\shortcite{yarowsky:acl-95} completely excluded manual
supervision by automatically acquiring the initial training data set
from a dictionary. In addition, Yarowsky used discourse constraints to
exclude noise from the decision list. To put it more precisely, when
significantly large number of examples associated with a given
discourse are annotated with a common sense in the training data, all
examples associated with that discourse are standardized to the same
sense annotation.  Yarowsky's experimental results show that the
performance of this method is equivalent to that achieved by
supervised learning.

However, one controversial issue would be the applicability of
bootstrapping to the disambiguation of {\em finer-grained\/} word
sense distinction~\cite{ng:acl-96}, in that this method has generally
been applied only to {\em coarse-grained\/} binary word distinctions,
such as {\it sake\/}~(``benefit''/``Japanese liquor''). Karov and
Edelman~\shortcite{karov:wvlc-96} used bootstrapping to automatically
enhance word sense classifiers using dictionary definitions and a
corpus. As with similar research, their method also focused on
coarse-grained binary sense distinction, such as {\it
  suit\/}~(``court''/``garment''). To sum up, the effectivity of
fully-automatic bootstrapping for the disambiguation of fine-grained
sense distinction remains an open question.

\paragraph{Spreading noise through monosemous words}

Yarowsky~\shortcite{yarowsky:coling-92} projected polysemy onto word
categories defined in Roget's thesaurus~\cite{chapman:84}, and trained
statistical classifiers for {\em categories\/} rather than individual
words.  Consequently, monosemous words associated with each category
provide reliable co-occurrence statistics, and noisy statistics
arising from polysemous words should be relatively dispersed
(tolerable), without any supervision in sense tagging.  For example,
the classifier for the category \capital{ANIMAL} is expected to be
reliably trained by monosemous words like {\it sparrow}, although
polysemous words like {\it crane\/} potentially introduce a certain
proportion of noise.  Luk~\shortcite{luk:acl-95} adopted a similar
approach to obtain statistics about word senses defined in
LDOCE~\cite{ldoce:78}.  While these methods completely remove any
overhead for supervision, the applicability of the methods is limited
to word senses mapped onto thesauri, where word entries are associated
with categories. This property does not apply to many dictionaries.

\paragraph{Automatic clustering}

Sch\"{u}tze~\shortcite{schutze:supercomp-92} reduced manual
supervision by using automatic clustering algorithms (proposed by
Cheeseman~\etal~\shortcite{cheeseman:cml-88} and
Cutting~\etal~\shortcite{cutting:sigir-92}, respectively). First,
clustering algorithms are used to divide the training data into a
certain number of clusters. Thereafter, a human expert examines a
small number of examples (from 10 to 20) contained in each cluster,
which are applied in determining an appropriate word sense for each
cluster. Thus, strictly speaking, this method does not constitute
supervised learning.  Given an input, the cluster (word sense) with
the maximum similarity to the input is selected as the interpretation
(see Equation~\ref{eq:cosine_sim} for this computation). Regrettably,
since reported experiments have not compared this semi-supervised
method with fully-supervised learning methods, it is not possible to
ascertain whether automatic clustering algorithms are expected to
reduce manual supervision without degrading the performance
demonstrated by supervised methods.

Pedersen and Bruce~\shortcite{pedersen:emnlp-97} used automatic
clustering algorithms relying on McQuitty's similarity
analysis~\cite{mcquitty:epm-66} and Ward's minimum-variance
method~\cite{ward:jasa-63}, respectively. Their training/test data
includes polysemous nouns, verbs and adjectives collected from the
ACL/DCI Wall Street Journal corpus~\cite{marcus:cl-93}, in which each
word is annotated with a single sense defined in LDOCE~\cite{ldoce:78}
or WordNet~\cite{miller:techrep-93}.  Their comparative experiments
showed that automatic methods still find it difficult to match the
accuracy achievable with fully-supervised learning methods in the
disambiguation of relatively fine-grained polysemy (respective
accuracies of roughly 66\% vs. 84\%). They also tested the expectation
maximization (EM) algorithm~\cite{dempster:jrss-77} for unsupervised
learning, which resulted in an accuracy of about 63\%.

\paragraph{Linguistic behavior}

Justeson and Katz~\shortcite{justeson:cl-95} proposed a method to
automatically acquire training examples for the disambiguation of
adjective senses. As described in Section~\ref{subsec:classification},
in this case, polysemy is projected onto adjective antonyms.  Their
acquisition method uses the following three principles:
\begin{quote}
  (a) antonyms often co-occur during direct comparison, in contrastive
  opposition, \\
  (b) antonyms are frequently joined by {\it and\/} or {\it or}, \\
  (c) antonyms frequently appear in noun phrases joined by
  prepositions and with the same head noun.
\end{quote}
Each principle can be understood through the following example
sentences presented in their paper~\cite{justeson:cl-95}, which
contain the adjectives {\it old\/}~(``not new''/``not young'') and
{\it hard\/}~(``not easy''/``not soft'').
\begin{quote}
  (a) They indicated that no $\underline{{\it new\/}~{\rm errors}}$ were
  being made and that all $\underline{{\it old\/}~{\rm errors}}$ would be
  corrected within 60 days. \\
  (b) It was pitiful to see the thin ranks of warriors,
  $\underline{{\it old\/}~{\rm and}~{\it young\/}}\ldots$ \\
  (c) But there is no sudden transition $\underline{{\rm from}~{\it
      hard\/}~{\rm to}~{\it soft\/}}$.
\end{quote}
The corresponding principles annotate the instances of {\it old\/} and
{\it hard\/} in sentences (a), (b) and (c) with the senses ``not
new'', ``not young'' and ``not soft'', respectively. These training
examples are used to identify indicator nouns (see
Section~\ref{subsec:classification}).  While this method successfully
performs on these examples, the following issues must be considered.
First, the above principles can be applied only to adjective sense
disambiguation in which senses are mapped onto concepts associated
with adjective antonyms.  Second, the coverage of this method is
relatively small: Justeson and Katz used a corpus consisting of 1.5
million sentences, and could obtain only about 1500 co-occurrences of
a target adjective and antonym.  Besides this, indicator nouns
identified from these co-occurrences could disambiguate only about
27\% of (open) test inputs.

\paragraph{Machine readable dictionaries}
\label{para:mrd}

Dictionaries provide definitions (and in some cases example sentences)
for each word sense, which contain a number of ``clue words''.  Let us
take the following sentences as example definitions for different
senses of the word {\it bank\/}~(``river edge''/``financial
institution''):
\begin{list}{}{}
\item
  \label{ex:bank}
  \newcounter{bank}
  \setcounter{bank}{\theexnum}
  \begin{tabular}{lll}
    (\exnum) & a. & land along side of river/lake. (``river edge'') \\ 
    & b. & place money kept. (``financial institution'')
  \end{tabular}
\end{list}
As can be seen, these definitions contain clue words, such as {\it
  river\/} or {\it money}, which are intuitively associated with the
respective senses of {\it bank} given above.  Supposing a given input
contains {\it bank\/} and {\it river}, one can easily select the
former sense for the interpretation of the input {\it bank}.  A number
of methods follow this
intuition~\cite{cowie:coling-92,demetriou:eacl-93,guthrie:acl-91,karov:wvlc-96,lesk:acm-sigdoc-86,okumura:nlprs-95,wilks:inbook-93}.
Given an input, these methods generally compute the number of (clue)
words appearing in the input and definitions, as the score for each
sense candidate. Thereafter, they select the sense with the maximum
score. A variation of these methods is to normalize the score by the
length of the input.

Methods of this type can be categorized as instances of
similarity-based methods (see Section~\ref{subsec:classification}),
because the score represents the similarity between the input and each
sense candidate\footnote{Other methods, such as the Naive-Bayes
  method, can also use clue words as a feature set.}.  Unlike
supervised methods, these methods do not involve an excessive overhead
for large-scaled sense annotation\footnote{Strictly speaking, human
  lexicographers need to provide sense definitions, a process which is
  associated with considerable overhead. However, given that a number
  of machine readable dictionaries are currently available, word sense
  disambiguation system developers are virtually released from the
  task of establishing sense definitions.}.

However, the quality and quantity of definitions, which highly
influence performance, are problematic for these
approaches. Definitions are relatively arbitrary, and thus different
dictionaries often provide different definitions for the same
sense. This problem is less crucial for human users because they can
rely on external knowledge. However, lacking sufficient knowledge,
computers can potentially fail to perform correctly given undesirable
definitions. As for the quantity problem, it is hard to obtain
adequate statistical information because definitions usually contain
small numbers of words, or in some cases simply comprise synonym
words.  The method proposed by Luk~\shortcite{luk:acl-95}, in which an
outside corpus (the Brown corpus) was combined to enhance the
statistics, would be one solution to this problem. Co-occurrence
statistics were obtained in terms of a ``control vocabulary'' obtained
from LDOCE~\cite{ldoce:78}. However, this method requires additional
manual compilation of the control vocabulary set (pruning rarely
occurring words, for example). Okumura and
Matsunaga~\shortcite{okumura:nlprs-95} also overcame the quantity
problem by expanding definitions through the EDR
thesaurus~\cite{edr:95}. However, these general methods do not
directly address the quality problem. To sum up, we claim that the use
of dictionary definitions is not a stand-alone approach: definitions
can be used as an initial resource, but must be combined with
additional (supervised) training data. This claim was experimentally
validated, as described in Chapter~\ref{cha:vsd}.

\section{Evaluation Methodology}
\label{sec:evaluation_methodology}

From a scientific point of view, performance evaluation is invaluable.
The procedure to evaluate a dedicated method is fundamentally to
simulate run-time usage of the method by providing a corpus as
training/test data. However, the performance of corpus-based methods
is generally strongly biased by the training data provided, as well as
test data.  To minimize this bias, most experiments iteratively carry
out the same trials, and average the results derived from each trial.
In each trial, a fixed number of examples are {\em randomly\/}
selected from the corpus, as the training and test data, respectively.
This evaluation methodology can be called ``cross validation''.
Strictly speaking, the performance of methods under evaluation is
inherently biased by the given corpus. However, since collecting
broader coverage sense-annotated corpora requires tremendous human
labor, most researchers have conducted experiments using relatively
smaller-sized corpora at hand at the time.  Another problem is that
since interpretation of \mbox{``performance''} seems to vary according
to the individual, a number of separate evaluation criteria exist.
The following sections detail evaluation methodologies used in past
word sense disambiguation methods.

\subsection{Common evaluation criteria}
\label{subsec:evaluation_criteria}

One common evaluation criterion is the degree to which a method can be
applied to run-time inputs, that is, the ``coverage''\footnote{The
  term ``coverage'' can have different interpretations. One example is
  the degree to which a system restricts the type of input
  (speech/on-line text). However, we confine interpretation of
  coverage to the narrower meaning given in the text.}. This criterion
is crucial especially for qualitative approaches (see
Section~\ref{subsec:classification}), because naive rules often reject
all the sense candidates and therefore no decision can be made.
Additionally, the ratio between the number of cases where the correct
decision was made and the total number of decisions made, termed as
``accuracy'', is commonly used. These two criteria can be summarized
as in Equations~\ref{eq:coverage/accuracy}.
\begin{equation}
  \label{eq:coverage/accuracy}
  \begin{array}{rll}
    \mbox{coverage} & = & \frac{\textstyle \mbox{\# of decisions
        made}}{\textstyle \mbox{total \# of inputs}} \\
    \noalign{\vskip 2ex}
    \mbox{accuracy} & = & \frac{\textstyle \mbox{\# of correct
        decisions made}}{\textstyle \mbox{total \# of decisions made}}
  \end{array}
\end{equation}
It should be noted that these criteria can also be applied to
quantitative approaches.  Given the degree of confidence about the
decision, the method can {\em purposefully\/} refuse to make a risky
decision. A number of instances of this principle can be found in the
evaluation of quantitative word sense disambiguation methods. Dagan
and Itai~\shortcite{dagan:cl-94} estimated the degree of confidence by
the statistical significance of the given training
data~\cite{woods:86}\footnote{Dagan and Itai~\shortcite{dagan:cl-94}
  used the criteria of ``applicability'' and ``precision'', which
  correspond to ``coverage'' and ``accuracy'', respectively, in this
  section.}.  Li~\etal~\shortcite{x.li:ijcai-95} empirically defined
the relation between their heuristic rules and the confidence
degree. In these cases, coverage and accuracy are contradictory
criteria.  However, most recent experiments have commonly employed a
``backing-off'' strategy (for example, random choice) when no decision
can be made, resulting in a coverage of virtually 100\% and
performance being evaluated simply through accuracy. One plausible
reason for this approach is that a single criterion is more
intuitively understandable than multiple criteria. In fact, in the
case of Li~\etal~\shortcite{x.li:ijcai-95} mentioned above, the
trade-off between coverage and accuracy can be interpreted as the
validation of the definition of confidence degree, rather than the
performance of the method itself.

One may notice that coverage and accuracy resemble ``recall'' and
``precision'', respectively, which have been commonly used in
performance evaluation of information retrieval (IR) and text
categorization (TC) tasks.  In fact, the definition of
recall/precision in experiments carried by
Ribas~\shortcite{ribas:eacl-95}, for example, is almost the same as
that of coverage/accuracy, as given above. Let us devote a little more
space to explaining the notion of recall and precision.  In the case
of IR, recall favors systems that retrieve as many documents salient
to a user query as possible (disregarding noise contained in the
retrieved data), while precision favors systems that retrieve as few
irrelevant documents as possible.  As can be seen, when all the
documents are retrieved, recall is always 100\%, potentially
sacrificing precision. On the other hand, in the case of TC, recall
favors systems that assign as many correct categories to each document
as possible, while precision favors systems that assign as few
incorrect categories to each document as possible.  Note that since a
single document can usually be associated with multiple categories
according to different points of view, evaluation relying on a single
criterion, i.e. accuracy, is less effective.  The notion of these two
contradictory criteria can be generalized as follows. Let us assume a
situation in which a subject has to answer ``yes'' or ``no'' to $N$
given questions (correct answers are not given to the subject, of
course). The results can be classified into four cases as shown in
Table~\ref{fig:contingency_table}, which is called a ``contingency
table''. In this table, capital letters (from A to D) denote the
number of instances associated with each case. Correspondingly, recall
and precision are defined as in Equation~\ref{eq:recall/precision}.
\begin{equation}
  \label{eq:recall/precision}
  \begin{array}{rll}
    \mbox{recall} & = & \frac{\textstyle A}{\textstyle A+C} \\
    \noalign{\vskip 2ex}
    \mbox{precision} & = & \frac{\textstyle A}{\textstyle A+B}
  \end{array}
\end{equation}
Note that in the case of IR, a question corresponds to a document, and
``yes'' means that the document is retrieved.  Therefore, \mbox{A+B}
and \mbox{A+C} denote the number of retrieved documents and salient
documents, respectively. On the other hand, in the case of TC, a
question corresponds to each combination of document and
category. Thus, \mbox{A+B} and \mbox{A+C} denote the number of
categories assigned to documents and correct categories assigned to
documents, respectively. We may note, in passing, that to integrate
recall and precision, ``F-measure'', which is expressed by
Equation~\ref{eq:f-measure}, is often used as an evaluation criterion.
\begin{equation}
  \label{eq:f-measure}
  \mbox{F-measure} = \frac{\textstyle (\beta^2 +
    1)\cdot\mbox{recall}\cdot\mbox{precision}}{\textstyle
    \beta^2\cdot\mbox{precision} + \mbox{recall}}
\end{equation}
Here, $\beta$ is a parametric constant used to control the preference
between recall and precision. One may notice that as a type of
categorization task, word sense disambiguation can equally be
evaluated as performed for text categorization. However, most
researchers coarsely assign a {\em single\/} sense to each word, and
seem to prefer accuracy as the evaluation criterion.

\begin{figure}[htbp]
  \begin{center}
    \small
    \leavevmode
    \begin{tabular}{|c|c|c|} \hline
      \backslashbox{subject}{correct} & yes & no \\ \hline
      yes & A & B \\ \hline
      no & C & D \\ \hline
    \end{tabular}
  \end{center}
  \caption{Contingency table}
  \label{fig:contingency_table}
\end{figure}

Finally, let us describe several entropy-driven criteria, which have
not been used for the past evaluation of word sense disambiguation.
As Resnik and Yarowsky~\shortcite{resnik:acl-siglex-ws-97} identified,
``cross-entropy'', which has commonly been used for the evaluation of
speech recognition systems, can also be used to evaluate probabilistic
word sense disambiguation systems.  Intuitively speaking, this
criterion allocates a higher score when the system assigns a higher
probability to a correct answer.  Kononenko and
Bratko~\shortcite{kononenko:ml-91} consider the prior distribution of
candidate classes as well as the posterior distribution, that is, a
higher score is allocated when the system improves on the prior
probability for the correct answer. This criterion is expected to make
it easier to compare results derived from different test data
collections. The same motivation can be found in the evaluation
methodology proposed by Gale~\etal~\shortcite{gale:acl-92}, in which
the system performance was compared with the lower bound performance
gained through the prior distribution of word senses.

\subsection{Multiple human judgement}
\label{subsec:human_judgement}

In using any of the evaluation criterion described in
Section~\ref{subsec:evaluation_criteria}, the system decision must be
compared with the correct answer.  In most cases, human judgement
provides the ``correct'' answer. However, there are cases where even
human experts cannot correctly perform word sense disambiguation,
because of differing interpretation of word senses between
individuals~\cite{gale:acl-92,ravin:acl-90}. Consequently, system
evaluation is biased by human judgement.  One can expect to minimize
this bias through the use of multiple human experts, that is, by
validating the judgement of one human through evaluation by other
humans.  Although the associated cost is greater than for a single
human judge, there have been a number of past cases of multiple judges
being used in word sense disambiguation research\footnote{Note that
  multiple human judges have also been used in the evaluation of other
  applications, such as speech recognition systems.}.

Ravin~\shortcite{ravin:acl-90} asked multiple humans (including both
linguists and non-linguists) to disambiguate the test data which was
intended for subsequent system evaluation, and used only those inputs
where humans agreed on the answer.  Gale~\etal~\shortcite{gale:acl-92}
also used multiple human judges to estimate the upper bound on their
system's performance. By ``upper bound'', they meant human performance
on disambiguation, which can be evaluated by way of comparison with
other human judgements. To estimate the upper bound more reliably,
they conducted the same trials by iteratively changing the human under
evaluation, to average the performance of different humans.
Hatzivassiloglou and McKeown~\shortcite{hatzivassiloglou:acl-93} used
multiple human judges to {\em directly\/} score individual decisions
made by their system (although they used this methodology to evaluate
the performance of adjective clustering). Intuitively speaking, their
system was attributed a higher score when it made a correct decision
which more humans agreed on the answer to.

\subsection{Empirical comparison}
\label{subsec:empirical_comparison}

Empirical (quantitative) comparison of a proposed method with various
existing methods is a common evaluation methodology, assuming some
well-defined evaluation criterion. In most cases, proposed (new)
methods are compared with lower bound techniques, that is, in the case
of word sense disambiguation, to systematically choose the most
frequently appearing sense in the training
data~\cite{gale:acl-92}. One aspect lacking in the empirical
comparison of word sense disambiguation systems is that there is no
contest for systems, as exists for information extraction systems in
the Message Understanding Conferences (MUCs). However, a number of
extensive comparisons involving a relatively large number of different
methods can be found in past experiments. Let us devote a little space
to describe these experiments, which fundamentally used the accuracy
as a evaluation criterion (some of the below disambiguation methods
are described in Section~\ref{subsec:classification}).

In the case of word sense disambiguation for English, the ACL/DCI Wall
Street Journal corpus~\cite{marcus:cl-93}, in which each word is
annotated with a single sense defined in LDOCE~\cite{ldoce:78} or
WordNet~\cite{miller:techrep-93}, has commonly been
used. Leacock~\etal~\shortcite{leacock:arpa-hlt-93} compared the
vector space model, the Naive-Bayes method and neural networks, and
reported that the Naive-Bayes method is marginally outperformed by the
two other methods. Mooney~\shortcite{mooney:emnlp-96} reported that
the Naive-Bayes method outperformed, variously, neural networks, C4.5,
$k$-NN (\mbox{$k = 3$}), and variations of the \mbox{``FOIL''} rule
induction algorithm~\cite{quinlan:ml-90}. An experiment conducted by
Ng~\shortcite{ng:emnlp-97} showed that
\mbox{``PEBLS''}~\shortcite{cost:ml-93} and the Naive-Bayes method are
comparable in performance.  Pedersen and
Bruce~\shortcite{pedersen:emnlp-97} compared three unsupervised
methods: two automatic clustering algorithms (namely, McQuitty's
similarity analysis~\cite{mcquitty:epm-66} and Ward's minimum-variance
method~\cite{ward:jasa-63}) and the expectation maximization (EM)
algorithm~\cite{dempster:jrss-77}. In the experimental results,
McQuitty's similarity clustering outperformed the other two methods.
Pedersen and Bruce~\shortcite{pedersen:aaai-iaai-97} reported that
\mbox{C4.5} slightly outperformed all of the Naive-Bayes method,
PEBLS, CN2 and variations of their decomposable
model~\cite{bruce:acl-94,pedersen:anlp-97}.

The relative results derived for these above experiments differ
because each method involves separate parameters, which can vary
between experiments. For example, a feature set, which would be
inconsistent between most of the methods above, is highly influential
in performance.

\begin{figure}[htbp]
  \begin{center}
    \leavevmode
    \psfig{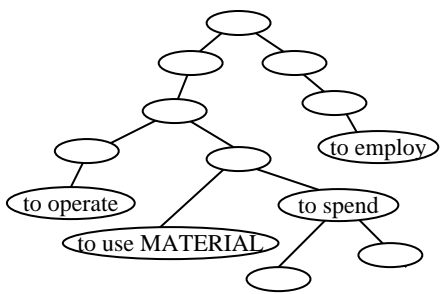}
  \end{center}
  \caption{A fragment of the EDR thesaurus including senses related
    to the Japanese verb {\it tsukau\/}}
  \label{fig:tsukau}
\end{figure}

Fujii~\etal~\shortcite{fujii:nlprs-97} compared a number of different
methods using the EDR Japanese corpus~\cite{edr:95}. According to
their results, the performance of the compared methods can be sorted
in descending order as follows: the thesaurus-driven similarity-based
method\footnote{Fujii~\etal~\shortcite{fujii:nlprs-97} used the {\it
    Bunruigoihyo\/} thesaurus~\cite{bgh:64} for the similarity
  computation.}, the vector space model, the Naive-Bayes method and
the qualitative method using selectional restriction. To derive more
general results, Fujii~\etal~\shortcite{fujii:nlprs-97} used a
different type of criterion along with the accuracy.  Instead of the
conventional binary (correct/incorrect) judgement (i.e. accuracy),
they applied an evaluation criterion based on the semantic similarity
between word sense candidates, as advocated by a number of
researchers~\cite{lin:acl-eacl-97,resnik:acl-siglex-ws-97}.  To
exemplify this notion, let us take the polysemous Japanese verb {\it
  tsukau}, which has multiple senses in EDR, such as ``to employ'',
``to operate'', ``to spend'' and ``to use \capital{MATERIAL}''. These
senses are associated with the EDR thesaurus~\cite{edr:95}.
Figure~\ref{fig:tsukau} shows a fragment of the thesaurus, in which an
oval denotes a word sense. As with most thesauri, the length of the
path between two word senses can be seen as the relative semantic
distance between them. As one can see, the verb sense ``to spend'' is
physically closer to ``to use \capital{MATERIAL}'' than to ``to
employ'' or ``to operate'', in structure. In fact, these two proximal
verb senses can be merged into one common category, that is, ``to use
up''.  Furthermore, they can be merged with ``to operate'' to form
``to use \capital{PHYSICAL OBJECT}'', as distinct from the remaining
sense of ``to employ (\capital{HUMAN}/\capital{CONCEPT})''.  Let the
correct sense of an input {\it tsukau\/} be ``to use
\capital{MATERIAL}'', and assume the system incorrectly outputs ``to
spend''. In this case, the error would be more acceptable than
outputting ``to employ'', which is totally different to the correct
interpretation of ``to use \capital{MATERIAL}''.  Therefore, we should
allot the system a scaled acceptability factor instead of a score of
zero.  Note that the binary judgment simply scores the system zero,
irrespective of the extent to which the error is acceptable given a
particular practical application.
Fujii~\etal~\shortcite{fujii:nlprs-97} formalized the scaled
acceptability factor (``acceptability'', hereafter) as in
Equation~\ref{eq:acceptability}.
\begin{equation}
  \label{eq:acceptability}
  A(x,\SS) = \left(\frac{\textstyle MAXLEN - EDR(x,\SS)}{\textstyle
    MAXLEN}\right)^\alpha
\end{equation}
Here, $x$ and $\SS$ are the system's interpretation and the correct
answer, respectively, and \mbox{$A(x,\SS)$} is the acceptability of
the given $x$ and $\SS$.  \mbox{$EDR(x,\SS)$} represents the length of
the path between $x$ and $\SS$ in the EDR thesaurus. \mbox{$MAXLEN$}
is the maximum length of the path between senses associated with an
input verb. For example, \mbox{$MAXLEN=7$} in Figure~\ref{fig:tsukau}
(the length of the path between ``to operate'' and ``to employ'').
$\alpha$ is a parametric constant, which acts to control the reduction
factor of \mbox{$A(x,\SS)$} for incorrect interpretations. With a
larger $\alpha$, \mbox{$A(x,\SS)$} becomes smaller for incorrect
interpretations, and becomes closer to the binary judgment. One can
notice that the acceptability ranges from 0 (where $x$ and $\SS$ are
the most dissimilar verb senses) to 1 ($x$ and $\SS$ are
identical). The result derived through the acceptability with
different $\alpha$ values showed the same tendency as the accuracy.

\subsection{Further considerations}
\label{subsec:further_consideration}

This section discusses several further issues which should be noted
for the evaluation of word sense disambiguation systems.

The first issue is that the granularity of word senses differs
depending on the viewpoint. This problem is closely related to the
problem that different viewpoints result in different distinctions of
word senses (as discussed in Section~\ref{subsec:human_judgement}).
Consequently, the performance of the system is likely to be degraded
when finer-grained sense definitions are used for the
evaluation. Conversely, the system would generally perform better for
applications that require coarser-grained sense definitions. The best
way to minimize this bias is to use multiple word sense definition
paradigms, with different levels of granularity.
Lin~\shortcite{lin:acl-eacl-97} used multiple granularity levels by
gradually relaxing sense distinction based on the taxonomy of
WordNet~\cite{miller:techrep-93}. To put it plainly, one can simulate
coarse-grained sense definition by regarding distinct senses dominated
by the same parent node (synset) in the taxonomy as the same sense.
The experimental results showed that Lin's method outperformed the
lower bound method as the granularity of sense distinction was
lessened.  However, at the same time, multiple evaluation criteria can
complicate the interpretation of the performance.

Second, one may notice that the most straightforward evaluation type
is ``task-based evaluation'', where the criterion is the degree to
which the system improved on the performance of another application.
However, a surprisingly small number of cases of task-based evaluation
can be found in past word sense disambiguation
research. Brown~\etal~\shortcite{brown:acl-91} improved on the
performance of French-English machine translation by introducing their
word sense disambiguation method. Fukumoto and
Suzuki~\shortcite{fukumoto:coling-96} reported that application of
word sense disambiguation improved on the performance of text
categorization\footnote{Sch\"{u}tze~\shortcite{schutze:cl-x} improved
  the quality of information retrieval through word sense {\em
    discrimination\/}, which differs from word sense disambiguation in
  that the former does not use predefined sense candidates.}.
Voorhees~\shortcite{voorhees:sigir-93} used word sense disambiguation
(with respect to WordNet senses) to enhance query terms for
information retrieval (IR), which unfortunately did not improve the
performance of a conventional IR system. One reason for this result
would be the rudimentary nature of existing word sense disambiguation
methods, especially for fine-grained sense definitions.  This poses
the question as to what degree word sense disambiguation should be
correctly performed for operational applications. Experiments
conducted by Sanderson~\shortcite{sanderson:sigir-94} aimed at an to
IR system application provide a pointer to the answer to this
question. Sanderson carried out experiments as follows.  First, due to
the lack of large-scaled sense-annotated data, artificial polysemous
words, namely
``pseudo-words''~\cite{schutze:supercomp-92,yarowsky:arpa-hlt-93} are
created, based on words appearing in a given document collection.  For
example, distinct words like {\it guerrilla\/} and {\it reptile\/} are
considered to be two distinct senses of a pseudo-word. In other words,
pseudo-words simulate extremely coarse-grained lexical
ambiguity. Based on this, one can artificially alter the performance
of (virtual) word sense disambiguation systems. For example, a system
with 50\% accuracy can be simulated by {\em knowingly\/} selecting
incorrect senses for half of the pseudo-words contained in the
document collection. The results show that (a) a theoretical perfect
WSD system improved on the baseline performance of the IR system (for
which polysemy of pseudo-words remains unresolved), and (b) WSD system
with 90\% accuracy rate did not improved on the baseline performance,
which means, at a bare minimum, more than 90\% accuracy is required to
enhance IR systems. At the same time, we note that the applicability
to practical issues of this conclusion derived from the artificial
evaluation, is arguable.

\section{Related NLP Research}
\label{sec:related_nlp_research}

Let us discuss NLP research which is expected to enhance word sense
disambiguation.

\paragraph{Morphological analysis}

As reviewed in Section~\ref{subsec:classification}, past word sense
disambiguation methods generally rely on the morphological content of
words in the input (or training data), such as the root form or
part-of-speech. In addition, in the case of agglutinative languages,
such as Chinese and Korean, lexical segmentation is also
poignant\footnote{Continuous speech inputs pose the same problem for
  non-agglutinative languages.}.  Note that numerous types of
morphological analyzers (in various languages) with reasonable
performance have been established for easy access.

\paragraph{Syntactic analysis}

Syntactic analysis can enhance the performance of word sense
disambiguation methods which rely on syntactic relations (associated
with polysemous words). This analysis is especially poignant when the
input comprises a complex sentence. Note that given the fact that full
parsing is still expensive, a number of methods for partial syntactic
analysis have also been applied to identify syntactic relations
associated with polysemous
words~\cite{cho:nlprs-95,fujii:nlprs-97,zernik:coling-90}. For
Japanese verb sense disambiguation,
Fujii~\etal~\shortcite{fujii:nlprs-97} used two simple heuristics for
extracting verb complements\footnote{In Japanese, a verb complement
  consists of noun phrase (case filler) and postposition (case marker
  suffix). See Section~\ref{sec:basis} for details.}, rather than
syntactic analysis on the whole input sentence. Note that verb
complements are useful features for verb sense disambiguation. The
heuristics utilized are given below:
\begin{itemize}
\item each complement is associated to the verb of highest proximity,
\item complements containing the genitive case marker {\it no\/} are
  not considered because they can constitute either possessive or
  nominative case markers, and are thus confusing.
\end{itemize}
They reported that the performance of verb sense disambiguation
combined with these heuristics is comparable to that combined with a
full syntactic analysis. In other words, the overhead required for
syntactic analysis can be reduced without degrading the system
performance.

\paragraph{Identification of idiomatic expressions}

Idiomatic (fixed or freezed) expressions, in which a specific word
collocation pattern stands for a certain meaning, is one obstacle to
word sense disambiguation. For example, similarity-based systems
generally fail to interpret inputs where polysemous words comprise
elements of idiomatic expressions, by semantic overgeneralization
through the use of a thesaurus.  Possible solutions would include one
proposed by Uramoto~\shortcite{uramoto:ieice-94}, in which idiomatic
expressions are described separately in the database so that the
system can control their overgeneralization. At the same time, given
the fact that there is no universal consensus as to idiomatic
expression types, automatic methods to identify idiomatic
expressions~\cite{church:cl-90,smadja:acl-90} may also prove useful.

\paragraph{Discourse analysis}

The discourse or domain dependency of each word sense, as been used in
a small proportion of past word sense disambiguation
techniques~\cite{guthrie:acl-91,nasukawa:tmi-93,yarowsky:acl-95}, can
provide strong external constraints. Given a single discourse segment,
polysemous words tend to coincide in interpretation.
Anaphora/ellipsis
resolution~\cite{hardt:cl-97,kehler:cl-97,murata:nlprs-97} is also
expected to enhance the contextual information.  Let us take the
following example, in which the pronoun {\it it\/} refers to {\it
  taxi\/} in the previous sentence:
\begin{list}{}{}
\item
  \label{ex:taxi}
  \newcounter{taxi}
  \setcounter{taxi}{\theexnum}
  \begin{tabular}{ll}
    (\exnum) & A taxi is coming. Let's take {\it it}.
  \end{tabular}
\end{list}
One may notice that correct anaphora resolution would make it easier
to disambiguate the inherent polysemy of {\it take\/} in the second
sentence.

\paragraph{Establishment of linguistic resources}

As Kilgarriff~\shortcite{kilgarriff:nlprs-97} identified, lexicography
is both a benefactor and beneficiary for word sense disambiguation.
That is, dictionaries provide sense candidates for polysemous words
and clue words for each sense.  In addition, automatic identification
of word
senses~\cite{fukumoto:coling-94,utsuro:coling-96,zernik:ijcai-89}
(perhaps combined with lexicography) is expected to avoid human bias
in sense distinction.

Establishment of thesauri is also useful for a number of
thesaurus-driven word sense disambiguation
methods~\cite{kurohashi:ieice-94,x.li:ijcai-95,lin:acl-eacl-97,ribas:eacl-95,rigau:acl-eacl-97,uramoto:ieice-94,yarowsky:coling-92}.
Conventional methods for automatic thesaurus construction have
utilized dictionary definitions to extract \capital{IS-A} or
hyper/hypo
relations~\cite{hearst:coling-92,michiels:coling-82,nakamura:coling-88}.
However, given the number of electronic thesauri currently available,
the recent trend seems to be to focus on enhancement of existing
thesauri (not development {\em from scratch\/}).  Besides this, given
the fact that most thesauri aim at general purpose applications, they
need to be adjusted for a particular
usage~\cite{basili:acl-siglex-ws-97,basili:cl-96}.
Tokunaga~\etal~\shortcite{tokunaga:acl-eacl-ws-97} and
Uramoto~\shortcite{uramoto:coling-96} (independently) extended
existing thesauri by appropriately positioning unregistered words in
the taxonomy. Hearst and Sch\"{u}tze~\shortcite{hearst:inbook-96}
proposed a method to adjust the general taxonomy defined in
WordNet~\cite{miller:techrep-93} for particular usages (although they
targetted the application toward text categorization systems).

Finally, the establishment of a sense-annotated corpus as a benchmark
collection is inevitable for the standardized evaluation of word sense
disambiguation methods~\cite{miller:arpa-hlt-94,miller:arpa-hlt-93}.

\section{Discussion}
\label{sec:past_discussion}

Pustejovsky and Boguraev~\shortcite{pustejovsky:ai-93} identified one
limitation of current word sense disambiguation methods in their
dependence on fixed numbers of predefined word senses, and that
therefore a {\em novel\/} word usage cannot be considered.  This
limitation can be especially problematic when one targets an {\em
  all-purpose\/} system, because exhaustive sense enumeration seems
practically impossible.  However, focusing on a particular domain
(sublanguage), where vocabulary size is relatively limited, this
problem is expected to be resolvable. One may argue that compilation
of word senses for different domains poses a considerable
overhead. However, automatic word sense
identification~\cite{fukumoto:coling-94,utsuro:coling-96,zernik:ijcai-89}
is expected to reduce this overhead.

In addition,
Basili~\etal~\shortcite{basili:aai-97,basili:acl-siglex-ws-97}
proposed a method to {\em tune\/} word senses taken from a general
purpose lexicon (in their case, WordNet~\cite{miller:techrep-93}) to a
specific sublanguage. They first (manually) identified some of higher
level word classes in WordNet~\cite{miller:techrep-93} as ``kernel
senses'' (25 for nouns and 15 for verbs), for each of which
representative verbs are automatically identified. Then, a sublanguage
corpus is used to train statistical classifiers for each kernel
sense\footnote{Basili~\etal~\shortcite{basili:aai-97,basili:acl-siglex-ws-97}
  used a statistical method proposed by
  Yarowsky~\cite{yarowsky:coling-92} for the classifier training.}.
Finally, statistical classifiers evaluate the membership of each
(prospective) polysemous word to the kernel senses. Consequently,
novel senses are acquired and irrelevant senses specified in the
sublanguage are discarded.

Thesaurus extending
methods~\cite{tokunaga:acl-eacl-ws-97,uramoto:coling-96} can also
potentially resolve the novel sense problem.
Tokunaga~\etal~\shortcite{tokunaga:acl-eacl-ws-97} used the ``SVMV
model''~\cite{iwayama:anlp-94} to train probabilistic classifiers of
thesaurus nodes (they use the {\it Bunruigoihyo\/}
thesaurus~\cite{bgh:64} as the core thesaurus). Each node is
represented by the co-occurrence vector associated with words
belonging to the node. Given a new word unlisted in the thesaurus
(which is also represented by a vector), classifiers compute the
probability that the word belongs to each node, and the word is
positioned in the node with maximum probability. One may notice that
this technique is equal to positioning a novel usage of a word in a
conceptual hierarchy (such as WordNet~\cite{miller:techrep-93} or
EDR~\cite{edr:95})\footnote{One may argue that, strictly speaking, the
  number of word classes (senses) defined in a thesaurus is already
  limited, that is, novel senses undefined in thesauri cannot be
  considered. The answer to this problem remains as an open
  question.}.

At the same time, we concede that this issue still needs to be further
explored, and do not pretend to draw any premature conclusions in this
research.

\section{Summary}
\label{sec:past_summary}

In this chapter, we surveyed past research associated with word sense
disambiguation. First, we elaborated on different classes of
corpus-based word sense disambiguation methods, which consists of
qualitative approaches (selectional restrictions, decision trees and
decision lists) and quantitative approaches (probabilistic models and
statistically/thesaurus-driven similarity-based methods). We
identified that one critical content is selection of useful features
(clues) for word sense disambiguation. In addition, the performance of
the similarity methods strongly depends on the way to compute
similarity between given examples. We also identified that past
methods generally use corpus, where each polysemous word is associated
with correct sense, as source knowledge.  Past experiments showed that
large-scaled corpora are required for operational systems.  In other
words, a considerable overhead is required when one tries to
supervise, i.e. manually annotate word senses to large corpora.  In
view of this problem, we second investigated how past research has
tried to minimize the overhead required for human supervision.  While
past methods reportedly successfully minimized (or completely
excluded) the overhead for supervision, we identified that
applicability of past methods seems to be limited to a certain
specified usage, such as machine translation oriented
system~\cite{dagan:cl-94}, coarse-grained word sense
distinction~\cite{karov:wvlc-96,schutze:supercomp-92,yarowsky:acl-95},
and disambiguation of adjective senses~\cite{justeson:cl-95}.  Third,
from a scientific point of view, we described evaluation methodologies
for word sense disambiguation, which turned out that so far there have
been no standardized evaluation criterion. This issue needs to be
further explored.  Fourth, we described a numerous NLP research which
is expected to enhance word sense disambiguation. Finally, we
discussed a limitation of current word sense disambiguation and a
possible solution for this problem.

In the following chapters, we will tackle some of identified problems.
Chapter~\ref{cha:vsd} describes overall design of our verb sense
disambiguation system, in which we propose a method to weigh the
degree to which each feature contributes to
disambiguation. Chapter~\ref{cha:sampling} proposes novel methods to
minimize the overhead required for our system, i.e. the overhead for
supervision on large corpora and the overhead for searching large
corpora. Finally, Chapter~\ref{cha:word_similarity} explores the
similarity computation integrating hand-crafted thesaurus and
statistical information, which is expected to enhance the
similarity-based word sense disambiguation.

\chapter{Verb Sense Disambiguation System}
\label{cha:vsd}

\begin{quote}
  \it This chapter describes a similarity-based method verb sense
  disambiguation system. The disambiguation process follows the
  nearest neighbor method.  Given an input consisting of a polysemous
  verb and its governing case complements (or cases), the system
  searches a database for the example most similar to the
  input. Thereafter, the polysemy of the verb is disambiguated by
  superimposing the sense of the verb associated with the retrieved
  example. The system uses a hand-crafted thesaurus or alternatively
  co-occurrence statistics, for the similarity computation.  The main
  interest in this chapter lies in the introduction of the notion of
  `case contribution to disambiguation' (CCD), which is a weighting
  schema for each case in similarity computation. Intuitively
  speaking, greater diversity of semantic range of case filler
  examples will lead to that case contributing more highly to verb
  sense disambiguation. We also report the results of comparative
  experiments, in which the performance of disambiguation is improved
  by considering the CCD factor.
\end{quote}

\section{Overall Design}
\label{sec:overall_design}

\begin{figure}[htbp]
  \begin{center}
    \leavevmode \psfig{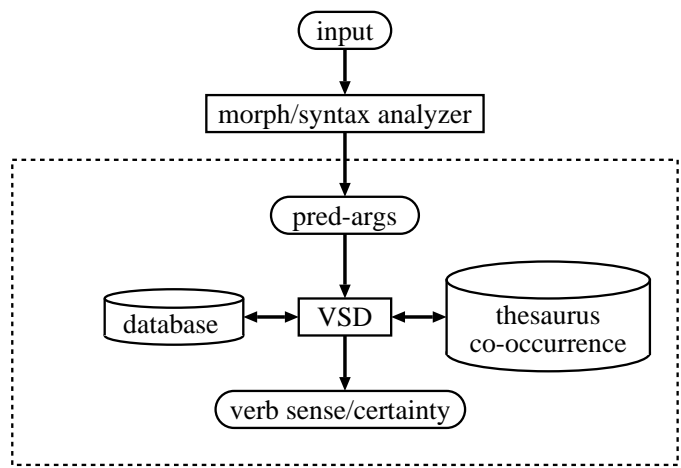}
  \end{center}
  \caption{The overall design of the verb sense disambiguation system}
  \label{fig:system}
\end{figure}

This chapter elaborates on our proposed similarity-based verb sense
disambiguation system\footnote{We originally used the term
  ``example-based''
  method~\cite{fujii:acl-eacl-ws-97,fujii:cl-x,fujii:wvlc-96,fujii:coling-96,fujii:jnlp-97}.
  However, we shall use the more general ``similarity-based''
  terminology in this research, because no explanatory distinction
  between similar approaches, such as the case-based and
  exemplar-based reasoning, has been identified.} and its evaluation
through experimentation. Although our system is currently implemented
for Japanese word sense disambiguation, the methodology can
theoretically be applied to other languages. Let us briefly explain
the disambiguation process based on Figure~\ref{fig:system}. In this
figure, ``input'' denotes a (complex/simple) sentence containing
polysemous verb(s).  Then, ``morphological and syntactic analyzer''
extracts predicate-argument structures (``pred-args'')\footnote{We
  shall interchangeably use the terms ``predicate-argument'' and
  ``verb-complement'' structures.} from the input. It should be noted
that in the case of Japanese, morphological analysis entails lexical
segmentation as well as part-of-speech tagging.  Thereafter, given the
pred-args, the core of the verb sense disambiguation system (``VSD'')
outputs plausible sense(s) for each verb in the pred-args. In this
process, the VSD uses supervised examples (in the ``database'') and
``thesaurus''/``co-occurrence'' data. The VSD additionally outputs the
degree on certainty of its decision (``certainty'').  Currently, we
rely on existing NLP tools for the morph/syntax analyzer, and this
component of the system is beyond the scope of this research. In fact,
for Japanese, a number of existing tools such as JUMAN (a
morphological analyzer)~\cite{matsumoto:93} and QJP (a morphological
and syntactic analyzer)~\cite{kameda:coling-96} have performed
promisingly.  Beside these tools, a number of simple heuristics to
identify predicate-argument structures have been proposed for verb
sense disambiguation~\cite{cho:nlprs-95,fujii:nlprs-97}, in an attempt
to minimize the overhead for syntactic analysis. To sum up, the focus
of this chapter can be identified by the portion of the overall system
enclosed within the dashed region of
Figure~\ref{fig:system}. Hereafter, ``system'' refers to the VSD, and
we shall use the term ``inputs'' to refer to the inputs which are
morphologically and syntactically analyzed (in other words,
predicate-argument structures).

Section~\ref{sec:basis} overviews the basis of our verb sense
disambiguation system, and then Section~\ref{sec:vsd_methodology}
elaborates on the disambiguation
methodology. Section~\ref{sec:vsd_experiment} evaluates our our system
by way of experiments. Section~\ref{sec:vsd_enhancement} describes a
way to further enhance our system, including the computation of the
certainty degree.

\section{Basic Idea}
\label{sec:basis}

The core mechanism in our verb sense disambiguation system is based on
the method proposed by Kurohashi and
Nagao~\shortcite{kurohashi:ieice-94} and later enhanced by
Fujii~\etaleos~\shortcite{fujii:coling-96,fujii:jnlp-97}. The system
uses an example-database (database, hereafter) containing examples of
collocations for each verb sense and its associated case frame(s). It
should be noted that while the term ``case'' has been used variously
by different researchers, including surface and deeper case level
senses, we shall consistently use ``case'' with reference to the
surface level\footnote{Conventional case systems for natural language
  are reviewed, for example, by
  Bruce~\shortcite{bruce:ai-75}.}. Figure~\ref{fig:database} shows a
fragment of the entry associated with the Japanese verb {\it
  toru}. The verb {\it toru\/} has multiple senses, a sample of which
are ``to take/steal'', ``to attain'', ``to subscribe'' and ``to
reserve''. The database specifies the case frame(s) associated with
each verb sense. In Japanese, a complement of a verb consists of a
noun phrase (case filler) and its case marker suffix, for example {\it
  ga\/} (nominative) or {\it wo\/} (accusative). The database also
lists several case filler examples for each case. In practice, the
database is usually compiled from machine readable dictionaries (MRDs)
and text corpora. The task of the system is ``to interpret'' the verbs
occurring in the input text, i.e. to choose one sense from among a set
of candidates\footnote{Note that unlike the automatic acquisition of
  word sense
  definitions~\cite{fukumoto:coling-94,pustejovsky:ai-93,utsuro:coling-96,zernik:ijcai-89},
  the task of the system is to identify the best matched category with
  a given input, from {\em predefined\/} candidates.}.

\begin{figure}[htbp]
  \begin{center}
    \footnotesize
    \leavevmode
    \begin{tabular}{|lll|} \hline
      & & \\
      \framebox{{\it toru\/}:} & & \\ & & \\ \hline
      $\left\{\begin{tabular}{ll}
      {\it suri\/} & (pickpocket) \\ {\it kanojo\/} & (she) \\
      {\it ani\/} & (brother) \end{tabular}\right\}$ {\it ga\/} &
      $\left\{\begin{tabular}{ll}
      {\it kane\/} & (money) \\ {\it saifu\/} & (wallet) \\
      {\it otoko\/} & (man) \\ {\it uma\/} & (horse) \\
      {\it aidea\/} & (idea) \end{tabular}\right\}$ {\it wo\/} &
      {\it toru\/} (to take/steal) \\ \hline
      $\left\{\begin{tabular}{ll}
      {\it kare\/} & (he) \\ {\it kanojo\/} & (she) \\
      {\it gakusei\/} & (student) \end{tabular}\right\}$ {\it ga\/} &
      $\left\{\begin{tabular}{ll}
      {\it menkyoshou\/} & (license) \\ {\it shikaku\/} & (qualification) \\
      {\it biza\/} & (visa) \end{tabular}\right\}$ {\it wo\/} &
      {\it toru\/} (to attain) \\ \hline
      $\left\{\begin{tabular}{ll}
      {\it kare\/} & (he) \\ {\it chichi\/} & (father) \\
      {\it kyaku\/} & (client) \end{tabular}\right\}$ {\it ga\/} &
      $\left\{\begin{tabular}{ll} {\it shinbun\/} & (newspaper) \\
      {\it zasshi\/} & (journal) \end{tabular}\right\}$ {\it wo\/} &
      {\it toru\/} (to subscribe) \\ \hline
      $\left\{\begin{tabular}{ll} {\it kare\/} & (he) \\
      {\it dantai\/} & (group) \\ {\it ryokoukyaku\/} & (passenger) \\
      {\it joshu\/} & (assistant) \end{tabular}\right\}$ {\it ga\/} &
      $\left\{\begin{tabular}{ll}
      {\it kippu\/} & (ticket) \\ {\it heya\/} & (room) \\
      {\it hikouki\/} & (airplane) \end{tabular}\right\}$ {\it wo\/} &
      {\it toru\/} (to reserve) \\ \hline
      {\hfill \centering $\vdots$ \hfill} & {\hfill \centering $\vdots$
        \hfill} & {\hfill \centering $\vdots$ \hfill} \\ \hline
    \end{tabular}
  \end{center}
  \caption{A fragment of the database, and the entry
    associated with the Japanese verb {\it toru\/}}
  \label{fig:database}
\end{figure}

\begin{figure}[htbp]
  \begin{center}
    \leavevmode \psfig{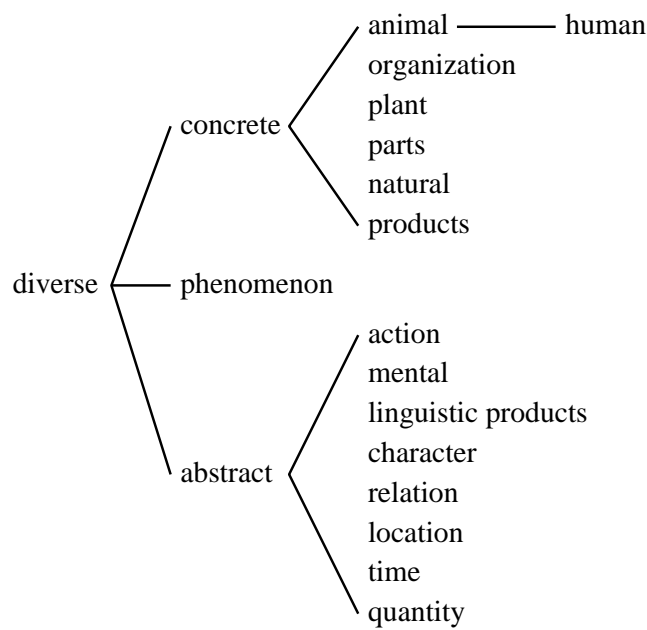}
  \end{center}
  \caption{The hierarchy of IPAL semantic features}
  \label{fig:ipal}
\end{figure}

All verb senses we use are defined in the machine readable dictionary
``IPAL''~\cite{ipa:87} (which parallels that of Kurohashi and
Nagao~\cite{kurohashi:ieice-94}). Let us devote a little space to
explaining IPAL. IPAL lists about 900 Japanese verbs and categorizes
each verb into multiple ``subentries'', based on verbal syntax and
semantics. In our case, subentries are equivalent to verb senses.
Figure~\ref{fig:IPAL_kakeru} shows a fragment of a subentry for the
Japanese verb {\it kakeru}, including the subentry number, definition
of the sense, case pattern, semantic features for case fillers, and
example case filler nouns. The ``case pattern'' entry defines the
obligatory cases associated with the verb sense, in which $N_i$
represents the $i$-th case filler\footnote{Note that IPAL describes
  only the {\em typical\/} case order, and that word order in Japanese
  is less strictly restricted than other language such as English.}.
In addition, optional cases are indicated in parentheses (such as
$N_2$-{\it ni\/} in Figure~\ref{fig:IPAL_kakeru}). IPAL defines 19
semantic feature types, which comprise a hierarchy, as shown in
Figure~\ref{fig:ipal}. In other words, these semantic features can be
used as selectional restrictions (see
Section~\ref{subsec:classification}). However, according to
experiments conducted by Kurohashi and
Nagao~\cite{kurohashi:ieice-94}, the IPAL semantic features are not
sufficiently fine-grained for rule-based verb sense disambiguation.
Kurohashi and Nagao thus implemented their similarity-based method
based on examples listed in each ``example nouns'' entry, and showed
its effectivity over the rule-based method.  Although Kurohashi and
Nagao used only examples listed in IPAL, their method is open for the
use of additional examples. In our case, we take from IPAL only the
``case pattern'' and ``example nouns'' entries to initialize the
database, and later enhance the database with additional supervised
examples.

\begin{figure}[htbp]
  \begin{center}
    \small
    \leavevmode
    \begin{tabular}{|c|c||c|} \hline
      \multicolumn{2}{|c||}{\# subentry} & 008 \\ \hline
      \multicolumn{2}{|c||}{definition} & attach something to a body
      \\ \hline
      \multicolumn{2}{|c||}{case pattern} & $N_1$-{\it
      ga\/}~~($N_2$-{\it ni\/})~~$N_3$-{\it wo\/} \\ \hline
      $N_1$ & semantic feature & human \\ \cline{2-3}
      & example nouns & {\it kare\/} (he), {\it kanojo\/} (she) \\
      \hline
      $N_2$ & semantic feature & parts \\ \cline{2-3}
      & example nouns & {\it kata\/} (shoulder), {\it te\/} (hand),
      {\it kubi\/} (neck) \\ \hline
      $N_3$ & semantic feature & products \\ \cline{2-3}
      & example nouns & {\it epuron\/} (apron), {\it megane\/}
      (glasses), {\it pendanto\/} (pendant) \\ \hline
    \end{tabular}
    \caption{An example of a subentry for the Japanese verb {\it kakeru\/}}
    \label{fig:IPAL_kakeru}
  \end{center}
\end{figure}

Let us now turn to the process of verb sense disambiguation. Given an
input, which consists of a polysemous verb and its governing
complements, the system identifies the verb sense on the basis of the
scored similarity between the input and the examples given for each
verb sense. Let us look at Figure~\ref{fig:database} again and take
the sentence below as an example input:
\begin{list}{}{\setlength{\leftmargin}{0mm}}
\item
  \label{ex:hisho}
  \newcounter{hisho}
  \setcounter{hisho}{\theexnum}
  \begin{tabular}{llll}
    (\exnum) & {\it hisho\/} {\it ga\/} & {\it shindaisha\/} {\it
      wo\/} & {\it toru}. \\ & (secretary-NOM) & (sleeping car-ACC) &
    (?)
  \end{tabular}
\end{list}
In this example, one may consider {\it hisho\/} (``secretary'') and
{\it shindaisha\/} (``sleeping car'') to be semantically similar to
{\it joshu\/} (``assistant'') and {\it hikouki\/} (``airplane'')
respectively, and since both collocate with the ``to reserve'' sense
of {\it toru}, one could infer that {\it toru\/} should be interpreted
as ``to reserve''. This resolution can be called the nearest neighbor
method because the verb in the input is disambiguated by superimposing
the sense of the verb appearing in the example of highest
similarity\footnote{Hereafter, ``similarity-based'' systems basically
  refers to those which are based on the nearest neighbor method.}. As
one can see, the similarity between an input and an example is
estimated based on the similarity between case fillers marked with the
same case. While the notion of nearest neighbor does not predefine the
type of similarity measurement used, we will explain two different
types of similarity measurement in the following sections.

Furthermore, since the restrictions imposed by the case fillers in
choosing the verb sense are not equally selective, we introduce a
weighted `case contribution to disambiguation (CCD)' of the verb
senses.  Let us consider another example input:
\begin{list}{}{\setlength{\leftmargin}{0mm}}
\item
  \label{ex:gakusei}
  \newcounter{gakusei}
  \setcounter{gakusei}{\theexnum}
  \begin{tabular}{llll}
    (\exnum) & {\it gakusei\/} {\it ga\/} & {\it shuukanshi\/} {\it
      wo\/} & {\it toru}.\\ & (student-NOM) & (magazine-ACC) & (?)
  \end{tabular}
\end{list}
The nominative, {\it gakusei\/} (``student''), in
sentence~(\thegakusei) is found in the ``to attain'' case frame of
{\it toru\/} and there is no other co-occurrence in any other sense of
{\it toru}. Therefore, the nominative supports an interpretation ``to
attain''.  On the other hand, the accusative, {\it shuukanshi\/}
(``magazine''), is most similar to the examples included in the
accusative of ``to subscribe'' and therefore the accusative supports
another interpretation ``to subscribe''. Although the most plausible
interpretation here is actually the latter, the former would be chosen
if one always relies equally on the similarity in the nominative and
the accusative.  However, in the case of {\it toru}, since the
semantic range of nouns collocating with the verb in the nominative
does not seem to have a strong delinearization in a semantic sense (in
Figure~\ref{fig:database}, the nominative of each verb sense displays
the same general concept, i.e. \capital{HUMAN}), it would be
difficult, or even risky, to properly interpret the verb sense based
on the similarity in the nominative. In contrast, since the semantic
ranges are disparate in the accusative, it would be feasible to rely
more strongly on the similarity here.

\begin{figure}[htbp]
  \begin{center}
    \leavevmode
    \leavevmode \psfig{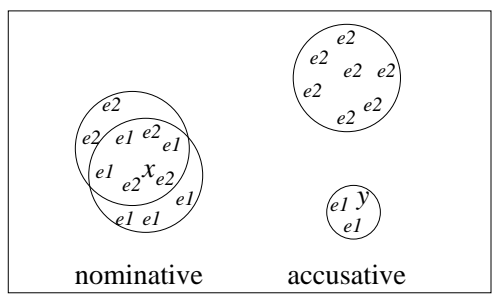}
  \end{center}
  \caption{The semantic ranges of the nominative and accusative for
    the verb {\it toru\/}}
  \label{fig:ccd}
\end{figure}

This argument can be illustrated as in Figure~\ref{fig:ccd}, in which
the symbols $e_1$ and $e_2$ denote example case fillers of different
case frames, and an input sentence includes two case fillers denoted
by $x$ and $y$.  The figure shows the distribution of example case
fillers for the respective case frames, denoted in a semantic space.
The semantic similarity between two given case fillers is represented
by the physical distance between the two symbols.  In the nominative,
since $x$ happens to be much closer to an $e_2$ than any $e_1$, $x$
may be estimated to belong to the range of $e_2$'s, although $x$
actually belongs to both sets of $e_1$'s and $e_2$'s.  In the
accusative, however, $y$ would be properly estimated to belong to the
set of $e_1$'s due to the disjunction of the two accusative case
filler sets, even though examples do not fully cover each of the
ranges of $e_1$'s and $e_2$'s. Note that this difference would be
critical if example data were sparse.  One may argue that this
property can be generalized as the notion that the system always
relies only on the similarity in the accusative for verb sense
disambiguation. Although some typical verbs show this general notion,
it is not guaranteed for any kind of verb. Our approach, which
computes the degree of contribution for each verb respectively, can
handle exceptional cases as well as typical ones. We will explain the
method used to compute CCD in Section~\ref{sec:vsd_methodology}.

\section{Methodology}
\label{sec:vsd_methodology}

To illustrate the overall algorithm, we will consider an abstract
specification of both an input and the database (see
Figure~\ref{fig:case-frame}). Let the input be
\mbox{\{$\Ni{\Ci{1}}$-$\Mi{\Ci{1}}$, $\Ni{\Ci{2}}$-$\Mi{\Ci{2}}$,
  $\Ni{\Ci{3}}$-$\Mi{\Ci{3}}$, $\V$\}}, where $\Ni{\Ci{i}}$
denotes the case filler for the case $\Ci{i}$, and
$\Mi{\Ci{i}}$ denotes the case marker for $\Ci{i}$, and assume
that the interpretation candidates for $\V$ are derived from the
database as $\SSi{1}$, $\SSi{2}$ and $\SSi{3}$. The database also
contains a set $\EXi{\SSi{i}}{\Ci{j}}$ of case filler examples
for each case $\Ci{j}$ of each sense $\SSi{i}$ (``---'' indicates that
the corresponding case is not allowed).

\begin{figure}[htbp]
  \begin{center}
    \small
    \leavevmode
    \begin{tabular}{|c|ccccl|} \hline
      input & $\Ni{\Ci{1}}$-$\Mi{\Ci{1}}$ & $\Ni{\Ci{2}}$-$\Mi{\Ci{2}}$ &
      $\Ni{\Ci{3}}$-$\Mi{\Ci{3}}$ & & $\V\:(?)$ \\ \hline \hline
      & $\EXi{\SSi{1}}{\Ci{1}}$ &
      $\EXi{\SSi{1}}{\Ci{2}}$ & $\EXi{\SSi{1}}{\Ci{3}}$ &
      --- & $\V\:(\SSi{1}$) \\
      database & $\EXi{\SSi{2}}{\Ci{1}}$ &
      $\EXi{\SSi{2}}{\Ci{2}}$ & $\EXi{\SSi{2}}{\Ci{3}}$ &
      $\EXi{\SSi{2}}{\Ci{4}}$ & $\V\:(\SSi{2}$) \\
      & --- &
      $\EXi{\SSi{3}}{\Ci{2}}$ & $\EXi{\SSi{3}}{\Ci{3}}$ &
      --- & $\V\:(\SSi{3}$) \\ \hline
    \end{tabular}
  \end{center}
  \caption{An input and the database}
  \label{fig:case-frame}
\end{figure}

During the verb sense disambiguation process, the system first
discards those candidates whose case frame does not fit the input. In
the case of Figure~\ref{fig:case-frame}, $\SSi{3}$ is discarded
because the case frame of \mbox{$\V\:(\SSi{3}$)} does not
subcategorize for the case~$\Ci{1}$\footnote{Since IPAL does not
  necessarily enumerate all the possible optional cases, the absence
  of case~$\Ci{1}$ from \mbox{$\V$~($\SSi{3}$)} in the figure may
  denote that $\Ci{1}$ is optional. If so, the interpretation
  $\SSi{3}$ should not be discarded in this stage. To avoid this
  problem, we use the same technique as used in Kurohashi's
  method. That is, we define several particular cases beforehand, such
  as the nominative, the accusative and the dative, to be obligatory,
  and impose the grammatical case frame constraints above only for
  these obligatory cases.  Optionality of case needs to be further
  explored.}.  In contrast, $\SSi{2}$ will not be rejected at this
step. This is based on the fact that in Japanese, cases can be easily
omitted if they are inferable from the given context. Note that no
omission of case fillers is allowed in the database for the reasonable
system coverage. In the case of IPAL, example case fillers cover every
slot.

In the next step the system computes the score of the remaining
candidates and chooses as the most plausible interpretation the one
with the highest score. The score of an interpretation is computed by
considering the weighted average of the similarity degrees of the
input case fillers with respect to each of the example case fillers
(in the corresponding case) listed in the database for the sense under
evaluation\footnote{$\EXi{\SSi{2}}{\Ci{4}}$ is not taken into
  consideration in the computation since $\Ci{4}$ does not appear in
  the input.}.  Formally, this is expressed by Equation~\ref{eq:score},
where \mbox{$Score(\SS)$} is the score of sense~$\SS$ of the input
verb, and \mbox{$SIM(\Ni{\C},\EXi{\SS}{\C})$} is the maximum
similarity degree between the input case filler~$\Ni{\C}$ and the
corresponding case fillers in the database example set $\EXi{\SS}{\C}$
(calculated through Equation~\ref{eq:sim}). \mbox{$CCD(\C)$} is the
weight factor of case~$\C$, which we will explain later in this
section.
\begin{equation}
\label{eq:score}
Score(\SS) = \frac{\textstyle \sum_{\C}
  SIM(\Ni{\C},\EXi{\SS}{\C})\cdot CCD(\C)}{\textstyle \sum_{\C}
  CCD(\C)}
\end{equation}
\begin{equation}
  \label{eq:sim}
  SIM(\Ni{\C},\EXi{\SS}{\C}) = {\displaystyle \max_{\E \in
      \EXi{\SS}{\C}}} sim(\Ni{\C},\E)
\end{equation}
It should be noted that while Equation~\ref{eq:score} proposes one
implementation of the score computation, there can be a number of
variations. For example, an alternative model could not consider cases
with diminished CCD values (for the verb under evaluation) in this
computation. In fact, one may notice that introduction of the CCD
factor can be characterized as one type of ``feature
selection''\footnote{See the item ``Probabilistic models''
  (p.~\pageref{para:prob}) in Section~\ref{subsec:classification} for
  past feature selection methods.}: cases associated with greater CCD
values consist of a useful feature set. In the extreme case, one can
rely solely on the case with greatest CCD
value\footnote{Yarowsky~\shortcite{yarowsky:acl-95}, for example,
  advocated a simple implementation using only the single most useful
  feature.}. Optimization of the model still remains as an open
question and needs further exploration.

The following two sections detail the computation of similarity
between case fillers, and the CCD factor.

\subsection{Computation of similarity between case filler nouns}
\label{subsec:simcomp}

\begin{figure}[htbp]
  \begin{center}
    \leavevmode \psfig{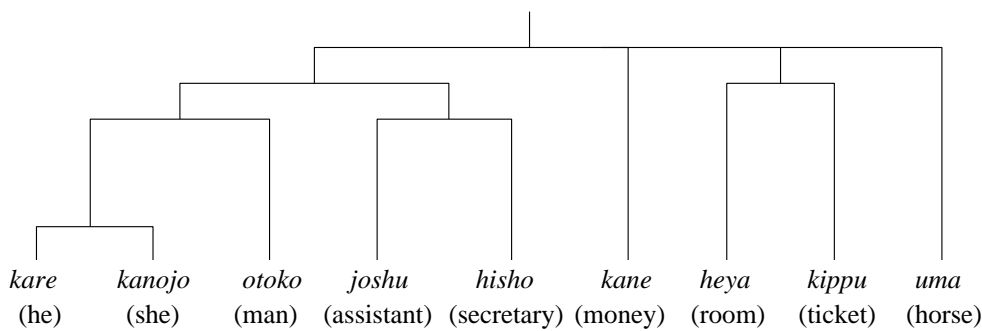}
  \end{center}
  \caption{A fragment of the {\it Bunruigoihyo\/} thesaurus}
  \label{fig:bgh}
\end{figure}

With regard to the computation of the similarity between two different
case fillers (\mbox{$sim(\Ni{\C},\E)$} in Equation~\ref{eq:score}), we
experimentally used two alternative approaches.  The first approach
uses semantic resources, that is, hand-crafted thesauri (such as
Roget's thesaurus~\cite{chapman:84} or
WordNet~\cite{miller:techrep-93} in the case of English, and {\it
  Bunruigoihyo\/}~\cite{bgh:64}, EDR~\cite{edr:95} or {\it
  Goi--Taikei\/}~\cite{ikehara:97}\footnote{{\it Goi-Taikei\/} is a
  relatively newly released Japanese dictionary.} in the case of
Japanese), based on the intuitively feasible assumption that words
located near each other within the structure of a thesaurus have
similar meaning.  Therefore, the similarity between two given words is
represented by the length of the path between them in the thesaurus
structure~\cite{fujii:coling-96,fujii:jnlp-97,kurohashi:ieice-94,x.li:ijcai-95,uramoto:ieice-94}\footnote{Different
  types of application of hand-crafted thesauri to word sense
  disambiguation have been proposed, for example, by
  Yarowsky~\shortcite{yarowsky:coling-92} (see
  Section~\ref{subsec:unsupervised_learning}).}.  We used the
similarity function empirically identified by Kurohashi and
Nagao~\shortcite{kurohashi:ieice-94}, in which the relation between
the length of the path in the {\it Bunruigoihyo\/} thesaurus and the
similarity, is defined as shown in Table~\ref{tab:sim}. In this
thesaurus, each entry is assigned a 7 digit class code.  In other
words, this thesaurus can be considered as a tree, 7 levels in depth,
with each leaf as a set of words.  Figure~\ref{fig:bgh} shows a
fragment of the {\it Bunruigoihyo\/} thesaurus including some of the
nouns in both Figure~\ref{fig:database} and the input sentence above.

\begin{table}[htbp]
  \caption{The relation between the length of the path between two nouns
    $n_1$ and $n_2$ in the {\it Bunruigoihyo\/} thesaurus
    ($len(n_1,n_2)$), and their relative similarity ($sim(n_1,n_2)$)}
  \label{tab:sim}
  \begin{center}
    \small
    \leavevmode
    \begin{tabular}{|c|ccccccc|} \hline
      $len(n_1,n_2)$ & 0 & 2 & 4 & 6 & 8 & 10 & 12 \\ \hline
      $sim(n_1,n_2)$ & 11 & 10 & 9 & 8 & 7 & 5 & 0 \\ \hline
    \end{tabular}
  \end{center}
\end{table}
  
The second approach is based on statistical modeling.  We adopted one
typical implementation called the ``vector space model''
(VSM)~\cite{frakes:92,leacock:arpa-hlt-93,niwa:coling-94,salton:83,schutze:supercomp-92},
which has a long history of application in information retrieval (IR)
and text categorization (TC) tasks.  In the case of IR/TC, VSM is used
to compute the similarity between documents, which is represented by a
vector comprising statistical factors of content words in a
document. Similarly, in our case, each noun is represented by a vector
comprising statistical factors, although statistical factors are
calculated in terms of the predicate-argument structure in which each
noun appears. Predicate-argument structures, which consist of
complements (case filler nouns and case markers) and verbs, have also
been used in the task of noun classification~\cite{hindle:acl-90}.
This can be expressed by Equation~\ref{eq:vector}, where $\vec{n}$ is
the vector for the noun in question, and items~$t_{i}$ represent the
statistics for predicate-argument structures including $n$.
\begin{equation}
  \label{eq:vector}
  \vec{n} = <t_{1},~t_{2},~\ldots,~t_{i},~\ldots>
\end{equation}
In regard to $t_{i}$, we used the notion of
TF$\cdot$IDF~\cite{salton:83}. TF (term frequency) gives each context
(a case marker/verb pair) importance proportional to the number of
times it occurs with a given noun. The rationale behind IDF (inverse
document frequency) is that contexts which rarely occur over
collections of nouns are valuable, and that therefore, the IDF of a
context is inversely proportional to the number of noun types that
appear in that context.  This notion is expressed by
Equation~\ref{eq:tf_idf}, where \mbox{$f(<\!n,c,v\!>)$} is the
frequency of the tuple \mbox{$<\!n,c,v\!>$}, \mbox{$nf(<\!c,v\!>)$} is
the number of {\em noun~types\/} which collocate with verb~$v$ in the
case~$c$, and $N$ is the number of noun types within the overall
co-occurrence data.
\begin{equation}
  \label{eq:tf_idf}
  t_{i} = f(<\!n,c,v\!>)\cdot\log\frac{\textstyle N}{\textstyle
    nf(<\!c,v\!>)}
\end{equation}
We then compute the similarity between \mbox{nouns $n_1$ and $n_2$} by
the cosine of the angle between the two \mbox{vectors $\vec{n_1}$ and
  $\vec{n_2}$}. This is realized by Equation~\ref{eq:vsm}.
\begin{equation}
  \label{eq:vsm}
  sim(n_1,n_2) = \frac{\textstyle \vec{n_1}\cdot \vec{n_2}}{\textstyle
    |\vec{n_1}||\vec{n_2}|}
\end{equation}
We extracted co-occurrence data from the RWC text base
\mbox{RWC-DB-TEXT-95-1~\cite{rwc:95}}. This text base consists of 4
years worth of \mbox{``Mainichi Shimbun''} newspaper
articles~\cite{mainichi:91-94}, which have been automatically
annotated with morphological tags. The total morpheme content is about
100 million. Since full parsing is usually expensive, a simple
heuristic rule was used in order to obtain collocations of nouns, case
markers and verbs in the form of tuples \mbox{$<\!n,c,v\!>$}. This
rule systematically associates each sequence of noun and case marker
to the verb of highest proximity, and produced 419,132 tuples. This
co-occurrence data was used in the preliminary experiment described in
Section~\ref{sec:vsd_experiment}\footnote{Note that each verb in
  co-occurrence data should ideally be annotated with its verb sense.
  However, there is no existing Japanese text base with sufficient
  volume of word sense tags.}.

\subsection{Computation of case contribution factor}
\label{subsec:ccd}

In Equation~\ref{eq:score}, \mbox{$CCD(\C)$} expresses the weight
factor of the contribution of case~$\C$ to (current) verb sense
disambiguation.  Intuitively speaking, preference should be given to
cases displaying case fillers which are classified in semantic
categories of greater disjunction.  As such, $\C$'s contribution to
the sense disambiguation of a given verb, \mbox{$CCD(\C)$}, is likely
to be higher if the example case filler sets
\mbox{\{$\EXi{\SSi{i}}{\C}~|~i = 1, \ldots, n$\}} share fewer elements
as in Equation~\ref{eq:ccd}.
\begin{equation}
  \begin{array}{l}
    CCD(\C) = {\displaystyle \left(\frac {\displaystyle
        1}{\displaystyle {\rm C}_{2}^{\it n\/}}\sum_{{\it i\/} =
        1}^{{\it n\/}-1}\sum_{{\it j\/} = {\it i\/} + 1}^{{\it
          n\/}}\frac{\displaystyle
        |\EXi{\SSi{i}}{\C}|+|\EXi{\SSi{j}}{\C}|-2|\EXi{\SSi{i}}{\C}\cap
        \EXi{\SSi{j}}{\C}|}{\displaystyle
        |\EXi{\SSi{i}}{\C}|+|\EXi{\SSi{j}}{\C}|}\right)^\alpha
      \label{eq:ccd}}
  \end{array}
\end{equation}
Here, $\alpha$ is a constant for parameterizing the extent to which
CCD influences verb sense disambiguation.  The larger $\alpha$, the
stronger CCD's influence on the system output. To avoid data
sparseness, we smooth each element (noun example) in
$\EXi{\SSi{i}}{\C}$. In practice, this involves generalizing each
example noun into a 5 digit class based on the {\it Bunruigoihyo\/}
thesaurus, as has been commonly used for smoothing.

\section{Experimentation}
\label{sec:vsd_experiment}

We estimated the performance of our verb sense disambiguation method
through a comparative experiment with other existing methods, in which
we compared the following five methods:
\begin{itemize}
\item lower bound (LB), in which the system systematically chooses the
  most frequently appearing verb sense in the
  database~\cite{gale:acl-92},
\item rule-based method (RB), in which the system uses a thesaurus to
  (automatically) identify appropriate semantic classes as selectional
  restrictions for each verb complement,
\item Naive-Bayes method (NB), in which the system interprets a given
  verb based on the probability that it takes each
  verb sense,
\item similarity-based method using the vector space model (VSM), in
  which the system uses the above mentioned co-occurrence data
  extracted from the RWC text base,
\item similarity-based method using the {\it Bunruigoihyo\/} thesaurus
  (BGH), in which the system uses Table~\ref{tab:sim} for the
  similarity computation.
\end{itemize}
Note that the last two similarity-based methods consider CCD factor
for the similarity computation.

In the rule-based method, the selectional restrictions are represented
by thesaurus classes, and allow only those nouns dominated by the
given class in the thesaurus structure as verb complements.  In order
to identify appropriate thesaurus classes, we used the association
measure proposed by Resnik~\shortcite{resnik:phd-93}, which computes
the information-theoretic association degree between case fillers and
thesaurus classes, for each verb sense\footnote{Note that previous
  research has applied this technique to tasks other than verb sense
  disambiguation, such as syntactic
  disambiguation~\cite{resnik:phd-93} and disambiguation of case
  filler noun senses~\cite{ribas:eacl-95}.}.  Equation~\ref{eq:assoc}
duplicates Equation~\ref{eq:ribas:95} here for the sake of enhanced
readability.
\begin{equation}
  \label{eq:assoc}
  A(\SS,\C,r) = P(r|\SS,\C)\cdot\log\frac{\textstyle
    P(r|\SS,\C)}{\textstyle P(r|\C)}
\end{equation}
Here, \mbox{$A(\SS,\C,r)$} is the association degree between verb
sense~$\SS$ and class~$r$ (restriction candidate) with respect to
case~$\C$. We used the semantic classes defined in the {\it
  Bunruigoihyo\/} thesaurus.  \mbox{$P(r|\SS,\C)$} is the conditional
probability that a case filler example associated with case~$\C$ of
sense~$\SS$ is dominated by class~$r$ in the {\it Bunruigoihyo\/}
thesaurus.  \mbox{$P(r|\C)$} is the conditional probability that a
case filler example for case~$\C$ (disregarding verb sense) is
dominated by class~$r$. Each probability is estimated based on the
distribution obtained from the training data. In practice, every $r$
whose association degree is above a certain threshold is chosen as a
selectional
restriction~\cite{resnik:phd-93,ribas:eacl-95}. Intuitively speaking,
by decreasing the value of the threshold, the system coverage can be
broadened while opening the way for irrelevant (noisy) selectional
restrictions.

The Naive-Bayes method, which is one type of Bayesian classification,
assumes that each feature (i.e. case filler) included in a given input
is conditionally independent of other case fillers: the system
approximates the probability that an input~$x$ takes a verb
sense~$\SS$, \mbox{$P(\SS|x)$}, simply by computing the product of the
probability that each verb sense~$\SS$ takes $\Ni{\C}$ as a case
filler for case~$\C$. The verb sense with maximal probability is then
selected as the interpretation (Equation~\ref{eq:prob})\footnote{A
number of experimental results have shown the effectivity of the
Naive-Bayes method for word sense
disambiguation~\cite{gale:ch-92,leacock:arpa-hlt-93,mooney:emnlp-96,ng:emnlp-97,pedersen:anlp-97}.}.
\begin{equation}
  \label{eq:prob}
  \begin{array}{lll}
    {\displaystyle \arg\max_{\SS} P(\SS|x)} & = & {\displaystyle
      \arg\max_{\SS} \frac{\textstyle P(\SS)\cdot P(x|\SS)}{\textstyle
        P(x)}} \\ \noalign{\vskip 1ex} & = & {\displaystyle
      \arg\max_{\SS} P(\SS)\cdot P(x|\SS)} \\ \noalign{\vskip 1ex} &
    \simeq & {\displaystyle \arg\max_{\SS} P(\SS) \prod_{\C}
      P(\Ni{\C}|\SS)}
  \end{array}
\end{equation}
Here, \mbox{$P(\Ni{\C}|\SS)$} is the probability that a case filler
associated with sense~$\SS$ for case~$\C$ in the training data is
$\Ni{\C}$. We estimated $P(\SS)$ based on the distribution of the verb
senses in the training data.  In practice, data sparseness leads to
not all case fillers~$\Ni{\C}$ appearing in the database, and as such,
we generalize each $\Ni{\C}$ into semantic class defined in the {\it
Bunruigoihyo\/} thesaurus.

A number of methods involve a parametric constant: the threshold value
for the association degree (RB), a generalization level for case
filler nouns (NB), and $\alpha$ in Equation~\ref{eq:ccd} (VSM and BGH).
For these parameters, we conducted several trials prior to the actual
comparative experiment, to determine the optimal parameter values over
a range of data sets.  For our method, we set $\alpha$ extremely
large, which is equivalent to virtually relying solely on the SIM of
the case with greatest CCD. However, note that when the SIM of the
case with greatest CCD is equal for multiple verb senses, the system
computes the SIM of the case with second highest CCD. This process is
repeated until only one verb sense remains.  When more than one verb
sense is selected for any given method (or none of them remains, for
the rule-based method), the system simply selects the verb sense that
appears most frequently in the database\footnote{One may argue that
  this goes against the basis of the rule-based method, in that, given
  a proper threshold value for the association degree, the system
  could improve on the accuracy (potentially sacrificing the
  coverage), and that the trade-off between the coverage and the
  accuracy is therefore a more appropriate evaluation criterion.
  However, our trials on the rule-based method with different
  threshold values did not show significant correlation between the
  improvement of the accuracy and the degeneration of the coverage.}.

In the experiment, we conducted six-fold cross validation, that is, we
divided the training/test data into six equal parts\footnote{Ideally
speaking, training and test data should be drawn from different
sources, to simulate a {\it real\/} application. However, the
sentences were already scrambled when provided to us, and therefore we
could not identify the original source corresponding to each
sentence.}, and conducted six trials in which a different part was
used as test data each time, and the rest as training data (the
database).  We evaluated the performance of each method according to
its accuracy, that is the ratio of the number of correct outputs,
compared to the total number of inputs.  The training/test data used
in the experiment contained about one thousand simple Japanese
sentences collected from news articles\footnote{Morph/syntax analyses
were {\em manually\/} conducted on the corpus to avoid errors
potentially caused by existing tools}.  Each sentence in the
training/test data contained one or more complement(s) followed by one
of the eleven verbs described in Table~\ref{tab:corpus}. In
Table~\ref{tab:corpus}, the column of ``English gloss'' describes
typical English translations of the Japanese verbs.  The column of
``\# of sentences'' denotes the number of sentences in the corpus, and
``\# of senses'' denotes the number of verb senses contained in
IPAL~\cite{ipa:87}. The column of ``accuracy'' shows the accuracy of
each method.

\begin{table}[htbp]
  \caption{The verbs contained in the corpus used, and the accuracy of
    the different verb sense disambiguation methods (LB: lower bound,
    RB: rule-based method, NB: Naive-Bayes method, VSM: vector space
    model, BGH: the {\it Bunruigoihyo\/} thesaurus)}
  \small
  \begin{center}
    \leavevmode
    \begin{tabular}{|c||c|c|c|c|c|c|c|c|} \hline
      & & \# of & \# of & \multicolumn{5}{c|}{accuracy (\%)}
      \\ \cline{5-9}
      verb & English gloss & sentences & senses & LB & RB & NB & VSM &
      BGH \\ \hline \hline
      {\it ataeru\/} & give & 136 & 4 &
      ~66.9~ & ~62.1~ & ~75.8~ & ~84.1~ & ~86.0~ \\ \hline
      {\it kakeru\/} & hang & 160 & 29 &
      25.6 & 24.6 & 67.6 & 73.4 & 76.2 \\ \hline
      {\it kuwaeru\/} & add & 167 & 5 &
      53.9 & 65.6 & 82.2 & 84.0 & 86.8 \\ \hline
      {\it motomeru\/} & require & 204 & 4 &
      85.3 & 82.4 & 87.0 & 85.5 & 85.5 \\ \hline
      {\it noru\/} & ride & 126 & 10 &
      45.2 & 52.8 & 81.4 & 80.5 & 85.3 \\ \hline
      {\it osameru\/} & govern & 108 & 8 &
      30.6 & 45.6 & 66.0 & 72.0 & 74.5 \\ \hline
      {\it tsukuru\/} & make & 126 & 15 &
      25.4 & 24.9 & 59.1 & 56.5 & 69.9 \\ \hline
      {\it toru\/} & take & 84 & 29 &
      26.2 & 16.2 & 56.1 & 71.2 & 75.9 \\ \hline
      {\it umu\/} & bear offspring & 90 & 2 &
      83.3 & 94.7 & 95.5 & 92.0 & 99.4 \\ \hline
      {\it wakaru\/} & understand & 60 & 5 &
      48.3 & 40.6 & 71.4 & 62.5 & 70.7 \\ \hline
      {\it yameru\/} & stop & 54 & 2 &
      59.3 & 89.9 & 92.3 & 96.2 & 96.3 \\ \hline \hline
      total & --- & 1315 & --- &
      51.4 & 54.8 & 76.6 & 78.6 & 82.3 \\ \hline
    \end{tabular}
  \end{center}
  \label{tab:corpus}
\end{table}

Looking at Table~\ref{tab:corpus}, one can see that our
similarity-based method outperformed the other methods (irrespective
of the similarity computation), although the Naive-Bayes method is
relatively comparable in performance. Surprisingly, despite the
relatively ad-hoc similarity definition utilized (see
Table~\ref{tab:sim}), the {\it Bunruigoihyo\/} thesaurus led to a
greater accuracy gain than the vector space model.  In order to
estimate the upper bound (limitation) of the disambiguation task, that
is, to what extent a human expert makes errors in
disambiguation~\cite{gale:acl-92}, we analyzed incorrect outputs and
found that roughly 30\% of the system errors using the {\it
Bunruigoihyo\/} thesaurus fell into this category.  It should be noted
that while the vector space model requires computational cost
(time/memory) of an order proportional to the size of the vector,
determination of paths in the {\it Bunruigoihyo\/} thesaurus comprises
a trivial cost\footnote{We will propose a method to optimize the
computational cost for the vector space model in
Chapter~\ref{cha:word_similarity}.}.

In order to evaluate the effectivity of the CCD factor, we also
investigated the accuracy rate for a similarity-based method which
does not consider the CCD factor (as performed by Kurohashi and
Nagao~\shortcite{kurohashi:ieice-94}). In other words, this method
computes the score for verb sense~$\SS$ simply by summing the
similarity degrees of the input case fillers with respect to each of
the example case fillers, as in Equation~\ref{eq:kurohashi}. We used
the {\it Bunruigoihyo\/} thesaurus for the similarity computation
between two case filler nouns.
\begin{equation}
  \label{eq:kurohashi}
  Score(\SS) = \sum_{\C} SIM(\Ni{\C},\EXi{\SS}{\C})
\end{equation}
We found that the CCD factor led to an accuracy gain from 76.9\% to
82.3\% (5.4\% gain).

\begin{figure}[htbp]
  \begin{center}
    \leavevmode \psfig{file=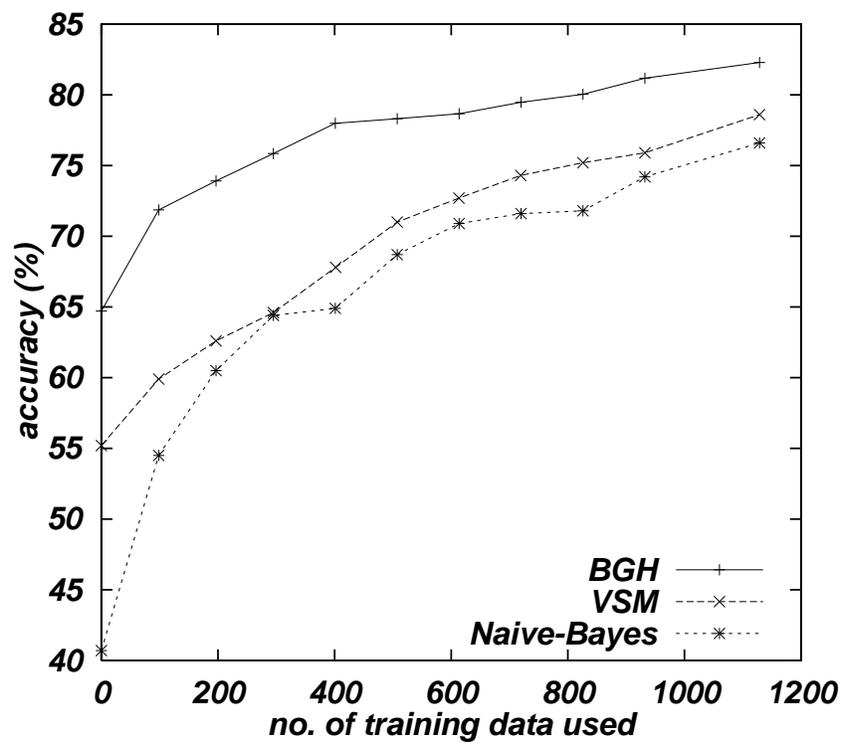,height=4in}
  \end{center}
  \caption{The relation between the number of training data used
    and accuracy of the different methods}
  \label{fig:vsd_experiment}
\end{figure}

We also investigated how the accuracy of each method improved as the
training data was increased, because the performance of corpus-based
methods generally depends on the size of training data. For this
purpose, we initially used only the examples taken from IPAL, and
progressively increased the size of the training data used.
Figure~\ref{fig:vsd_experiment} shows the the relation between the
number of the training data used and the accuracy of different
methods. In this figure, zero on the x-axis represents the system
using only the IPAL examples, which imitates a dictionary
definition-based method\footnote{See the item ``Machine readable
  dictionaries'' (p.~\pageref{para:mrd}) in
  Section~\ref{subsec:unsupervised_learning}.}. We derive from this
figure that our method using the {\it Bunruigoihyo\/} outperformed the
other methods, irrespective of the size of training data, and that
enhancing the volume of training data significantly improved on the
accuracy using only the IPAL examples. The latter observation
justifies the necessity of supervision on large-sized training data,
as shown in past
experiments~\cite{mooney:emnlp-96,ng:acl-siglex-ws-97}.

One may argue that given sufficient co-occurrence statistics, the
vector space model should outperform hand-crafted thesauri, in other
words, human lexicographers' knowledge is no longer needed.  We
investigate this prediction in Table~\ref{tab:nfreq}, which shows the
the relation between the frequency of nouns appearing in the
co-occurrence data extracted from the RWC text base (see
Section~\ref{sec:vsd_methodology}) and the accuracy of verb sense
disambiguation, in which the ``frequency'' entry denotes the threshold
of the frequency of nouns. The ``coverage'' entry denotes the ratio
between the number of inputs including at least one noun with
frequency over a given threshold, and the total number of inputs. The
last two entries show the accuracy with different similarity measures,
for each coverage. Surprisingly, not only the accuracy of VSM but also
the accuracy of the {\it Bunruigoihyo\/} thesaurus increased as the
threshold of the frequency increased, and VSM did not outperform the
{\it Bunruigoihyo\/} thesaurus for any of the thresholds. We could
assume frequently appearing nouns are used so commonly that even human
lexicographers can reasonably define the similarity between them in a
thesaurus. In addition, nouns which frequently appear in the
co-occurrence data also appear in the training data, and therefore
they provide the maximal similarity (that is, ``exact matching'')
independent of which similarity measure is used. We would like to note
that human knowledge is useful in the task of word sense
disambiguation, as with other NLP research~\cite{klavans:96}.

\begin{table}[htbp]
  \caption{The frequency of nouns and resultant accuracy of verb sense
    disambiguation}
  \begin{center}
    \small
    \leavevmode
    \begin{tabular}{|c||c|c|c|} \hline
      frequency & $\geq$100 & $\geq$1000 & $\geq$10000 \\ \hline
      coverage & 72.4\% & 57.2\% & 16.3\% \\ \hline \hline
      VSM & 83.5 & 87.9 & 91.4 \\ \hline
      BGH & 88.4 & 92.0 & 94.4 \\ \hline
    \end{tabular}
  \end{center}
  \label{tab:nfreq}
\end{table}

We also investigated errors made by the rule-based method to find a
rational explanation for its inferiority.  We identified that the
association measure in Equation~\ref{eq:assoc} tends to give a greater
value to less frequently appearing verb senses and lower level (more
specified) classes, and therefore chosen rules are generally
overspecified\footnote{This problem has also been identified by
  Charniak~\shortcite{charniak:93}.}.  Consequently, frequently
appearing verb senses are likely to be rejected. On the other hand,
when attempting to enhance the rule set by setting a smaller threshold
value for the association score, overgeneralization can be a problem.
We also note that one of the theoretical differences between the
rule-based and similarity-based methods is that the former statically
generalizes examples (prior to system usage), while the latter does so
dynamically. Static generalization would appear to be relatively risky
for sparse training data.

Although comparison of different approaches to word sense
disambiguation should be further investigated, this experimental
result gives us good motivation to explore similarity-based verb sense
disambiguation approaches in the following sections and chapters.

\section{Further Enhancement}
\label{sec:vsd_enhancement}

\subsection{Computation of interpretation certainty}
\label{subsec:certainty}

Since, as shown in Table~\ref{tab:corpus}, the system still finds it
difficult to achieve a 100\% accuracy, it is important to select
presumably correct outputs from the overall outputs (potentially
sacrificing system coverage), for practical applications. Integrated
with other NLP systems, it is desirable that other systems can vary
the degree of reliance on the output of the verb sense disambiguation
system.  To achieve this, it is useful to estimate the degree of
certainty as to the interpretation, so that one can gain higher
accuracy selecting only outputs with greater certainty degree.

A number of methods have been proposed to compute the interpretation
certainty in word sense
disambiguation~\cite{dagan:cl-94,yarowsky:acl-95} and text
categorization~\cite{lewis:sigir-94}. These methods estimate
interpretation certainty as the ratio between the probability of the
most plausible category (word sense/text category), and the
probability of any other category, excluding the most probable one.
Similarly, in the verb sense disambiguation system, we introduce the
notion of interpretation certainty based on the following preference
conditions~\cite{fujii:wvlc-96}:
\begin{enumerate}
\item the highest interpretation score is greater,
\item the difference between the highest and second highest
  interpretation scores is greater.
\end{enumerate}
The rationale for these conditions is given below. Consider
Figure~\ref{fig:certainty}, where each symbol denotes an example in a
given corpus, with symbols~$x$ as unsupervised examples and
symbols~$e$ as supervised examples.  The curved lines delimit the
semantic vicinities (extents) of the two verb senses 1 and 2,
respectively\footnote{Note that this method can easily be extended for
  a verb which has more than two senses.}. The semantic similarity
between two examples is graphically portrayed by the physical distance
between the two symbols representing them. In
Figure~\ref{fig:certainty-a}, $x$'s located inside a semantic vicinity
are expected to be interpreted as being similar to the appropriate
example~$e$ with high certainty, a fact which is in line with
condition~1 above. However, in Figure~\ref{fig:certainty-b}, the
degree of certainty for the interpretation of any $x$ which is located
inside the intersection of the two semantic vicinities cannot be
great.  This occurs when the case fillers associated with two or more
verb senses are not selective enough to allow for a clear cut
delineation between them. This situation is explicitly rejected by
condition~2.

\begin{figure}[htbp]
  \begin{center}
    \leavevmode
    \begin{minipage}[t]{.47\textwidth}
      \centering
      \psfig{file=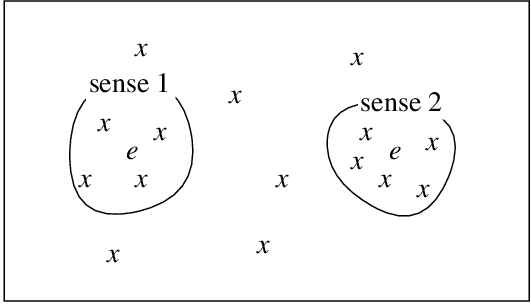,height=1.5in}
      \subcaption{The case where the interpretation certainty of the
        enclosed $x$'s is great}
      \label{fig:certainty-a}
    \end{minipage}
    \hfill
    \begin{minipage}[t]{.47\textwidth}
      \centering
      \psfig{file=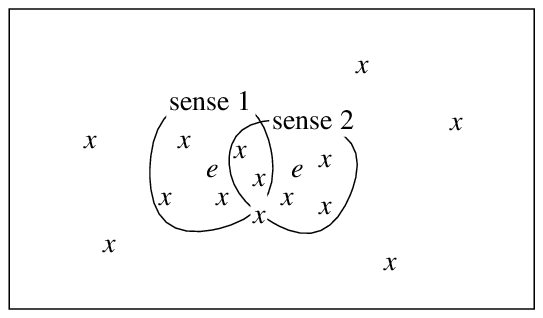,height=1.5in}
      \subcaption{The case where the interpretation certainty of the
        $x$'s contained in the intersection of senses 1 and 2 is
        small}
      \label{fig:certainty-b} \medskip
    \end{minipage}
  \end{center}
  \caption{The concept of interpretation certainty}
  \label{fig:certainty}
\end{figure}

Based on the above two conditions, we compute interpretation
certainties using Equation~\ref{eq:certainty}, where \mbox{$C(x)$} is
the interpretation certainty of an example~$x$. \mbox{$Score_1(x)$}
and \mbox{$Score_2(x)$} are the highest and second highest scores for
$x$, respectively. $\lambda$, which ranges from 0 to 1, is a
parametric constant used to control the degree to which each condition
affects the computation of \mbox{$C(x)$}.
\begin{equation}
  \label{eq:certainty}
  C(x) = \lambda\cdot Score_1(x) + (1 - \lambda)\cdot(Score_1(x) -
  Score_2(x))
\end{equation}

We estimate the validity of the notion of the interpretation certainty,
by the trade-off between the accuracy and coverage of the system. Note
that in this experiment, the accuracy is the ratio between the number
of correct outputs, and the number of cases where the interpretation
certainty of the output is above a certain threshold.  The coverage is
the ratio between the number of cases where the interpretation
certainty of the output is above a certain threshold, and the number
of inputs.  By raising the value of the threshold, the accuracy also
increases (at least theoretically), while the coverage decreases.

\begin{figure}[htbp]
  \begin{center}
    \leavevmode \psfig{file=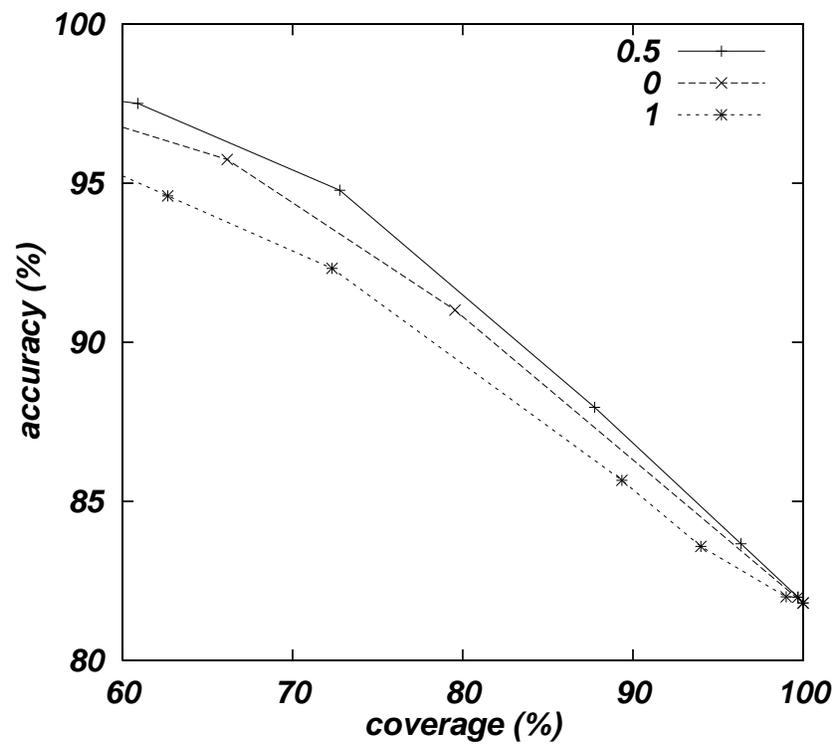,height=4in}
  \end{center}
  \caption{The relation between coverage and accuracy with different
    $\lambda$'s}
  \label{fig:cov-acc}
\end{figure}

The system used the {\it Bunruigoihyo\/} thesaurus for the similarity
computation, and was evaluated by way of six-fold cross validation
using the same corpus as that used for the experiment described in
Section~\ref{sec:vsd_experiment}.  Figure~\ref{fig:cov-acc} shows the
result of the experiment with several values of $\lambda$, from which
the optimal $\lambda$ value seems to be in the range around 0.5. It
can be seen that, as we assumed, both of the above conditions are
essential for the estimation of the interpretation certainty.

\subsection{Incorporation of contextual constraints}
\label{subsec:context}

A number of researchers have pointed out that words tend to maintain
the same sense within a given
context~\cite{nasukawa:tmi-93,yarowsky:acl-95}. In other words, when
the same verb appears multiply in the same context, we can generally
assume that it will take the same sense. The crucial issue then
becomes which verb sense to select if each verb occurrence is
interpreted with different sense by a dedicated method (for example,
the methods compared in Section~\ref{sec:vsd_experiment}). Here, we
newly introduce a method to propagate contextual constraint through
the degree of interpretation certainty. Let us assume that a given
input includes multiple distinct verbs of common lexical content, and
that one of them is interpreted with (unreasonably) smaller certainty
than the others. In the following, we list cases where the similarity
cannot be reasonably computed:
\begin{itemize}
\item insufficient supervised training data,
\item input case markers are topicalized or substituted for,
\item input cases are omitted,
\item input case filler nouns are unlisted in the
  thesaurus\footnote{Most thesauri, including WordNet and the {\it
      Bunruigoihyo\/} thesaurus, lack proper nouns and technical
    terms.} (or unlisted in the co-occurrence data in the case of the
  vector space model),
\item input case fillers comprise compound nouns.
\end{itemize}
In such cases, we superimpose interpretations with higher certainty
onto those with lower certainty. It should be noted that while the
four problems above can be individually solved through different
approaches (some of which are discussed in
Section~\ref{subsec:vsd_remaining_problems}), our approach can be used
as a {\em complementary\/} solution in a general formalism.  To
exemplify our contextual propagation method, let us take the following
example (taken from the EDR Japanese corpus~\cite{edr:95}):
\begin{list}{}{\setlength{\leftmargin}{0mm}}
\item
  \label{ex:satou}
  \newcounter{satou}
  \setcounter{satou}{\theexnum}
  \begin{tabular}{llllll}
    (\exnum) & {\it satou\/} {\it wo\/} & {\it tsuka-wazuni\/} &
    $\ldots$ & {\it parachinousu\/} {\it wo\/} & {\it tsukau}. \\ &
    (sugar-ACC) & (do not ``use \capital{MATERIAL}'') & $\ldots$ &
    (palatinose-ACC) & (?)
  \end{tabular}
\end{list}
This example contains two occurrences of the polysemous Japanese verb
{\it tsukau\/}, which has multiple senses in the EDR
dictionary~\cite{edr:95}, such as ``to employ'', ``to operate'', ``to
spend'' and ``to use \capital{MATERIAL}''. While the sense of the
former {\it tsukau\/} can be correctly identified as ``to use
\capital{MATERIAL}'', the disambiguation of the latter {\it tsukau\/}
fails because the {\it Bunruigoihyo\/} thesaurus does not list the
technical term {\it parachinousu\/}~(``palatinose''), which is a
artificial sweetener.  However, we can identify the correct
interpretation for the latter {\it tsukau\/} by superimposing the
interpretation for the former one, i.e. ``use
\capital{MATERIAL}''. Computation of the degree of certainty is
performed using the method proposed in Equation~\ref{eq:certainty}
(see Section~\ref{subsec:certainty}).

For the evaluation of this method, we used the EDR Japanese
corpus~\cite{edr:95}, which was originally extracted from news
articles.  Note that the corpus used for experiments in
Section~\ref{sec:vsd_experiment} was not applicable to this evaluation
because this corpus was compiled on a simple sentence basis. The EDR
corpus provides sense information for each word, based on the EDR
dictionary, and we used this as a means of checking the
interpretation\footnote{It should be noted that according to our
  preliminary observation, the EDR corpus contains a number of sense
  annotation errors and ``nil'' verb senses (unanalysed/unanalyseable
  verb senses). In view of this problem, we did not use this corpus
  for other experiments in this research.}. From the EDR corpus, we
first collected sentences containing one of ten frequently appearing
verbs, producing a total of 10,880 sentences. Thereafter, we collected
sentences in which a polysemous verb appears more than once, from the
aforementioned collection. We limited the range of context to one
sentence because the EDR corpus does not provide wider contextual
information, such as paragraph boundaries and sentence genres.  The
number of derived sentences was 462 (out of 10,880), which means the
applicability of this method is relatively small. However, the use of
contextual information enhanced the accuracy of the similarity-based
method using the {\it Bunruigoihyo\/} thesaurus from 60.4\% to 64.1\%.

\subsection{Remaining problems}
\label{subsec:vsd_remaining_problems}

Let us discuss how further enhancements to our verb sense
disambiguation system could be made.

First, it should be noted that in Japanese, case markers can be
omitted or topicalized (i.e. marked with postposition {\it wa\/}), an
issue which our framework does not currently consider. In addition,
polysemous verbs in relative clauses, where the surface case markers
of the head noun is omitted, pose a similar problem. Kurohashi and
Nagao~\shortcite{kurohashi:ieice-94} proposed a way of modeling such
(irregular) verb complements, by matching them to complements followed
by {\it ga}, {\it ni\/} or {\it wo\/} based on the similarity between
respective case fillers. Baldwin~\etal~\shortcite{baldwin:nlprs-97}
and Baldwin~\shortcite{baldwin:anlppgws-98} proposed a head gapping
method for Japanese relative clauses, which identifies appropriate
case slots for head nouns in relative clauses. Anaphora/ellipsis
resolution is expected to overcome the case omission problem. Given
that this processing can be carried out successfully, the similarity
between an input and examples is expected to be more reliable.

Second, our system is currently limited to the vocabulary defined in
the {\it Bunruigoihyo} thesaurus.  Thesaurus extending
methods~\cite{tokunaga:acl-eacl-ws-97,uramoto:coling-96} are expected
to counter this problem. The problem of vocabulary size is also
problematic when case filler nouns comprise compound nouns, and
therefore extraction of the semantic head is a crucial task. Note that
in Japanese, compound nouns lack lexical segmentation.  It has been
empirically shown that in more than 90\% of cases of four {\em
  kanji\/} character Japanese compound nouns\footnote{{\it Kanji\/}
  characters are basic ideograms in Japanese.}, the semantic head
consists of the last two
characters~\cite{kobayashi:coling-94}. However, analysis of longer
compound nouns still remains as a challenging task.
Kobayashi~\etal~\shortcite{kobayashi:coling-94,kobayashi:nlprs-95},
for example, proposed a method of analyzing the syntactic/semantic
structure of Japanese compound nouns, which can further enhance our
system coverage.

Third, some idiomatic expressions represent highly restricted
collocations, and overgeneralizing them semantically through the use
of a thesaurus can cause further errors. Possible solutions would
include one proposed by Uramoto, in which idiomatic expressions are
described separately in the database so that the system can control
their overgeneralization~\cite{uramoto:ieice-94} (as discussed in
Section~\ref{sec:related_nlp_research}).

Finally, external information such as the discourse or domain
dependency of each word
sense~\cite{guthrie:acl-91,nasukawa:tmi-93,yarowsky:acl-95} is
expected to lead to system improvement\footnote{Note that we limited
  the range of context to one sentence in
  Section~\ref{subsec:context}, because the EDR corpus does not
  provide wider contextual information, such as paragraph boundaries
  and sentence genres.}.

\section{Summary}
\label{sec:vsd_summary}

This chapter described our similarity-based verb sense disambiguation
system~\cite{fujii:coling-96,fujii:jnlp-97}. The basis of the system
is as follows.  First, provided with sentences containing a polysemous
verb, the system searches the database for the most similar example to
the input, following the nearest neighbor method. The database
consists of supervised examples which are manually annotated with
correct verb senses.  Thereafter, the verb is disambiguated by
superimposing the sense of the verb appearing in the most similar
supervised example. The similarity (score) between the input and an
example, or more precisely, the weighted average of the similarity
between case filler nouns included in them, is computed based on
either a statistical measure or an existing thesaurus. For the
statistical similarity measure, we use the ubiquitous vector space
model~\cite{frakes:92,leacock:arpa-hlt-93,niwa:coling-94,salton:83,schutze:supercomp-92}.
In this, each case filler noun is represented as a vector comprising
statistical factors about its collocation, which are taken from a
large-scaled text base, and the similarity between two nouns is
computed as the cosine of the angle between their associated
vectors. As for the thesaurus-driven similarity
measure~\cite{kurohashi:ieice-94,x.li:ijcai-95,uramoto:ieice-94}, we
applied a method proposed by Kurohashi and
Nagao~\cite{kurohashi:ieice-94}, which determines the similarity
between two nouns as the length of the path between them in the
hand-crafted {\it Bunruigoihyo\/} thesaurus~\cite{bgh:64}. In
practice, morphological and syntactic analyses are needed prior to the
verb sense disambiguation process.  We currently assume the use of
existing tools for morphological and syntactic analysis on the input,
and do not focus on these modules in this research.

Let us summarize the main points that have been made in this chapter.

First, we introduced the weighting factor of case contribution to
disambiguation (CCD), which computes the degree to which each case
filler contributes to verb sense disambiguation. Intuitively speaking,
greater diversity of semantic range of case filler examples will lead
to greater contribution of that case to verb sense disambiguation.  We
then compute the similarity between the input and each supervised
example, placing emphasis on cases with greater CCD values. We also
identified that the introduction of the CCD factor can be seen as a
feature selection method, that is, cases with greater CCD values
consist of a useful feature set.  In order to investigate the
effectivity of CCD, we compared our system with the rule-based and
Naive-Bayes methods. Empirical results showed that the
similarity-based system combined with the CCD factor improved on the
lower bound performance to a larger degree than the rule-based and
Naive-Bayes methods. We also showed that the use of the {\it
  Bunruigoihyo\/} thesaurus is comparable with the vector space model
in the similarity computation. While criticism has been made of
inherent limitations in hand-crafted resources, our experimental
results advocate the effectivity of these resources for NLP
applications.  In addition, we showed that additional supervised
training examples significantly improved on the performance, relying
solely on a small number of examples taken from the machine readable
dictionary IPAL~\cite{ipa:87}.

Second, in order to achieve higher accuracy, we selected presumably
correct outputs from the overall outputs by use of the notion of
interpretation certainty. The interpretation certainty is greater when
(a) the score associated with the selected verb sense is greater, and
(b) the difference between the highest and second highest scores is
greater. We showed the effectivity of this computation by way of
experiments.

Finally, our prototype method of propagating contextual constraints,
further improved on the performance of the similarity-based method.

\chapter{Selective Sampling of Effective Example Sets}
\label{cha:sampling}

\begin{quote}
  \it This chapter proposes an efficient example sampling method for
  similarity-based word sense disambiguation systems. To construct a
  database of a practical size, a considerable overhead for manual
  sense disambiguation (``overhead for supervision'') is required. In
  addition, the time complexity of searching a large-sized database
  poses a considerable problem (``overhead for search''). To counter
  these problems, our method selectively samples a smaller-sized
  effective subset from a given example set for use in word sense
  disambiguation.  Our method is characterized by its reliance on the
  notion of `training utility': the degree to which each example is
  informative for future example sampling when used for the training
  of the system. The system progressively collects examples by
  selecting those with greatest utility.  This paper reports on the
  effectivity of our method through experiments on about one thousand
  sentences. Compared to experiments with other example sampling
  methods, our method reduced both the overhead for supervision and
  the overhead for search, without degenerating the performance of the
  system.
\end{quote}

\section{Motivation}
\label{sec:sampling_motivation}

In Chapter~\ref{cha:vsd}, we described the similarity-based verb sense
disambiguation system, and showed its effectivity through comparative
experiments. Following the nearest neighbor method, our system uses
the database, which contains example sentences associated with each
verb sense. Given an input sentence containing a polysemous verb, the
system chooses the most plausible verb sense from predefined
candidates.  In this process, the system computes a scored similarity
between the input and examples in the database, and choses the verb
sense associated with the example which maximizes the score.  To
realize this, we have to manually disambiguate polysemous verbs
appearing in examples, prior to their use by the system. We shall call
these examples ``supervised examples''.  A preliminary experiment
conducted in Section~\ref{sec:vsd_experiment} on eleven polysemous
Japanese verbs showed that (a) the more supervised examples we
provided to the system, the better it performed, and (b) in order to
achieve a reasonable result (say over 80\% accuracy), the system
needed a hundred-order supervised example set for each
verb. Therefore, in order to build an operational system, the
following problems have to be taken into account\footnote{Note that
these problems are associated with corpus-based approaches in general,
and have been identified by a number of
researchers~\cite{engelson:acl-96,lewis:sigir-94,uramoto:coling-94,yarowsky:acl-95}.}:
\begin{itemize}
\item given human resource limitations, it is not reasonable to
  supervise every example in large corpora (``overhead for
  supervision''),
\item given the fact that similarity-based systems, including our system,
  search the database for the examples most similar to the input, the
  computational cost becomes prohibitive if one works with a very
  large database size (``overhead for search'').
\end{itemize}
Empirically speaking, we observed that about 84\% of the 900 or so
verbs defined in IPAL~\cite{ipa:87} are polysemous, and that some of
these verbs are associated with debilitatingly large numbers of senses
(the maximum number of senses is 32). This statistics shows that the
overhead for supervision in a real application is never trivial.  It
should be noted that the overhead for supervision is also crucial when
one tries to customize a WSD system to several distinct domains,
because (a) word sense distinctions are often different depending on
domains, and therefore (b) multiple overhead for supervision in each
domain is required.

One may argue that unsupervised WSD methods (for example, those
described in Section~\ref{subsec:unsupervised_learning}) can overcome
the former problem. To investigate this assumption, we preliminary
applied the bootstrapping method to the corpus used in
Section~\ref{sec:vsd_experiment}. In this experiment, we first
provided the similarity-based system (using the {\it Bunruigoihyo\/}
thesaurus) with an initial database, consisting of examples taken from
IPAL~\cite{ipa:87} (see the ``example nouns'' entry listed in
Figure~\ref{fig:IPAL_kakeru}).

Thereafter, the system repeats the training process and automatically
incorporates the example with maximal interpretation certainty into
the database, until no training data remains. We used
Equation~\ref{eq:certainty} for the computation of interpretation
certainty (see Section~\ref{subsec:certainty}).  Finally, we evaluated
the accuracy of verb sense disambiguation on open test data. For this
purpose, we divided the corpus into test and training data and
conducted six-fold cross validation, as carried out in
Section~\ref{sec:vsd_experiment}.  We found that the accuracy was
66.1\%, which is less than satisfactory when compared with the
accuracy of the supervised methods listed in
Table~\ref{tab:corpus}. In fact, the bootstrapping method did not
significantly improve on the accuracy rate achieved only using the
initial database (64.7\%). This experimental result advocates that we
still need supervised methods in certain cases, although we concede
that unsupervised methods also need to be explored to counter the
overhead for supervision.

\begin{figure}[htbp]
  \begin{center}
    \leavevmode \psfig{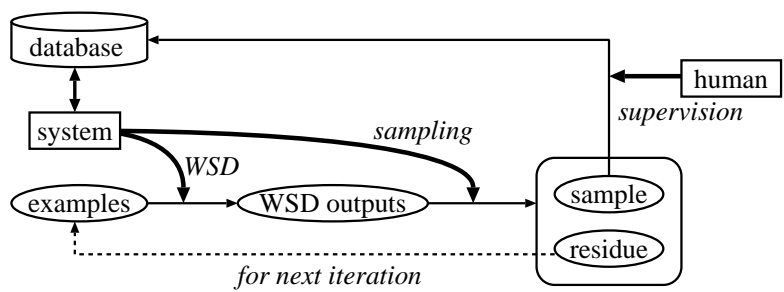}
  \end{center}
  \caption{Flow of control of the selective sampling}
  \label{fig:sampling}
\end{figure}

Motivated by these above arguments, we propose a different approach,
namely to {\em select\/} a small number of optimally informative
examples from given corpora~\cite{fujii:cl-x,fujii:wvlc-96}.
Hereafter we will call these examples ``samples''. Our example
sampling method, based on the utility maximization principle, decides
on the preference for including a given example in the database.  This
decision procedure is usually called ``selective
sampling''~\cite{cohn:ml-94}. The overall control flow of selective
sampling systems can be depicted as in Figure~\ref{fig:sampling},
where ``system'' refers to our verb sense disambiguation system, and
``examples'' refers to an unsupervised example set\footnote{The
  ``system'' in Figure~\ref{fig:sampling} includes the ``VSD'',
  ``thesaurus/co-occurrence'' and ``morph/syntax analyzer'' from
  Figure~\ref{fig:system}. However, note that the morphological and
  syntactic analyses for ``examples'' are needed only at the first
  stage of the iteration.}.  The sampling process basically cycles
between the word sense disambiguation (WSD) and training
phases. During the WSD phase, the system generates an interpretation
for each polysemous verb contained in the input example (``WSD
outputs'').  This phase is equivalent to normal word sense
disambiguation execution.  During the training phase, the system
selects samples for training from the previously produced
outputs. During this phase, a human expert supervises samples, that
is, provides the correct interpretation for the verbs appearing in the
samples. Thereafter, samples are simply incorporated into the database
without any computational overhead (as would be associated with
globally re-estimating parameters in statistics-based systems),
meaning that the system can be trained on the remaining examples
(``residue'') for the next iteration. Iterating between these two
phases, the system progressively enhances the database. It should be
noted that the selective sampling procedure gives us an optimally
informative database of a given size irrespective of the stage at
which processing is terminated.

Several researchers have proposed this type of approach for NLP
applications.  Engelson and Dagan~\shortcite{engelson:acl-96} proposed
a committee-based sampling method, which is currently applied to HMM
training for part-of-speech tagging. This method sets several models
(committee) taken from a given supervised data set, and selects
samples based on the degree of disagreement among the committee
members as to the output.  However, as this method is implemented for
statistics-based models, there is a need to explore how to formalize
and map the concept of selective sampling into similarity-based
approaches.  Lewis and Gale~\shortcite{lewis:sigir-94} proposed an
uncertainty sampling method for statistics-based text
classification. In this method, the system always samples outputs with
an uncertain level of correctness.  However, we should take into
account the training effect a given example has on other unsupervised
examples, introduced as `training utility' in our method. We devote
Section~\ref{sec:sampling_experiment} to further comparison of our
approach and other related works.

With respect to the problem of overhead for search, possible solutions
would include the generalization of similar examples through a
thesaurus taxonomy~\cite{kaji:coling-92,nomiyama:tipsj-93} or the
reconstruction of the database using a small portion of useful
instances selected from a given supervised example
set~\cite{aha:ml-91,smyth:ijcai-95}.  Aha~\etal~\shortcite{aha:ml-91}
select examples that significantly contribute to the performance gain
on test data.  However, such approaches imply a significant overhead
for supervision of each example prior to the system's execution. This
shortcoming is precisely what our approach aims to avoid: reduction of
the overhead for supervision as well as the overhead for search.

In the followings, Section~\ref{sec:sampling_methodology} first
elaborates on the methodology of our selective sampling
method. Section~\ref{sec:sampling_experiment} then evaluates our
method by way of experiments. Section~\ref{sec:related} describes
related work. Finally, discussion is added in
Section~\ref{sec:sampling_discussion}.

\section{Methodology}
\label{sec:sampling_methodology}

\subsection{Overview}
\label{subsec:overview}

Let us look again at Figure~\ref{fig:sampling}, in which ``WSD
outputs'' refers to a corpus in which each sentence is assigned an
expected verb interpretation during the WSD phase. In the training
phase, the system stores supervised samples (with each interpretation
simply checked or appropriately corrected by a human) in the database,
to be used in a later WSD phase. In this section, we turn to the
problem as to which examples should be selected as samples.

Lewis and Gale~\shortcite{lewis:sigir-94} proposed the notion of
uncertainty sampling for the training of statistics-based text
classifiers. Their method selects those examples that the system
classifies with minimum certainty, based on the assumption that there
is no need for teaching the system the correct answer when it has
answered with sufficiently high certainty. However, we should take
into account the training effect a given example has on other
remaining (unsupervised) examples. In other words, we would like to
select samples such as to be able to correctly disambiguate as many
examples as possible in the next iteration.  If this is successfully
done, the number of examples to be supervised will decrease. We
consider maximization of this effect by means of a training utility
function aimed at ensuring that the example with the greatest training
utility factor, is the most useful example at a given point in time.
Intuitively speaking, the training utility of an example is greater
when we can expect greater increase in the interpretation certainty of
the remaining examples after training using that example (see
Section~\ref{subsec:certainty} for the computation of interpretation
certainty).

To explain this notion intuitively, let us take
Figure~\ref{fig:yameru} as an example corpus.  In this corpus, all
sentences contain the verb {\it yameru}, which has two senses
according to IPAL, $s_1$ (``to stop (something)'') and $s_2$ (``to
quit (occupation)''). In this figure, sentences $e_1$ and $e_2$ are
supervised examples associated with the senses $s_1$ and $s_2$,
respectively, and $x_i$'s are unsupervised examples.  For the sake of
enhanced readability, the examples $x_i$'s are partitioned according
to their verb senses, that is, $x_1$ to $x_5$ correspond to sense
$s_1$, and $x_6$ to $x_9$ correspond to sense $s_2$.  In addition, one
may notice that examples in the corpus can be readily categorized
based on case similarity, that is, into clusters
\mbox{$\{x_1,x_2,x_3,x_4\}$} (``someone/something stops service''),
\mbox{$\{e_2,x_6,x_7\}$} (``someone leaves organization''),
\mbox{$\{x_8,x_9\}$} (``someone quits occupation''), $\{e_1\}$ and
$\{x_5\}$.  Let us simulate the sampling procedure with this example
corpus.  In the initial stage with \mbox{$\{e_1,e_2\}$} in the
database, $x_6$ and $x_7$ can be interpreted as $s_2$ with greater
certainty than for other $x_i$'s, because these two examples are
similar to $e_2$.  Therefore, uncertainty sampling selects any example
excepting $x_6$ and $x_7$ as the sample.  However, it is evident that
any of $x_1$ to $x_4$ is more desirable because by way of
incorporating one of these examples, we can obtain more $x_i$'s with
greater certainty. Assuming that $x_1$ is selected as the sample and
incorporated into the database with sense $s_1$, either of $x_8$ and
$x_9$ will be more highly desirable than other unsupervised $x_i$'s in
the next stage.

\begin{figure}[htbp]
  \begin{center}
    \leavevmode
    \small
    \begin{tabular}{|clll|} \hline
      $e_1$: & {\it seito ga\/} (student-NOM) & {\it shitsumon wo\/}
      (question-ACC) & {\it yameru\/} ($\SSi{1}$) \\
      $e_2$: & {\it ani ga\/} (brother-NOM) & {\it kaisha~wo\/}
      (company-ACC) & {\it yameru\/} ($\SSi{2}$) \\ \hline \hline
      $x_1$: & {\it shain ga\/} (employee-NOM) & {\it eigyou wo\/}
      (sales-ACC) & {\it yameru\/} (?) \\
      $x_2$: & {\it shouten ga\/} (store-NOM) & {\it eigyou wo\/}
      (sales-ACC) & {\it yameru\/} (?) \\
      $x_3$: & {\it koujou ga\/} (factory-NOM) & {\it sougyou wo\/}
      (operation-ACC) & {\it yameru\/} (?) \\
      $x_4$: & {\it shisetsu ga\/} (facility-NOM) & {\it unten wo\/}
      (operation-ACC) & {\it yameru\/} (?) \\
      $x_5$: & {\it sensyu ga\/} (athlete-NOM) & {\it renshuu wo\/}
      (practice-ACC) & {\it yameru\/} (?) \\
      $x_6$: & {\it musuko ga\/} (son-NOM) & {\it kaisha wo\/}
      (company-ACC) & {\it yameru\/} (?) \\
      $x_7$: & {\it kangofu ga\/} (nurse-NOM) & {\it byouin wo\/}
      (hospital-ACC) & {\it yameru\/} (?) \\
      $x_8$: & {\it hikoku ga\/} (defendant-NOM) & {\it giin wo\/}
      (congressman-ACC) & {\it yameru\/} (?) \\
      $x_9$: & {\it chichi ga\/} (father-NOM) & {\it kyoushi wo\/}
      (teacher-ACC) & {\it yameru\/} (?) \\ \hline
    \end{tabular}
  \end{center}
  \caption{Example of a given corpus associated with the verb {\it
      yameru\/}}
  \label{fig:yameru}
\end{figure}

Let $\set{S}$ be a set of sentences, i.e. a given corpus, and
$\set{D}$ be the subset of supervised examples stored in the database.
Further, let $\set{X}$ be the set of unsupervised examples, realizing
Equation~\ref{eq:corpus}.
\begin{equation}
  \label{eq:corpus}
  \set{S} = \set{D} \cup \set{X}
\end{equation}
The example sampling procedure can be illustrated as:
\begin{enumerate}
\item $WSD(\set{D}, \set{X})$
\item $e \leftarrow \arg\max_{x\in\set{X}}TU(x)$ 
\item $\set{D} \leftarrow \set{D} \cup\{e\},~~\set{X}
  \leftarrow \set{X}\cap \overline{\{e\}}$
\item goto 1
\end{enumerate}
where \mbox{$WSD(\set{D}, \set{X})$} is the verb sense disambiguation
process on input $\set{X}$ using $\set{D}$ as the database. In this
disambiguation process, the system outputs the following for each
input: (a) a set of verb sense candidates with interpretation scores,
and (b) an interpretation certainty. These factors are used for the
computation of \mbox{$TU(x)$}, newly introduced in our method. 
\mbox{$TU(x)$} computes the training utility factor for an example
$x$.  The sampling algorithm gives preference to examples of maximum
utility.

We will explain in the following sections how one can estimate
\mbox{$TU(x)$}, based on the estimation of the interpretation
certainty.

\subsection{Computation of training utility}
\label{subsec:utility}

\begin{figure}[htbp]
  \begin{center}
    \leavevmode
    \begin{minipage}[t]{.47\textwidth}
      \centering
      \psfig{file=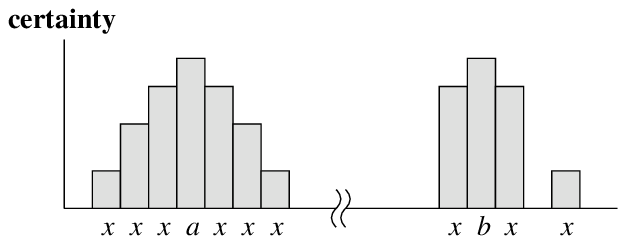,height=1in}
      \subcaption{The case where the training utility of $a$
        is greater than that of $b$ because $a$ has
        more unsupervised neighbors}
      \label{fig:tu-a}
    \end{minipage}
    \hfill
    \begin{minipage}[t]{.47\textwidth}
      \centering
      \psfig{file=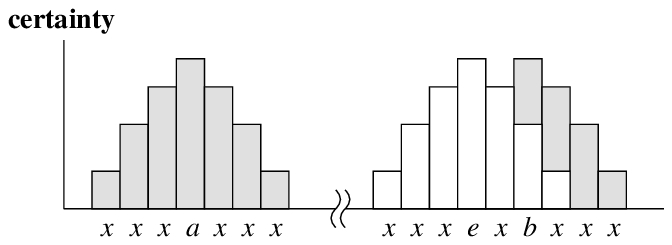,height=1in}
      \subcaption{The case where the training utility of $a$
        is greater than that of $b$ because $b$
        closely neighbors $e$, contained in the database}
      \label{fig:tu-b} \medskip
    \end{minipage}
  \end{center}
  \caption{The concept of training utility}
  \label{fig:tu}
\end{figure}

The training utility of an example $a$ is greater than that of another
example $b$ when the total interpretation certainty of unsupervised
examples increases more after training with example $a$ than with
example $b$. Let us consider Figure~\ref{fig:tu}, in which the x-axis
mono-dimensionally denotes the semantic similarity between two
unsupervised examples, and the y-axis denotes the interpretation
certainty of each example.  Let us compare the training utility of the
examples $a$ and $b$ in Figure~\ref{fig:tu-a}.  Note that in this
figure, whichever example we use for training, the interpretation
certainty for each unsupervised example ($x$) neighboring the chosen
example increases based on its similarity to the supervised example. 
Since the increase in the interpretation certainty of a given $x$
becomes smaller as the similarity to $a$ or $b$ diminishes, the
training utility of the two examples can be represented by the shaded
areas.  It is obvious that the training utility of $a$ is greater as
it has more neighbors than $b$.  On the other hand, in
Figure~\ref{fig:tu-b}, $b$ has more neighbors than $a$.  However,
since $b$ is semantically similar to $e$, which is already contained
in the database, the total {\em increase\/} in interpretation
certainty of its neighbors, i.e. the training utility of $b$, is
smaller than that of $a$.

Let \mbox{$\Delta C(x\!=\!\SS,y)$} be the difference in the
interpretation certainty of \mbox{$y\in\set{X}$} after training with
\mbox{$x\in\set{X}$}, taken with the sense $\SS$. We compute the
interpretation certainty of an example using
Equation~\ref{eq:certainty} in Section~\ref{subsec:certainty}.
\mbox{$TU(x\!=\!\SS)$}, which is the training utility function for $x$
taken with sense $\SS$, can be computed by way of
Equation~\ref{eq:utility}.
\begin{equation}
  \label{eq:utility}
  TU(x\!=\!\SS) = \sum_{y \in \set{X}}\Delta C(x\!=\!\SS,y)
\end{equation}
It should be noted that in Equation~\ref{eq:utility}, we can replace
$\set{X}$ with a subset of $\set{X}$ which consists of neighbors of
$x$. However, in order to facilitate this, an efficient algorithm to
search for neighbors of an example is required. We will discuss this
problem in Section~\ref{sec:sampling_discussion}.

Since there is no guarantee that $x$ will be supervised with any given
sense $\SS$, it can be risky to rely solely on \mbox{$TU(x\!=\!\SS)$}
for the computation of \mbox{$TU(x)$}. We estimate \mbox{$TU(x)$} by
the {\em expected\/} value of $x$, calculating the average of each
\mbox{$TU(x\!=\!\SS)$}, weighted by the probability that $x$ takes
sense $\SS$. This can be realized by Equation~\ref{eq:utility_sum},
where \mbox{$P(\SS|x)$} is the probability that $x$ takes the sense
$\SS$.
\begin{equation}
  \label{eq:utility_sum}
  TU(x) = \sum_{\SS}P(\SS|x)\cdot TU(x\!=\!\SS)
\end{equation}
Given the fact that (a) \mbox{$P(\SS|x)$} is difficult to estimate in
the current formulation, and (b) the cost of computation for each
\mbox{$TU(x\!=\!\SS)$} is not trivial, we temporarily approximate
\mbox{$TU(x)$} as in Equation~\ref{eq:utility_temp}, where $\set{K}$
is a set of the $k$-best verb sense(s) of $x$ with respect to the
interpretation score in the current state.
\begin{equation}
  \label{eq:utility_temp}
  TU(x) \simeq \frac{\textstyle 1}{\textstyle k}\sum_{\SS \in \set{K}}
  TU(x\!=\!\SS)
\end{equation}

\subsection{Enhancement of computation}
\label{subsec:computation}

In this section, we discuss how to enhance the computation associated
with our example sampling algorithm.

\begin{figure}[htbp]
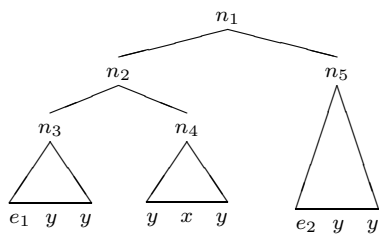

  \begin{center}
    \scriptsize
    \leavevmode
    \unitlength=.05ex
    \tree{
      \node{$n_1$}
      {\Ln7{$n_2$}
        {\Ln5{$n_3$}\tangle8{$e_1$~~$y$~~~$y$}}
        {\Rn5{$n_4$}\tangle8{$y$~~~$x$~~~$y$}}}
      {\Rn7{$n_5$}
        \tangle4{$e_2$~~$y$~~~$y$}}}
  \end{center}
  \caption{A fragment of the thesaurus including neighbors of $x$
    associated with case $\C$}
  \label{fig:neighbor}
\end{figure}

First, we note that computation of \mbox{$TU(x\!=\!\SS)$} in
Equation~\ref{eq:utility} (see Section~\ref{subsec:utility}) becomes
time consuming because the system is required to search the whole set
of unsupervised examples for examples whose interpretation certainty
will increase after $x$ is used for training.  To avoid this problem,
we could potentially apply a method used in efficient database search
techniques, by which the system can search for neighbor examples of
$x$ with optimal time complexity~\cite{utsuro:coling-94}.  However, in
this section, we will explain another efficient algorithm to identify
neighbors of $x$, in which neighbors of case fillers are taken as
being given directly by the thesaurus structure\footnote{Utsuro's
  method requires the construction of large-scaled similarity
  templates prior to similarity computation~\cite{utsuro:coling-94},
  and this is what we would like to avoid.}. The basic idea is the
following: the system searches for neighbors of each case filler of
$x$ instead of $x$ as a whole, and merges them as a set of neighbors
of $x$.  Note that by dividing examples along the lines of each case
filler, we can retrieve neighbors based on the structure of the {\it
  Bunruigoihyo\/} thesaurus (instead of the conceptual semantic space
as in Figure~\ref{fig:certainty}).  Let \mbox{$\seti{N}{x=\SS,\C}$} be
a subset of unsupervised neighbors of $x$ whose interpretation
certainty will increase after $x$ is used for training, considering
only case $\C$ of sense $\SS$. The {\em actual\/} neighbor set of $x$
with sense $\SS$ (\mbox{$\seti{N}{x=\SS}$}) is then defined as in
Equation~\ref{eq:neighbor}.
\begin{equation}
  \label{eq:neighbor}
  \seti{N}{x=\SS} = \bigcup_{\C}\seti{N}{x=\SS,\C}
\end{equation}
Figure~\ref{fig:neighbor} shows a fragment of the thesaurus, in which
$x$ and the $y$'s are unsupervised case filler examples.  Symbols
$e_1$ and $e_2$ are case filler examples stored in the database taken
as senses $\SSi{1}$ and $\SSi{2}$, respectively. The triangles
represent subtrees of the structure, and the labels $n_i$ represent
nodes.  In this figure, it can easily be seen that the interpretation
score of $\SSi{1}$ never changes for examples other than the children
of $n_4$, after $x$ is used for training with sense $\SSi{1}$. In
addition, incorporating $x$ into the database with sense $\SSi{1}$
never changes the score of examples $y$ for other sense candidates.
Therefore, $\seti{N}{x=\SSi{1},\C}$ includes only examples dominated
by $n_4$, in other words, examples which are more closely located to
$x$ than $e_1$ in the thesaurus structure.  Since, during the WSD
phase, the system determines $e_1$ as the supervised neighbor of $x$
for sense $\SSi{1}$, identifying $\seti{N}{x=\SSi{1},\C}$ does not
require any extra computational overhead. One may notice that the
technique presented here is not applicable when the vector space model
(see Section~\ref{sec:vsd_methodology}) is used for the similarity
computation. However, automatic clustering algorithms, which give a
set of words hierarchy based on the similarity between them (for
example, one proposed by Tokunaga et
al.~\shortcite{tokunaga:ijcai-95}), could potentially facilitate the
application of this retrieval method to the vector space model.

Second, sample size at each iteration should ideally be one, so as to
avoid the supervision of similar examples. On the other hand, a small
sampling size generates a considerable computation overhead for each
iteration of the sampling procedure.  This can be a critical problem
for statistics-based approaches, as the reconstruction of statistic
classifiers is expensive. However, similarity-based systems
fortunately do not require reconstruction, and examples simply have to
be stored in the database.  Furthermore, in each disambiguation phase,
our similarity-based system needs only compute the similarity between
each newly stored example and its unsupervised neighbors, rather than
between every example in the database and every unsupervised example.
Let us reconsider Figure~\ref{fig:neighbor}. As mentioned above, when
$x$ is stored in the database with sense $\SSi{1}$, only the
interpretation score of $y$'s dominated by $n_4$,
i.e. $\seti{N}{x=\SSi{1},\C}$, will be changed with respect to sense
$\SSi{1}$.  This algorithm reduces the time complexity of each
iteration from \mbox{$O(N^2)$} to \mbox{$O(N)$}, given that $N$ is the
total number of examples in a given corpus.

\section{Experimentation}
\label{sec:sampling_experiment}

In order to investigate the effectivity of our example sampling
method, we conducted an experiment, in which we compared the following
four sampling methods:
\begin{itemize}
\item a control (random), in which a certain proportion of a given
  corpus is randomly selected for training,
\item uncertainty sampling (US), in which examples with minimum
  interpretation certainty are selected~\cite{lewis:sigir-94},
\item committee-based sampling (CBS)~\cite{engelson:acl-96},
\item our method based on the notion of training utility (TU).
\end{itemize}
We elaborate on uncertainty sampling and committee-based sampling in
Section~\ref{sec:related}. We compared these sampling methods by
evaluating the relation between the number of training examples
sampled and the performance of the system.  We conducted six-fold
cross validation and carried out sampling on the training set. With
regard to the training/test data set, we used the same corpus as that
used for the experiment described in
Section~\ref{sec:vsd_experiment}. Each sampling method uses
examples from IPAL to initialize the system, with the number of
example case fillers for each case being an average of about 3.7.  For
each sampling method, the system uses the {\it Bunruigoihyo\/}
thesaurus for the similarity computation.  In Table~\ref{tab:corpus}
(in Section~\ref{sec:vsd_experiment}), the column of ``accuracy''
for ``BGH'' denotes the accuracy of the system with the entire set of
training data contained in the database. Each of the four sampling
methods achieved this figure at the conclusion of training.

\begin{figure}[htbp]
  \begin{center}
    \leavevmode
    \psfig{file=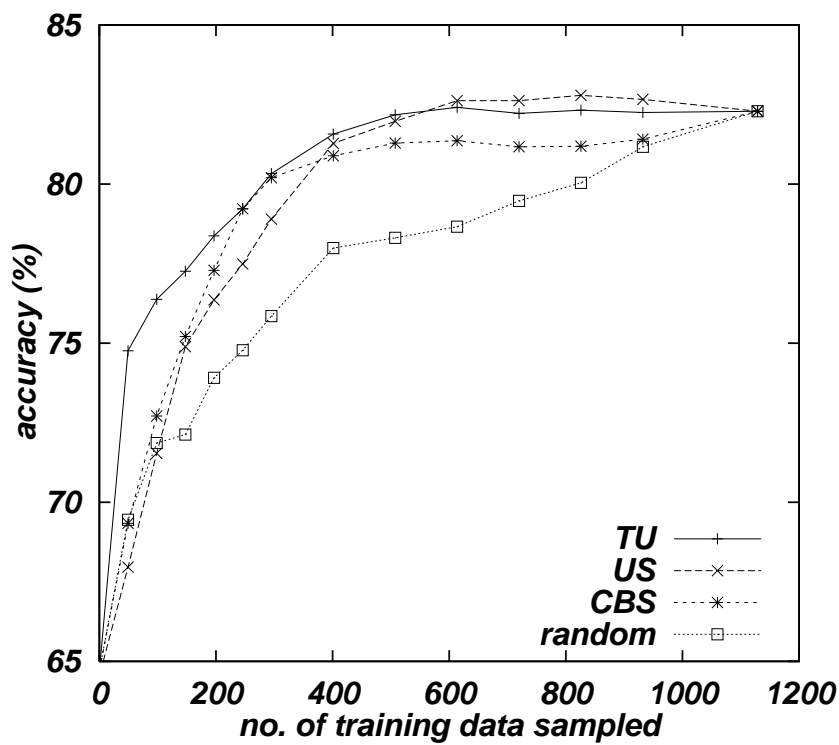,height=4in}
  \end{center}
  \caption{The relation between the number of training data sampled
    and accuracy of the system}
  \label{fig:accuracy_IPAL}
\end{figure}

We evaluated each system performance according to its accuracy, that
is the ratio of the number of correct outputs, compared to the total
number of inputs.  For the purpose of this experiment, we set the
sample size to 1 for each iteration, \mbox{$\lambda = 0.5$} for
Equation~\ref{eq:certainty}, and \mbox{$k = 1$} for
Equation~\ref{eq:utility_temp}. Based on a preliminary experiment,
increasing the value of $k$ either did not improve the performance
over that for \mbox{$k = 1$}, or lowered the overall performance.
Figure~\ref{fig:accuracy_IPAL} shows the relation between the number
of the training data sampled and the accuracy of the system. In
Figure~\ref{fig:accuracy_IPAL}, zero on the x-axis represents the
system using only the examples provided by IPAL.  Looking at
Figure~\ref{fig:accuracy_IPAL} one can see that compared with random
sampling and committee-based sampling, our sampling method reduced the
number of the training data required to achieve any given accuracy. 
For example, to achieve a accuracy of 80\%, the number of the training
data required for our method was roughly one-third of that for random
sampling.  Although the accuracy for our method was surpassed by that
for uncertainty sampling for larger sizes of training data, this
minimal difference for larger data sizes is overshadowed by the
considerable performance gain attained by our method for smaller data
sizes.

\begin{figure}[htbp]
  \begin{center}
    \leavevmode
    \psfig{file=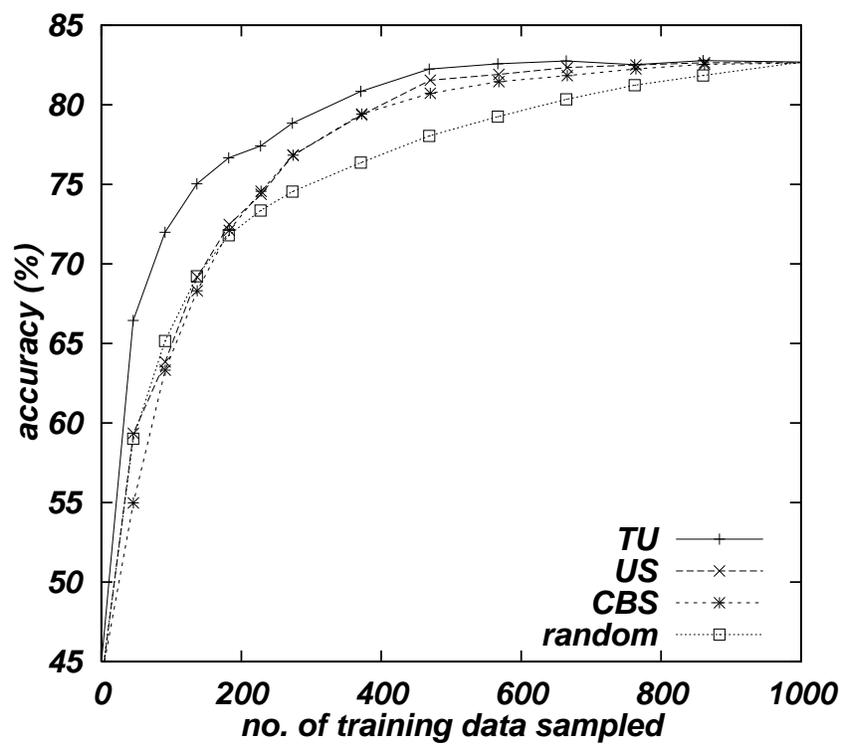,height=4in}
  \end{center}
  \caption{The relation between the number of training data sampled
    and accuracy of the system without using examples from IPAL}
  \label{fig:accuracy}
\end{figure}

Since IPAL has, in a sense, been manually selectively sampled in an
attempt to model the maximum verb sense coverage, the performance of
each method is biased by the initial contents of the database. To
counter this effect, we also conducted an experiment involving the
construction of the database from scratch, without using examples from
IPAL. During the initial phase, the system randomly selected one
example for each verb sense from the training set, and a human expert
provided the correct interpretation to initialize the
system\footnote{In order to minimize the potential bias caused by
  selection of the initial examples, we conducted the same trials with
  three different initial example sets and averaged the results for
  each trial.}. Figure~\ref{fig:accuracy} shows the performance of the
various methods, from which the same general tendency as seen in
Figure~\ref{fig:accuracy_IPAL} is observable.  However, in this case,
our method was generally superior to other methods.  Through these
comparative experiments, we can conclude that our example sampling
method is able to decrease the number of the training data, i.e.  the
overhead for both supervision and searching, without degrading the
system performance.

\section{Related Work}
\label{sec:related}

\subsection{Uncertainty sampling}
\label{subsec:lewis}

The procedure for uncertainty sampling~\cite{lewis:sigir-94} is as
follows, where \mbox{$C(x)$} represents the interpretation certainty
for an example $x$ (see our sampling procedure in
Section~\ref{subsec:overview} for comparison):
\begin{enumerate}
\item $WSD(\set{D}, \set{X})$
\item $e \leftarrow \arg\min_{x\in\set{X}}C(x)$ 
\item $\set{D} \leftarrow \set{D} \cup\{e\},~~\set{X}
  \leftarrow \set{X}\cap \overline{\{e\}}$
\item goto 1
\end{enumerate}

Let us discuss the theoretical difference between this and our method.
Considering Figure~\ref{fig:tu} again, one can see that the concept of
training utility is supported by the following properties:
\renewcommand{\theenumi}{\alph{enumi}}
\def\labelenumi{(\theenumi)}
\begin{enumerate}
\item an example which neighbors more unsupervised examples is more
  informative (Figure~\ref{fig:tu-a}),
\item an example less similar to one already existing in the database
  is more informative (Figure~\ref{fig:tu-b}).
\end{enumerate}
\renewcommand{\theenumi}{\arabic{enumi}} \def\labelenumi{\theenumi.}
Uncertainty sampling directly addresses property~(b), but ignores
property~(a). It differs from our method more crucially when more
unsupervised examples remain, because these unsupervised examples have
a greater influence on the computation of training utility.  This
assumption can also be seen in the comparative experiments in
Section~\ref{sec:sampling_experiment}, in which our method
outperformed uncertainty sampling to the highest degree in early
stages.

\subsection{Committee-based sampling}
\label{subsec:engelson}

In committee-based sampling~\cite{engelson:acl-96}, which follows the
``query by committee'' principle~\cite{seung:acm-ws-92}, the system
selects samples based on the degree of disagreement between models
randomly taken from a given training set (these models are called
``committee members'').  This is achieved by iteratively repeating the
steps given below, in which the number of committee members is given
as two without loss of generality:
\begin{enumerate}
\item draw two models randomly,
\item classify unsupervised example $x$ according to each model,
  producing classifications $C_1$ and $C_2$,
\item if $C_1 \neq C_2$ (the committee members disagree), select $x$
  for the training of the system.
\end{enumerate}

\begin{figure}[htbp]
  \begin{center}
    \leavevmode
    \psfig{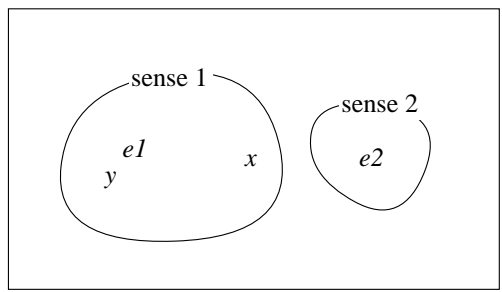}
  \end{center}
  \caption{A case where either $x$ or $y$ can be selected in
    committee-based sampling}
  \label{fig:cbs}
\end{figure}

Figure~\ref{fig:cbs} shows a typical disparity evident between
committee-based sampling and our sampling method. The basic notation
in this figure is the same as in Figure~\ref{fig:certainty}, and both
$x$ and $y$ denote unsupervised examples, or more formally
\mbox{$\set{D} = \{e_1,e_2\}$}, and \mbox{$\set{X} = \{x,y\}$}. Assume
a pair of committee members $\{e_1\}$ and $\{e_2\}$ have been selected
from the database $\set{D}$.  In this case, the committee members
disagree as to the interpretations of both $x$ and $y$, and
consequently, both examples can potentially be selected as a sample
for the next iteration. In fact, committee-based sampling tends to
require a number of similar examples (similar to $e_1$ and $y$) in the
database, otherwise committee members taken from the database will
never agree. This feature provides a salient contrast to our method
for which a similar example is less informative, and $x$ is therefore
preferred to $y$ as a sample.  This contrast can also correlate to the
fact that committee-based sampling is currently applied to
statistics-based language models (HMM classifiers), in other words,
statistical models generally require that the distribution of the
training data reflects that of the overall text.  Through this
argument, one can assume that committee-based sampling is better
suited to statistics-based systems, while our method is more suitable
for similarity-based systems.

Engelson and Dagan~\shortcite{engelson:acl-96} criticized uncertainty
sampling~\cite{lewis:sigir-94}, which they call a ``single model''
approach, as distinct from their ``multiple model'' approach:
\begin{quote}
  \it $\ldots$ sufficient statistics may yield an accurate 0.51
  probability estimate for a class $c$ in a given example, making it
  certain that $c$ is the $\underline{\it appropriate\/}$
  classification\footnote{By appropriate classification, Engelson and
    Dagan mean the classification given by a perfectly-trained
    model.}. However, the certainty that $c$ is the $\underline{\it
    correct\/}$ classification is low, since there is a 0.49 chance
  that $c$ is the wrong class for the example. A single model can be
  used to estimate only the second type of uncertainty, which does not
  correlate directly with the utility of additional training.
\end{quote}
We note that this criticism cannot be applied to our sampling method,
despite the fact that our method falls into the category of a single
model approach. In our sampling method, given sufficient statistics,
the increment of the certainty degree for unsupervised examples, i.e. 
the training utility of additional supervised examples, becomes small
(theoretically, for both similarity-based and statistics-based systems). 
As such, the utility factor can be considered to correlate directly
with additional training, for our method.

\section{Discussion}
\label{sec:sampling_discussion}

\subsection{Sense ambiguity of case fillers in selective sampling}
\label{subsec:nsa}

\begin{figure}[htbp]
  \begin{center}
    \leavevmode
    \begin{minipage}[t]{.47\textwidth}
      \centering
      \psfig{file=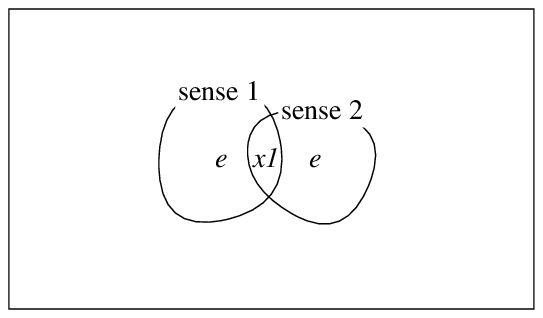,height=1.5in}
      \subcaption{Interpretation certainty of $x$ is small because $x$
        lies in the intersection of distinct verb senses}
      \label{fig:nsa-a}
    \end{minipage}
    \hfill
    \begin{minipage}[t]{.47\textwidth}
      \centering
      \psfig{file=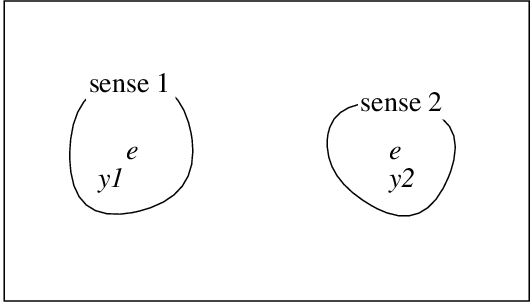,height=1.5in}
      \subcaption{Interpretation certainty of $y$ is small because $y$
        is semantically ambiguous}
      \label{fig:nsa-b} \medskip
    \end{minipage}
  \end{center}
  \caption{Two separate scenarios in which the interpretation certainty
    of $x$ is small}
  \label{fig:nsa}
\end{figure}

The semantic ambiguity of case fillers (nouns) should be taken into
account during selective sampling. Figure~\ref{fig:nsa}, which uses
the same basic notation as Figure~\ref{fig:certainty}, illustrates one
possible problem caused by case filler ambiguity.  Let $x_1$ be a
sense of a case filler $x$, and $y_1$ and $y_2$ be different senses of
a case filler $y$.  On the basis of Equation~\ref{eq:certainty}, the
interpretation certainty of $x$ and $y$ is small in
Figures~\ref{fig:nsa-a} and \ref{fig:nsa-b}, respectively.  However,
in the situation shown in Figure~\ref{fig:nsa-b}, since (a) the task
of distinguishing between the {\em verb\/} senses 1 and 2 is easier,
and (b) instances where the sense ambiguity of case fillers
corresponds to distinct verb senses will be rare, training using
either $y_1$ or $y_2$ will be less effective than using a case filler
of the type of $x$.  It should also be noted that since {\it
  Bunruigoihyo\/} fundamentally associates each word with a single
concept, this problem is not critical in our case.  However, given
other existing thesauri like the EDR electronic
dictionary~\cite{edr:95} or WordNet~\cite{miller:techrep-93}, these
two situations should be strictly differentiated.

\subsection{Applicability of our selective sampling method}
\label{subsec:limitation}

First, we note that our selective sampling method is expected to be
applicable to most similarity-based methods, or more precisely, those
which follow the nearest neighbor method, although in this chapter, we
describe the sampling method being applied only to our verb sense
disambiguation method (from Chapter~\ref{cha:vsd}). However, our
sampling method does not seem effective for some similarity-based
methods, such as the one proposed by Ng and Lee~\shortcite{ng:acl-96}.
Ng and Lee used the similarity measure as shown in
Equation~\ref{eq:cost_dist} (see page~\pageref{eq:cost_dist}).  It
should be noted that their similarity computation uses the statistics
obtained from supervised examples.  {\em Selective\/} sampling for the
supervised examples potentially biases the estimation of
\mbox{$P(\SS|F\!=\!f)$}, i.e. the conditional probability that
sense~$\SS$ occurs, given that feature~$F$ takes value~$f$. This issue
needs to be further explored.  On the other hand, our sampling method
can be applied to similarity-based methods which use resources {\em
  independent\/} of the supervised examples, for the similarity
computation~\cite{cho:nlprs-95,kurohashi:ieice-94,niwa:coling-94,uramoto:ieice-94}.

\begin{figure}[htbp]
  \begin{center}
    \leavevmode
    \psfig{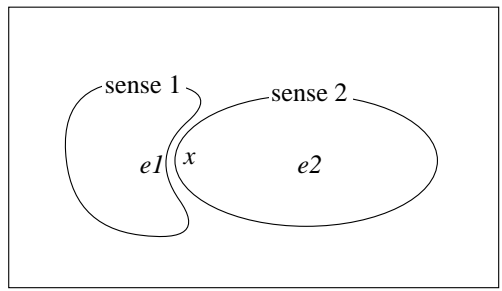}
  \end{center}
  \caption{The case where informative example $x$ is not selected}
  \label{fig:limitation}
\end{figure}

Second, Figure~\ref{fig:limitation} exemplifies a limitation of our
sampling method. The basic notation is the same as in
Figure~\ref{fig:certainty}.  In this figure, the only supervised
examples contained in the database are $e_1$ and $e_2$, and $x$
represents an unsupervised example belonging to sense 2. Given this
scenario, $x$ is informative because (a) it clearly evidences the
semantic vicinity of sense 2, and (b) without $x$ as sense 2 in the
database, the system may misinterpret other examples neighboring
$x$. However, in our current implementation, the training utility of
$x$ would be small because it would be mistakenly interpreted as sense
1 with great certainty due to its relatively close semantic proximity
to $e_1$.  Even if $x$ has a number of unsupervised neighbors, the
total increment of their interpretation certainty cannot be expected
to be large.  This shortcoming often presents itself when the semantic
vicinities of different verb senses are closely aligned or their
semantic ranges are not disjunctive.  Here, let us consider
Figure~\ref{fig:ccd} again, in which the nominative case would
parallel the semantic space shown in Figure~\ref{fig:limitation} more
closely than the accusative.  Relying more on the similarity in the
accusative (the case with greater CCD) as is done in our system, we
aim to map the semantic space in such a way as to achieve higher
semantic disparity and minimize this shortcoming.

\section{Summary}
\label{sec:sampling_summary}

Corpus-based approaches have recently pointed the way to a promising
trend in word sense disambiguation. However, these approaches tend to
require a considerable overhead for supervision in constructing a
large-sized database, additionally resulting in a computational
overhead to search the database.  To overcome these problems, our
method, which is currently applied to an similarity-based verb sense
disambiguation system, selectively samples a smaller-sized subset from
a given example set~\cite{fujii:cl-x,fujii:wvlc-96}.  This method is
expected to be applicable to other similarity-based
systems. Applicability for other types of systems needs to be further
explored.

The process basically iterates through two phases: (normal) word sense
disambiguation and a training phase.  During the disambiguation phase,
the system is provided with sentences containing a polysemous verb,
and searches the database for the most semantically similar example to
the input (nearest neighbor method).  Thereafter, the verb is
disambiguated by superimposing the sense of the verb appearing in the
supervised example.  The similarity between the input and an example,
or more precisely the similarity between the case fillers included in
them, is computed based on an existing thesaurus (or the vector space
model is alternatively applicable). In the training phase, a sample is
then selected from the system outputs and provided with the correct
interpretation by a human expert.  Through these two phases, the
system iteratively accumulates supervised examples into the database.
The critical issue in this process is to decide which example should
be selected as a sample in each iteration.  To resolve this problem,
we considered the following properties: (a) an example which neighbors
more unsupervised examples is more influential for subsequent
training, and therefore more informative, and (b) since our verb sense
disambiguation is based on nearest neighbor resolution, an example
similar to one already existing in the database is redundant.
Motivated by these properties, we introduced and formalized the
concept of training utility as the criterion for example selection.
Our sampling method always gives preference to that example which
maximizes training utility.

We reported on the performance of our sampling method by way of
experiments, in which we compared our method with random sampling,
uncertainty sampling~\cite{lewis:sigir-94} and committee-based
sampling~\cite{engelson:acl-96}. The result of the experiments showed
that our method reduced both the overhead for supervision and the
overhead for searching the database to a larger degree than any of the
above three methods, without degrading the performance of verb sense
disambiguation. Through the experiment and discussion, we claim that
uncertainty sampling considers property (b) mentioned above, but lacks
property (a). We also claim that committee-based sampling differs from
our sampling method in terms of its suitability to statistics-based
systems as compared to similarity-based systems.

\chapter{Exploration of Similarity Computation}
\label{cha:word_similarity}

\begin{quote}
  \it This chapter explores the computation of word similarity, on
  which the performance of our similarity-based verb sense
  disambiguation system is highly dependent. The statistics-based
  computation of word similarity has been popular in recent research,
  but is associated with significant computational cost.  On the other
  hand, the use of hand-crafted thesauri as semantic resources is
  simple to implement, but lacks mathematical rigor.  To integrate the
  advantages of these two approaches, we aim at calculating a
  statistical weight for each branch of a thesaurus, so that we can
  compute word similarity based simply on the length of the path
  between two words in the thesaurus. Our experiment on Japanese nouns
  shows that this framework upheld the inequality of statistics-based
  word similarity with an accuracy of more than 70\%. We also applied
  our framework of word similarity computation to our verb sense
  disambiguation system.
\end{quote}

\section{Motivation}
\label{sec:ws_intro}

As with other similarity-based systems, the performance of our system
(as described in Chapter~\ref{cha:vsd}) is highly dependent on the
computation of the (relative) similarity between two examples, which
is further decomposed into the similarity between two case filler
nouns (we shall call this ``word similarity'', hereafter).  By the
performance of the system, we refer to the following different
viewpoints:
\begin{itemize}
\item the relative number of correct outputs (``accuracy''),
\item the cost needed for the similarity computation (``efficiency'').
\end{itemize}
We described two previous approaches for word similarity computation
in Section~\ref{sec:vsd_methodology}. Here, let us summarize these
approaches again.

The first approach for word similarity computation is statistics-based
methods~\cite{brown:cl-92,charniak:93,dagan:acl-94,grishman:coling-94,hindle:acl-90,pereira:acl-93,tokunaga:ijcai-95}.
From a number of different implementations, we used the vector space
model
(VSM)~\cite{frakes:92,leacock:arpa-hlt-93,niwa:coling-94,salton:83,schutze:supercomp-92},
for our verb sense disambiguation system. In the vector space model,
each word is represented by a vector consisting of co-occurrence
statistics, such as relative frequency, with respect to other words.
Note that in our case, we used only the co-occurrence between a noun
and verbs. The similarity between two given words is then measured
computationally using the two vectors representing those words.  One
typical implementation computes the similarity as the cosine of the
angle between the two vectors, a method which is also commonly used in
information retrieval and text categorization systems to measure the
similarity between documents~\cite{salton:83}. Since it is based on
mathematical rationale, this type of similarity measurement has been
popular. Besides this, since the similarity is computed based on given
co-occurrence data, word similarity can easily be adjusted according
to the domain. However, data sparseness is an inherent problem. This
fact was observed in our preliminary experiment (see
Section~\ref{sec:vsd_experiment}), despite using statistical
information taken from news articles for as many as {\em four\/}
years.  Furthermore, in this approach, each vector requires
\mbox{$O(N)$} computational cost, given that $N$ is the length of the
vector. Note that although, in our case, each vector term is confined
to a verb, additional parts-of-speech for vector terms are
problematic. In particular, the vector length can potentially be
overly great when the vectors comprise {\em noun\/} terms, because
nouns are generally constitute an open class.

The other category of word similarity approach uses a hand-crafted
thesaurus (such as Roget's thesaurus~\cite{chapman:84} or
WordNet~\cite{miller:techrep-93} in the case of English, and {\it
  Bunruigoihyo\/}~\cite{bgh:64} or EDR~\cite{edr:95} in the case of
Japanese), based on the intuitively feasible assumption that words
located near each other within the structure of a thesaurus have
similar meaning.  Therefore, the similarity between two given words is
represented by the length of the path between them in the thesaurus
structure~\cite{kurohashi:ieice-94,x.li:ijcai-95,uramoto:ieice-94}.
Unlike the former approach, the required computational cost can be
restricted to constant order, because only a list of semantic codes
for each word is required. For example, the commonly used Japanese
{\it Bunruigoihyo\/} thesaurus~\cite{bgh:64} represents each semantic
code with only 7 digits\footnote{The revised version of the {\it
    Bunruigoihyo\/} thesaurus assigns an 8 digit code each word.}.
However, computationally speaking, the relation between the similarity
(namely the semantic length of the path), and the physical length of
the path is not clear\footnote{Most researchers {\em heuristically\/}
  define functions between the similarity and physical path
  length~\cite{kurohashi:coling-92,kurohashi:ieice-94,x.li:ijcai-95,uramoto:ieice-94}.}.
Furthermore, since most thesauri aim at a {\it general\/} word
hierarchy, the similarity between words used in specific domains
(technical terms) cannot be measured to the desired level of accuracy.

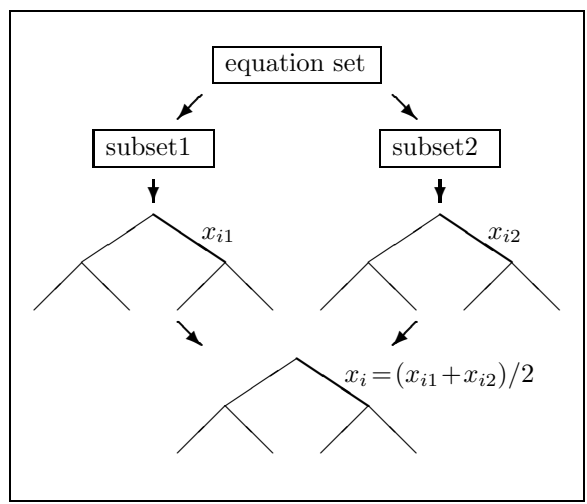
\begin{figure}[htbp]
  \begin{center}
    \leavevmode
    \input{thesaurus.latex}
  \end{center}
  \caption{A fragment of the thesaurus associated with $x_1$ to $x_6$}
  \label{fig:ws_thesaurus}
\end{figure}

To sum up, the use of hand-crafted thesauri is preferable to
statistics-based approaches in terms of ``efficiency''. As for the
``accuracy'', while our experiments in
Section~\ref{sec:vsd_experiment} quantitatively showed that the use of
the {\it Bunruigoihyo\/} thesaurus outperformed the vector space
model, there has been no qualitative explanation for this general
tendency. In addition, new and improved models can potentially be
introduced for statistics-based approaches.  In view of these
arguments, we aim at intergrating the advantages of the two above
methodological types, or more precisely, realizing statistics-based
word similarity based on the length of the thesaurus
path~\cite{fujii:acl-eacl-ws-97}. Consequently, our approach allows us
to measure the statistics-based word similarity, while retaining
optimal computational cost (\mbox{$O(1)$}).  The crucial concern in
this process is how to determine the `statistics-based length' (SBL)
of each branch in a thesaurus.  We tentatively use the {\it
  Bunruigoihyo\/} thesaurus, in which each word corresponds to a leaf
in the tree structure.  Let us take Figure~\ref{fig:ws_thesaurus},
which shows a fragment of the thesaurus. In this figure, $w_i$'s
denote words and $x_i$'s denote the statistics-based length of each
branch $i$.  Let the statistics-based (vector space model) word
similarity between $w_1$ and $w_2$ be \mbox{$vsm(w_1, w_2)$}.  We hope
to estimate this similarity by the length of the path through branches
3 and 4, and derive an equation \mbox{``$x_3+x_4 = sim(w_1,w_2)$''}.
Intuitively speaking, any combination of $x_3$ and $x_4$ which
satisfies this equation can constitute the SBLs for branches 3 and 4.
Formalizing equations for other pairs of words in the same manner, we
can derive the set of simultaneous equations shown in
Figure~\ref{fig:ws_simeq}. That is, we can assign the SBL for each
branch by way of finding answers for each $x_i$.

\begin{figure}[htbp]
  \begin{center}
    \leavevmode \framebox{\parbox{7cm}{
    \begin{eqnarray*}
      x_1 + x_2 + x_3 + x_5 & = & vsm(w_1,w_3) \\
      x_1 + x_2 + x_3 + x_6 & = & vsm(w_1,w_4) \\
      x_1 + x_2 + x_4 + x_5 & = & vsm(w_2,w_3) \\
      & \vdots &
    \end{eqnarray*}
    }}
  \end{center}
  \caption{A fragment of the set of simultaneous equations associated
    with Figure~\protect\ref{fig:ws_thesaurus}}
  \label{fig:ws_simeq}
\end{figure}

In Section~\ref{sec:ws_methodology}, we elaborate on the methodology
of our word similarity measurement. We then evaluate our method by way
of an experiment in Section~\ref{sec:ws_experiment} and apply this
method to the task of word sense disambiguation in
Section~\ref{sec:ws_application}.

\section{Methodology}
\label{sec:ws_methodology}

\subsection{Overview}
\label{subsec:ws_overview}

Our word similarity measurement proceeds in the following way:
\begin{enumerate}
\item compute the statistics-based similarity of every combination of
  given words,
\item set up a set of simultaneous equations through use of the
  thesaurus and previously computed word similarities, and find
  solutions for the statistics-based length (SBL) of the corresponding
  thesaurus branch (see Figures~\ref{fig:ws_thesaurus} and
  \ref{fig:ws_simeq}),
\item the similarity between two given words is measured by the sum of
  SBLs included in the path between those words.
\end{enumerate}
For step~1, we used the vector space model, for which the reader is
referred to Section~\ref{subsec:simcomp}. However, note that our
framework is independent of the implementation of the similarity
computation. We elaborate on steps 2 and 3 in the following sections.

\subsection{Resolution of the simultaneous equations}
\label{subsec:ws_resolution}

\begin{figure}[htbp]
  \begin{center}
    \leavevmode
    \input{subsets.latex}
  \end{center}
  \caption{Approximation of the statistics-based length $x_i$}
  \label{fig:ws_subsets}
\end{figure}
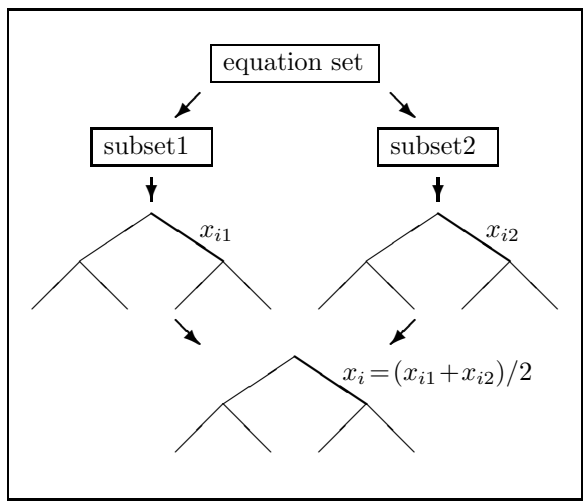

The set of simultaneous equations used in our method is expressed by
Equation~\ref{eq:ws_simeq}, where $A$ is a matrix comprising only the
values 0 and 1, and $B$ is the list of $vsm$'s (i.e. statistics-based
similarities) for all possible combinations of given words.  $X$ is a
list of variables, which represents the statistics-based length (SBL)
for the corresponding branch in the thesaurus.
\begin{equation}
  \label{eq:ws_simeq}
  AX = B
\end{equation}
Here, let the $i$-th similarity in $B$ be \mbox{$vsm(a,b)$}, and let
\mbox{$path(a,b)$} denote the path between words $a$ and $b$ in the
thesaurus.  Each equation contained in the set of simultaneous
equations is represented by Equation~\ref{eq:ws_eq}, where $x_j$ is
the statistics-based length (SBL) for branch $j$, and $\alpha_{ij}$ is
either 0 or 1, as in Equation~\ref{eq:ws_a}.
\begin{equation}
  \label{eq:ws_eq}
  \left[\alpha_{i1}~\alpha_{i2}~\ldots~\alpha_{ij}~\ldots\right]
    \left[
      \begin{array}{c}
        x_1 \\ x_2 \\ \vdots \\ x_j \\ \vdots
      \end{array}
    \right] = vsm(a,b)
\end{equation}
\begin{equation}
  \label{eq:ws_a}
  \alpha_{ij} = \left\{
  \begin{array}{ll}
    1 & \mbox{if $j\in path(a,b)$} \\ 0 & \mbox{otherwise}
  \end{array}
  \right.
\end{equation}
By finding the solutions for $X$, we can assign SBLs to branches.
However, the set of similarity values outnumbers the variables. For
example, the revised version of the {\it Bunruigoihyo\/} thesaurus
contains about 55,000 noun entries, and therefore, the number of
similarity values for those nouns becomes about 1.5$\times$10$^9$
($C_{2}^{55,000}$). On the other hand, the number of the branches is
only about 53,000. As such, overly many equations are redundant, and
the time complexity to solve the simultaneous equations becomes a
crucial problem.  To counter this problem, we randomly divide the
overall equation set into equal parts, which can be solved reasonably.
Thereafter we approximate the solution for $x$ by averaging the
solutions for $x$ derived from each subset. Let us take
Figure~\ref{fig:ws_subsets}, in which the number of subsets is given
as two without loss of generality. In this figure, $x_{i1}$ and
$x_{i2}$ denote the answers for branch $i$ individually derived from
subsets 1 and 2, and $x_i$ is approximated by the average of $x_{i1}$
and $x_{i2}$ (that is, \mbox{$\frac{x_{i1}+x_{i2}}{2}$}).  To
generalize this notion, let $x_{ij}$ denote the solution associated
with branch $i$ in subset $j$. The approximate solution for branch $i$
is given by Equation~\ref{eq:ws_answer}, where $n$ is the number of
divisions of the equation set.
\begin{equation}
  \label{eq:ws_answer}
  x_i = \frac{\textstyle 1}{\textstyle n}\sum_{j=1}^{n} x_{ij}
\end{equation}
With regard to resolving the simultaneous equations, we used the
mathematical analysis tool ``MATLAB''\footnote{Developed by Cybernet
  System, Inc.}.

\subsection{Word similarity using SBL}
\label{subsec:ws_sim_sbl}

Let us reconsider Figure~\ref{fig:ws_thesaurus}. In this figure, the
similarity between $w_1$ and $w_2$, for example, is measured by the
sum of $x_3$ and $x_4$.  In general, the similarity between words $a$
and $b$ using SBL (\mbox{$sbl(a,b)$}, hereafter) is realized by
Equation~\ref{eq:ws_sbl}, where $x_i$ is the SBL for branch $i$, and
\mbox{$path(a,b)$} is the path that includes thesaurus branches
located between $a$ and $b$.
\begin{equation}
  \label{eq:ws_sbl}
  sbl(a,b) = \sum_{i\in path(a,b)} x_i
\end{equation}

\section{Experimentation}
\label{sec:ws_experiment}

\begin{figure}[htbp]
  \begin{center}
    \leavevmode
    \psfig{file=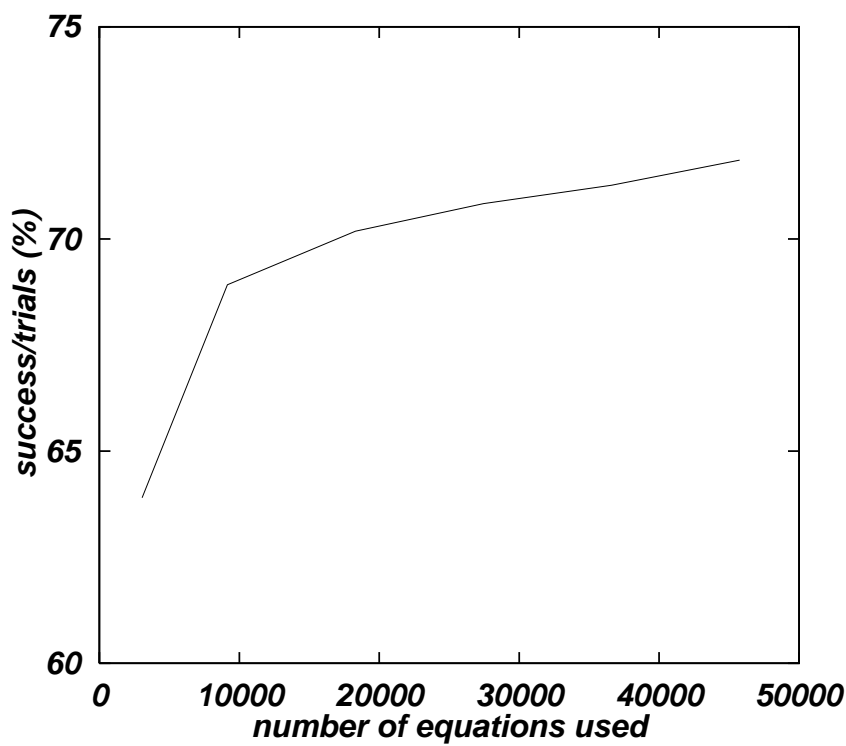,height=4in}
  \end{center}
  \caption{The relation between the number of equations used and the
    accuracy}
  \label{fig:ws_experiments}
\end{figure}

Here, we evaluate the degree to which simultaneous equations are
successfully approximated through the use of the technique described
in Section~\ref{sec:ws_methodology}. In other words, we analyze to
what extent the (original) statistics-based word similarity can be
realized by our framework.  We conducted this evaluation in the
following way.  Let the statistics-based similarity between words $a$
and $b$ be \mbox{$vsm(a,b)$}, and the similarity based on SBL be
\mbox{$sbl(a,b)$}.  Here, let us assume the inequality
\mbox{``$vsm(a,b) > vsm(c,d)$''} for words $a$, $b$, $c$ and $d$. If
this inequality can be maintained for our method, that is,
\mbox{``$sbl(a,b) > sbl(c,d)$''}, the similarity measurement is taken
to be successful.  The accuracy is then estimated by the ratio between
the number of successful measurements and the total number of
trials. Since resolution of equations is time-consuming, we
tentatively generalized 23,223 nouns into 303 semantic classes
(represented by the first 4 digits of the semantic code given in the
{\it Bunruigoihyo\/} thesaurus), reducing the total number of
equations to 45,753.  Figure~\ref{fig:ws_experiments} shows the
relation between the number of equations used and the accuracy: we
divided the overall equation set into $n$ equal subsets\footnote{We
  arbitrarily set \mbox{$n = 15$} so as to be able to resolve
  equations reasonably.} (see Section~\ref{subsec:ws_resolution}), and
progressively increased the number of subsets used in the computation.
When the whole set of equations was provided, the accuracy became
about 72\%.  We also estimated the lower bound of this evaluation,
that is, we also conducted the same trials using the {\it
  Bunruigoihyo\/} thesaurus. In this case, if word $a$ is more closely
located to $b$ than $c$ is to $d$ and \mbox{``$vsm(a,b) >
  vsm(c,d)$''}, that trial measurement is taken to be successful.  We
found that the lower bound method produced an accuracy of about 56\%,
and therefore, that our framework outperformed this method.

\section{Application to verb sense disambiguation system}
\label{sec:ws_application}

We further evaluated our word similarity technique in the task of verb
sense disambiguation. The evaluation methodology is simple, that is,
the performance of word similarity measurement is evaluated in the
context of the improvement on our similarity-based system. In
practice, Equation~\ref{eq:ws_sbl} is used as a substitution for $sim$
in Equation~\ref{eq:sim} (see Section~\ref{sec:vsd_methodology} for
details).  As performed in Section~\ref{sec:vsd_experiment}, this
experiment involved six-fold cross validation, that is, we divided the
training/test data into six equal parts, and conducted six trials in
which a different part was used as test data each time, and the rest
as the database.  We evaluated the performance of the system according
to its accuracy, that is the ratio of the number of correct outputs,
compared to the total number of inputs.  We used as training/test data
the same corpus as in Section~\ref{sec:vsd_experiment}, where
sentences contain one of eleven polysemous verbs, producing a total of
1315 sentences (see Table~\ref{tab:corpus}).  We found that the
combined performance with SBLs did not differ from that for the {\it
  Bunruigoihyo\/} thesaurus (see the column ``BGH'' in
Table~\ref{tab:corpus}), which means our method still finds it
difficult to enhance existing hand-crafted thesauri.  However, it
should be noted that given the inferiority of the vector space model
as shown in Table~\ref{tab:corpus}, our word similarity measure cannot
theoretically exceed the performance for the {\it Bunruigoihyo\/}
thesaurus, because the objective of SBL is to {\em simulate\/} word
similarity based on the vector space model. The use of SBL is expected
to improve other applications where the vector space model outperforms
hand-crafted thesauri.

\section{Related Work}
\label{sec:ws_related}

In this chapter, we focused solely on the ``efficiency'' problem
related to the word similarity computation, disregarding the
``accuracy'' problem. Resnik~\shortcite{resnik:ijcai-95} (later
enhanced by Jiang and Conrath~\shortcite{jiang:rocling-97}) integrates
a thesaurus taxonomy and information content for similarity
computation, focusing on the accuracy problem.  Resnik's method
computes similarity between given word classes (synsets) taken from
WordNet~\cite{miller:techrep-93}, based on the degree of {\em
  commonality\/} between them. In other words, the more information
two synsets share in common, the more similar they are.  Formally
speaking, the similarity between two synsets~$c_1$~and~$c_2$ is
computed by Equation~\ref{eq:ws_resnik}.
\begin{equation}
  \label{eq:ws_resnik}
  sim(c_1,c_2) = \max_{c\in S(c_1,c_2)} [-\log P(c)]
\end{equation}
Here, \mbox{$S(c_1,c_2)$} is the set of synsets that dominate both
$c_1$ and $c_2$. \mbox{$P(c)$} is estimated based on the distribution
of words associated with synset~$c$, obtained from a corpus.
Resnik~\shortcite{resnik:ijcai-95} (and Jiang and
Conrath~\shortcite{jiang:rocling-97}) showed the effectivity of this
method using 30 pairs of words, which seems a relatively small data
size. We applied Resnik's method to our verb sense disambiguation
system, using the {\it Bunruigoihyo\/} as a core thesaurus.  We
conducted the same procedure as performed in
Section~\ref{sec:vsd_experiment}, and found that the accuracy of verb
sense disambiguation was about 23\%. It should be noted that this
result represents only one example among the numerous usages of word
similarity, and does not discount the validity of Resnik's method.
However, we claim that enhancement of the accuracy problem for
similarity computation still remains difficult, and needs to be
further explored.

\section{Summary}
\label{sec:ws_summary}

The previous methods for word similarity computation can be divided
into two approaches.  Statistics-based approaches have been popular
because of their mathematical rigor. However, these approaches
generally require prohibitive computational cost. Remaining
approaches, which use a hand-crafted thesaurus, are computationally
cheaper. However, the quality of these approaches is highly dependent
on a priori human judgement. In this chapter, we proposed a new method
to integrate two different approaches for word similarity
computation~\cite{fujii:acl-eacl-ws-97}. That is, we realized
statistics-based word similarity computation with complexity
equivalent to that for thesaurus-based approaches. Our objective is to
determine the statistical weight for each thesaurus branch, so that
one can measure statistics-based similarity by simply traversing the
thesaurus structure. For this purpose, we set up a set of simultaneous
equations, in which each variable and value correspond to a thesaurus
branch and statistics-based similarity, respectively. By resolving the
simultaneous equations, we can expect to assign appropriate
statistical weights to each branch. By way of this method, we can
expect to optimize the computational efficiency required for our verb
sense disambiguation.

\chapter{Conclusion}
\label{cha:conclusion}

\section{Contribution}
\label{sec:contribution}

In this research, we targetted various problems associated with
corpus-based word sense disambiguation, focusing on verbal
polysemy. Let us summarize the main points that have been made in this
research.
\begin{itemize}
\item First, in Chapter~\ref{cha:vsd}, we described our
  similarity-based verb sense disambiguation system and demonstrated
  its effectivity. One of the major features of this chapter was that
  we modeled and computationally implemented the linguistic behavior
  of `case contribution to disambiguation' (CCD). Through comparative
  experiments, we confirmed that the performance of our system was
  improved by considering the CCD factor. Our experiments also showed
  that a hand-crafted thesaurus (in our case, the {\it Bunruigoihyo\/}
  thesaurus~\cite{bgh:64}) is an effective resource for word sense
  disambiguation. Due to the lack of a large-sized sense-annotated
  test collection for Japanese, we used a relatively small-sized
  corpus for our experimentation. However, our experimental results
  are expected to reflect general performance tendencies to a certain
  degree.  We further enhanced our system by way of computation of the
  degree of interpretation certainty, so as to obtain more reliable
  outputs. In addition, we proposed a prototype method of propagating
  contextual constraints when a unique polysemous verb appears in an
  input sentence.  It should be noted that while our experiments were
  carried out on Japanese, our proposed system can be applied to other
  languages.
\item Although the performance of our similarity-based system proved
  to be relatively satisfactory, we identified the overhead for manual
  sense annotation (overhead for supervision) and the overhead
  searching a database (overhead for search) as being limitations on
  operational applications.  Given this observation, in
  Chapter~\ref{cha:sampling}, we proposed a novel sampling method,
  which minimizes the overhead for supervision and the overhead for
  search. Note that these overheads can be major drawbacks for
  corpus-based NLP approaches in general.  Our sampling method is
  characterized by its reliance on the notion of `training utility':
  the degree to which each example is informative for future sampling
  when incorporated into the database.  Comparative experiments showed
  that our sampling method reduced both overheads to a larger degree
  than any existing sampling methods, without degrading the
  performance of verb sense disambiguation.  It should be noted that
  example sets sampled by our method can also be used as initial
  training data to further improve the quality of unsupervised
  learning, as occurs in bootstrapping.
\item Finally, in Chapter~\ref{cha:word_similarity}, we proposed a new
  method for word similarity computation, given that the performance
  of our system is highly dependent on the similarity computation
  between given words. Our method integrates the advantages of the two
  different approaches for similarity computation: the mathematical
  rigor of statistics-based approaches and easy implementation of
  thesaurus-based approaches. The core process is to determine the
  `statistics-based length' (SBL) of each branch in a thesaurus,
  reflecting statistically computed word similarities. In this way,
  our method realizes statistics-based word similarity by simply
  traversing the thesaurus structure.
\end{itemize}

\section{Outstanding Issues}
\label{sec:future}

Out standing issue is the problem of balancing qualitative and
quantitative
approaches~\cite{basili:ai-96,klavans:96,uramoto:ieice-94} (see
Chapter~\ref{subsec:classification} for a description of these
approaches). Among a number of past research attempts to tackle this
issue, Uramoto~\shortcite{uramoto:ieice-94}, for example, used
selectional restrictions and scalable similarity based on a thesaurus
taxonomy, as (relatively shallow) constraints and preference
conditions, respectively. However, given that Uramoto did not compare
his integrated approach with monotonic approaches, the empirical study
of balancing different approaches requires further exploration and is
expected to improve on current word sense disambiguation methods.

Another direction would be the establishment of standardized
evaluation criteria for word sense disambiguation systems, which is
expected to streamline system comparison/enhancement\footnote{ACL
  SIGLEX is going to hold a workshop (co-ordinated by Adam Kilgarriff)
  in September 1998, to initiate the standardized evaluation of word
  sense disambiguation systems for English, French, German, Italian
  and Spanish.}. However, as discussed in
Section~\ref{sec:evaluation_methodology}, numerous problems exist
associated with evaluation methodologies. Ultimately speaking, we
think that task-based (application-based) evaluation is the most
straightforward, because the performance of a system is directly
related to benefits to real-world applications.  In addition, word
sense distinction, which can be one major obstacle in standardized
evaluation, can be fixed relatively easily for specific
applications. However, on the other hand, numerous sense-annotated
corpora are required for different applications. Our sampling method
(as well as past efforts targetting unsupervised methods) are expected
to reduce the human overhead for establishing these types of
sense-annotated corpora.

\renewcommand{\baselinestretch}{1}
\cleardoublepage
\addcontentsline{toc}{chapter}{Bibliography}
\bibliographystyle{jplain}

\end{document}

%% file: thesaurus.latex
\setlength{\unitlength}{0.012500in}%
\begingroup\makeatletter\ifx\SetFigFont\undefined
\def\x#1#2#3#4#5#6#7\relax{\def\x{#1#2#3#4#5#6}}%
\expandafter\x\fmtname xxxxxx\relax \def\y{splain}%
\ifx\x\y   
\gdef\SetFigFont#1#2#3{%
  \ifnum #1<17\tiny\else \ifnum #1<20\small\else
  \ifnum #1<24\normalsize\else \ifnum #1<29\large\else
  \ifnum #1<34\Large\else \ifnum #1<41\LARGE\else
     \huge\fi\fi\fi\fi\fi\fi
  \csname #3\endcsname}%
\else
\gdef\SetFigFont#1#2#3{\begingroup
  \count@#1\relax \ifnum 25<\count@\count@25\fi
  \def\x{\endgroup\@setsize\SetFigFont{#2pt}}%
  \expandafter\x
    \csname \romannumeral\the\count@ pt\expandafter\endcsname
    \csname @\romannumeral\the\count@ pt\endcsname
  \csname #3\endcsname}%
\fi
\fi\endgroup
\begin{picture}(240,205)(10,600)
\thicklines
\multiput( 80,675)(0.40000,-0.40000){26}{\makebox(0.4444,0.6667){\SetFigFont{7}{8.4}{rm}.}}
\put( 90,665){\vector( 1,-1){0}}
\multiput(180,675)(-0.40000,-0.40000){26}{\makebox(0.4444,0.6667){\SetFigFont{7}{8.4}{rm}.}}
\put(170,665){\vector(-1,-1){0}}
\put( 70,735){\vector( 0,-1){ 10}}
\put(190,735){\vector( 0,-1){ 10}}
\multiput( 90,770)(-0.40000,-0.40000){26}{\makebox(0.4444,0.6667){\SetFigFont{7}{8.4}{rm}.}}
\put( 80,760){\vector(-1,-1){0}}
\multiput(170,770)(0.40000,-0.40000){26}{\makebox(0.4444,0.6667){\SetFigFont{7}{8.4}{rm}.}}
\put(180,760){\vector( 1,-1){0}}
\thinlines
\put( 10,600){\framebox(240,205){}}
\put( 45,740){\framebox(50,15){}}
\put(165,740){\framebox(50,15){}}
\put( 20,680){\line( 1, 1){ 20}}
\put( 40,700){\line( 1,-1){ 20}}
\put( 80,680){\line( 1, 1){ 20}}
\put(100,700){\line( 1,-1){ 20}}
\put( 40,700){\line( 3, 2){ 30}}
\put(140,680){\line( 1, 1){ 20}}
\put(160,700){\line( 1,-1){ 20}}
\put(200,680){\line( 1, 1){ 20}}
\put(220,700){\line( 1,-1){ 20}}
\put(160,700){\line( 3, 2){ 30}}
\put( 80,620){\line( 1, 1){ 20}}
\put(100,640){\line( 1,-1){ 20}}
\put(140,620){\line( 1, 1){ 20}}
\put(160,640){\line( 1,-1){ 20}}
\put(100,640){\line( 3, 2){ 30}}
\thicklines
\put( 70,720){\line( 3,-2){ 30}}
\put(190,720){\line( 3,-2){ 30}}
\put(130,660){\line( 3,-2){ 30}}
\thinlines
\put( 95,775){\framebox(70,15){}}
\put( 90,710){\makebox(0,0)[lb]{\smash{\SetFigFont{10}{12.0}{it}$x_{i1}$}}}
\put(210,710){\makebox(0,0)[lb]{\smash{\SetFigFont{10}{12.0}{it}$x_{i2}$}}}
\put(150,650){\makebox(0,0)[lb]{\smash{\SetFigFont{10}{12.0}{it}$x_i\!=\!(x_{i1}\!+\!x_{i2})/2$}}}
\put( 50,745){\makebox(0,0)[lb]{\smash{\SetFigFont{10}{12.0}{rm}subset1}}}
\put(170,745){\makebox(0,0)[lb]{\smash{\SetFigFont{10}{12.0}{rm}subset2}}}
\put(100,780){\makebox(0,0)[lb]{\smash{\SetFigFont{10}{12.0}{rm}equation set}}}
\end{picture}

%% file: subsets.latex
\setlength{\unitlength}{0.012500in}%
\begingroup\makeatletter\ifx\SetFigFont\undefined
\def\x#1#2#3#4#5#6#7\relax{\def\x{#1#2#3#4#5#6}}%
\expandafter\x\fmtname xxxxxx\relax \def\y{splain}%
\ifx\x\y   
\gdef\SetFigFont#1#2#3{%
  \ifnum #1<17\tiny\else \ifnum #1<20\small\else
  \ifnum #1<24\normalsize\else \ifnum #1<29\large\else
  \ifnum #1<34\Large\else \ifnum #1<41\LARGE\else
     \huge\fi\fi\fi\fi\fi\fi
  \csname #3\endcsname}%
\else
\gdef\SetFigFont#1#2#3{\begingroup
  \count@#1\relax \ifnum 25<\count@\count@25\fi
  \def\x{\endgroup\@setsize\SetFigFont{#2pt}}%
  \expandafter\x
    \csname \romannumeral\the\count@ pt\expandafter\endcsname
    \csname @\romannumeral\the\count@ pt\endcsname
  \csname #3\endcsname}%
\fi
\fi\endgroup
\begin{picture}(240,205)(10,600)
\thicklines
\multiput( 80,675)(0.40000,-0.40000){26}{\makebox(0.4444,0.6667){\SetFigFont{7}{8.4}{rm}.}}
\put( 90,665){\vector( 1,-1){0}}
\multiput(180,675)(-0.40000,-0.40000){26}{\makebox(0.4444,0.6667){\SetFigFont{7}{8.4}{rm}.}}
\put(170,665){\vector(-1,-1){0}}
\put( 70,735){\vector( 0,-1){ 10}}
\put(190,735){\vector( 0,-1){ 10}}
\multiput( 90,770)(-0.40000,-0.40000){26}{\makebox(0.4444,0.6667){\SetFigFont{7}{8.4}{rm}.}}
\put( 80,760){\vector(-1,-1){0}}
\multiput(170,770)(0.40000,-0.40000){26}{\makebox(0.4444,0.6667){\SetFigFont{7}{8.4}{rm}.}}
\put(180,760){\vector( 1,-1){0}}
\thinlines
\put( 10,600){\framebox(240,205){}}
\put( 45,740){\framebox(50,15){}}
\put(165,740){\framebox(50,15){}}
\put( 20,680){\line( 1, 1){ 20}}
\put( 40,700){\line( 1,-1){ 20}}
\put( 80,680){\line( 1, 1){ 20}}
\put(100,700){\line( 1,-1){ 20}}
\put( 40,700){\line( 3, 2){ 30}}
\put(140,680){\line( 1, 1){ 20}}
\put(160,700){\line( 1,-1){ 20}}
\put(200,680){\line( 1, 1){ 20}}
\put(220,700){\line( 1,-1){ 20}}
\put(160,700){\line( 3, 2){ 30}}
\put( 80,620){\line( 1, 1){ 20}}
\put(100,640){\line( 1,-1){ 20}}
\put(140,620){\line( 1, 1){ 20}}
\put(160,640){\line( 1,-1){ 20}}
\put(100,640){\line( 3, 2){ 30}}
\thicklines
\put( 70,720){\line( 3,-2){ 30}}
\put(190,720){\line( 3,-2){ 30}}
\put(130,660){\line( 3,-2){ 30}}
\thinlines
\put( 95,775){\framebox(70,15){}}
\put( 90,710){\makebox(0,0)[lb]{\smash{\SetFigFont{10}{12.0}{it}$x_{i1}$}}}
\put(210,710){\makebox(0,0)[lb]{\smash{\SetFigFont{10}{12.0}{it}$x_{i2}$}}}
\put(150,650){\makebox(0,0)[lb]{\smash{\SetFigFont{10}{12.0}{it}$x_i\!=\!(x_{i1}\!+\!x_{i2})/2$}}}
\put( 50,745){\makebox(0,0)[lb]{\smash{\SetFigFont{10}{12.0}{rm}subset1}}}
\put(170,745){\makebox(0,0)[lb]{\smash{\SetFigFont{10}{12.0}{rm}subset2}}}
\put(100,780){\makebox(0,0)[lb]{\smash{\SetFigFont{10}{12.0}{rm}equation set}}}
\end{picture}